\documentclass[twoside,11pt]{article}

\def\useJmlr{1}  %

\usepackage{jmlr2e}

\usepackage[utf8]{inputenc}

\usepackage{amsmath}

\usepackage{color}

\usepackage{subcaption} 
\usepackage{booktabs}
\usepackage{float}

\usepackage{array}
\usepackage{lastpage}

\PassOptionsToPackage{hyphens}{url}\usepackage{hyperref}
\usepackage{url}

\newcommand\vocab{\emph}
\newcommand\risk{L}
\newcommand\emprisk{\hat{L}}
\newcommand\objective{J}
\newcommand\opt[1]{{#1}^\star}
\newcommand\loss{\ell}
\newcommand\grad{\nabla}
\newcommand\defeq{=}

\newcommand\R{\mathbb{R}}
\newcommand\E{\mathop{\mathbb{E}}}

\newcommand\mD{\mathcal{D}}

\newcommand\mX{\mathcal{X}}
\newcommand\mY{\mathcal{Y}}
\newcommand\mZ{\mathcal{Z}}

\newcommand{\nth}{\textsuperscript{th}}

\newcommand*\samethanks[1][\value{footnote}]{\footnotemark[#1]}

\if \useJmlr 1

    \jmlrheading{20}{2019}{1-\pageref{LastPage}}{11/18}{7/19}{18-789}{Christopher J.~Shallue, Jaehoon Lee, Joseph Antognini, Jascha Sohl-Dickstein, Roy Frostig, and George E.~Dahl}

    \ShortHeadings{Measuring the Effects of Data Parallelism on Neural Network Training}{Shallue, Lee, Antognini, Sohl-Dickstein, Frostig, and Dahl}
    
    \firstpageno{1}
\fi

\begin{document}

\title{Measuring the Effects of Data Parallelism\\on Neural Network Training}

\author{\name Christopher J.~Shallue\thanks{Both authors contributed equally.} \email shallue@google.com
      \AND \name Jaehoon Lee\samethanks[1]\, \thanks{Work done as a member of the Google AI Residency program \href{https://g.co/airesidency}{(g.co/airesidency)}.} \email jaehlee@google.com
      \AND \name Joseph Antognini\samethanks[2] \email joe.antognini@gmail.com
      \AND \name Jascha Sohl-Dickstein \email jaschasd@google.com
      \AND \name Roy Frostig \email frostig@google.com
      \AND \name George E.~Dahl \email gdahl@google.com
      \AND
      \addr{Google Brain\\
      1600 Amphiteatre Parkway\\
      Mountain View, CA, 94043, USA}}

\editor{Rob Fergus}

\maketitle  %

\begin{abstract}%
    Recent hardware developments have dramatically increased the scale of data parallelism available for neural network training. Among the simplest ways to harness next-generation hardware is to increase the batch size in standard mini-batch neural network training algorithms. In this work, we aim to experimentally characterize the effects of increasing the batch size on training time, as measured by the number of steps necessary to reach a goal out-of-sample error.
    We study how this relationship varies with the training algorithm, model, and data set, and find extremely large variation between workloads. Along the way, we show that disagreements in the literature on how batch size affects model quality can largely be explained by differences in metaparameter tuning and compute budgets at different batch sizes. We find no evidence that larger batch sizes degrade out-of-sample performance. Finally, we discuss the implications of our results on efforts to train neural networks much faster in the future. Our experimental data is publicly available as a database of 71,638,836 loss measurements taken over the course of training for 168,160 individual models across 35 workloads.
\end{abstract}

\if \useJmlr 1
\fi

\section{Introduction}

Neural networks have become highly effective at a wide variety of prediction tasks, including image classification, machine translation, and speech recognition. The dramatic improvements in predictive performance over the past decade have partly been driven by advances in hardware for neural network training, which have enabled larger models to be trained on larger datasets than ever before. However, although modern GPUs and custom accelerators have made training neural networks orders of magnitude faster, training time still limits both the predictive performance of these techniques and how widely they can be applied. For many important problems, the best models are still improving at the end of training because practitioners cannot afford to wait until the performance saturates. In extreme cases, training must end before completing a single pass over the data \citep[e.g.][]{anil2018large}. Techniques that speed up neural network training can significantly benefit many important application areas. Faster training can facilitate dramatic improvements in model quality by allowing practitioners to train on more data \citep{hestness2017deep}, and by decreasing the experiment iteration time, allowing researchers to try new ideas and configurations more rapidly. Faster training can also allow neural networks to be deployed in settings where models have to be updated frequently, for instance when new models have to be produced when training data get added or removed.

\vocab{Data parallelism} is a straightforward and popular way to accelerate neural network training. For our purposes, data parallelism refers to distributing training examples across multiple processors to compute gradient updates (or higher-order derivative information) and then aggregating these locally computed updates. As long as the training objective decomposes into a sum over training examples, data parallelism is model-agnostic and applicable to any neural network architecture. In contrast, the maximum degree of \vocab{model parallelism} (distributing parameters and computation across different processors for the same training examples) depends on the model size and structure. Although data parallelism can be simpler to implement, ultimately, large scale systems should consider all types of parallelism at their disposal. In this work, we focus on the costs and benefits of data parallelism in the synchronous training setting. 

Hardware development is trending towards increasing capacity for data parallelism in neural network training. Specialized systems using GPUs or custom ASICs \citep[e.g.][]{jouppi2017datacenter} combined with high-performance interconnect technology are unlocking unprecedented scales of data parallelism where the costs and benefits have not yet been well studied. On the one hand, if data parallelism can provide a significant speedup at the limits of today's systems, we should build much bigger systems. On the other hand, if additional data parallelism comes with minimal benefits or significant costs, we might consider designing systems to maximize serial execution speed, exploit other types of parallelism, or even prioritize separate design goals such as power use or cost.

There is considerable debate in the literature about the costs and benefits of data parallelism in neural network training and several papers take seemingly contradictory positions. Some authors contend that large-scale data parallelism is harmful in a variety of ways, while others contend that it is beneficial. 
The range of conjectures, suggestive empirical results, and folk knowledge seems to cover most of the available hypothesis space. Answering these questions definitively has only recently become important (as increasing amounts of data parallelism have become practical), so it is perhaps unsurprising that the literature remains equivocal, especially in the absence of sufficiently comprehensive experimental data.

In this work, we attempt to provide the most rigorous and extensive experimental study on the effects of data parallelism on neural network training to date. In order to achieve this goal, we consider realistic workloads up to the current limits of data parallelism. We try to avoid making assumptions about how the optimal metaparameters vary as a function of batch size. Finally, in order to guide future work, we consider any remaining limitations in our methodology, and we discuss what we see as the most interesting unanswered questions that arise from our experiments.

\subsection{Scope}\label{sec:scope}

We restrict our attention to variants of mini-batch stochastic gradient descent (SGD), which are the dominant algorithms for training neural networks. These algorithms iteratively update the model's parameters using an estimate of the gradient of the training objective.
The gradient is estimated at each step using a different subset, or \vocab{(mini-) batch}, of training examples.
See Section~\ref{sec:setup-algos} for a more detailed description of these algorithms.
A data-parallel implementation computes gradients for different training examples in each batch in parallel, and so, in the context of mini-batch SGD and its variants, we equate the batch size with the amount of data parallelism.\footnote{Mini-batch SGD can be implemented in a variety of ways, including data-serially, but a data-parallel implementation is always possible given appropriate hardware.} We restrict our attention to synchronous SGD because of its popularity and advantages over asynchronous SGD \citep{chen2016revisiting}.

Practitioners are primarily concerned with out-of-sample error and the cost they pay to achieve that error. Cost can be measured in a variety of ways, including training time and hardware costs. Training time can be decomposed into number of steps multiplied by average time per step, and hardware cost into number of steps multiplied by average hardware cost per step. The per-step time and hardware costs depend on the practitioner's hardware, but the number of training steps is hardware-agnostic and can be used to compute the total costs for any hardware given its per-step costs.
Furthermore, in an idealized data-parallel system where the communication overhead between processors is negligible, training time depends only on the number of training steps (and not the batch size) because the time per step is independent of the number of examples processed. 
Indeed, this scenario is realistic today in systems like TPU pods\footnote{ \url{https://www.blog.google/products/google-cloud/google-cloud-offer-tpus-machine-learning/}.}, where there are a range of batch sizes for which the time per step is almost constant.
Since we are primarily concerned with training time, we focus on number of training steps as our main measure of training cost.

An alternative hardware-agnostic measure of training cost is the number of training examples processed, or equivalently the number of passes (\vocab{epochs}) over the training data. This measure is suitable when the per-step costs are proportional to the number of examples processed (e.g. hardware costs proportional to the number of floating point operations).
However, the number of epochs is not a suitable measure of training time in a data-parallel system---it is possible to reduce training time by using a larger batch size and processing \textit{more} epochs of training data, provided the number of training steps decreases.

In light of practitioners' primary concerns of out-of-sample error and the resources needed to achieve it, we believe the following questions are the most important to study to understand the costs and benefits of data parallelism with mini-batch SGD and its variants:
\begin{enumerate}
    \item What is the relationship between batch size and number of training steps to reach a goal out-of-sample error?
    \item What governs this relationship?
    \item Do large batch sizes incur a cost in out-of-sample error?
\end{enumerate}

\subsection{Contributions of This Work}

\begin{enumerate}
    \item We show that the relationship between batch size and
    number of training steps to reach a goal out-of-sample error has the same characteristic form across six different families of neural network, three training algorithms, and seven data sets.

    Specifically, for each workload (model, training algorithm, and data set), increasing the batch size initially decreases the required number of training steps proportionally, but eventually there are diminishing returns until finally increasing the batch size no longer changes the required number of training steps. 
    To the best of our knowledge, we are the first to experimentally validate this relationship across models, training algorithms, and data sets while independently tuning the learning rate, momentum, and learning rate schedule (where applicable) for each batch size.
    Unlike prior work that made strong assumptions about these metaparameters, our results reveal a universal relationship that holds across all workloads we considered, across different error goals, and when considering either training error or out-of-sample error. 
    
    \item We show that the maximum useful batch size varies significantly between workloads and depends on properties of the model, training algorithm, and data set. Specifically, we show that:
    \begin{enumerate}
        \item SGD with momentum (as well as Nesterov momentum) can make use of much larger batch sizes than plain SGD, suggesting future work to study the batch size scaling properties of other algorithms.
        \item Some models allow training to scale to much larger batch sizes than others. We include experimental data on the relationship between various model properties and the maximum useful batch size, demonstrating that the relationship is not as simple as one might hope from previous work (e.g. wider models do \textit{not} always scale better to larger batch sizes).
        \item The effect of the data set on the maximum useful batch size tends to be smaller than the effects of the model and training algorithm, and does not depend on data set size in a consistent way.
    \end{enumerate}
    
    \item We show that the optimal values of training metaparameters do not consistently follow any simple relationships with the batch size. In particular, popular learning rate heuristics---such as linearly scaling the learning rate with the batch size--- do not hold across all problems or across all batch sizes.
    
    \item Finally, by reviewing the specifics of the experimental protocols used in prior work, we at least partially reconcile conflicting stances in the literature on whether increasing the batch size degrades model quality.
    Specifically, we show that assumptions about computational budgets and the procedures for selecting metaparameters at different batch sizes can explain many of the disagreements in the literature. We find no evidence that increasing the batch size necessarily degrades model quality, but additional regularization techniques may become important at larger batch sizes. 

\end{enumerate}

\subsection{Experimental Data}

We release our raw experimental data for any further analysis by the research community.\footnote{\url{https://github.com/google-research/google-research/tree/master/batch_science}} Our database contains 454 combinations of workload (model, data set, training algorithm) and batch size, each of which is associated with a metaparameter search space and a set of models trained with different configurations sampled from the search space. In total, our data contains 71,638,836 loss measurements taken over the course of training for 168,160 individual models. Together, these measurements make up the training curves of all of the individual models we trained, and can be used to reproduce all plots in this paper.\footnote{\url{https://colab.research.google.com/github/google-research/google-research/blob/master/batch_science/reproduce_paper_plots.ipynb}}

\section{Setup and Background}\label{sec:setup}

In this section we set up the basic definitions and background concepts used throughout the paper.

\subsection{Learning}\label{sec:setup-learning}
A \vocab{data distribution} is a probability distribution $\mD$ over a data domain $\mZ$. For example, we might consider a supervised learning task over a domain $\mZ = \mX \times \mY$, where $\mX$ is the set of 32-by-32-pixel color images and $\mY$ is the set of possible labels denoting what appears in the image. A \vocab{training set} $z_1, \ldots, z_n \in \mZ$ is a collection of \vocab{examples} from the data domain, conventionally assumed to be drawn i.i.d.\ from the data distribution $\mD$. 

A machine learning \vocab{model} is a function that, given \vocab{parameters} $\theta$ from some set
$\Theta \subset \R^d$, and given a data point $z \in \mZ$, produces a prediction whose quality is measured by a differentiable non-negative scalar-valued loss function.\footnote{Technically, the loss need only be sub-differentiable. Extending our setup to this end is straightforward.} We denote by $\loss(\theta; z)$ the loss of a prediction made by the model, under parameters $\theta$, on the data point $z$. We denote by $\risk$ the \vocab{out-of-sample loss} or \vocab{expected loss}:
\begin{align}
    \risk(\theta) &\defeq \E_{z \sim \mD} \left[ \loss(\theta; z) \right],
    \label{eq:risk}
\end{align}
and by $\emprisk$ the \vocab{empirical average loss} under a data set $S = (z_1, \ldots, z_n)$:
\begin{align}
    \emprisk(\theta; S) &\defeq \frac 1 n \sum_{i=1}^n \loss(\theta; z_i).
    \label{eq:emprisk}
\end{align}
When $S$ is the training set, we call $\emprisk$ the \textit{average training loss}. We will say that the data source $\mD$, loss $\loss$, and model with parameter set $\Theta$ together specify a learning \vocab{task}, in which our aim is to find parameters $\theta$ that achieve low out-of-sample loss (Equation~\ref{eq:risk}), while given access only to $n$ training examples. A common approach is to find parameters of low average training loss (Equation~\ref{eq:emprisk}) as an estimate of the out-of-sample loss \citep{shalev2014understanding}.

When minimizing average training loss $\emprisk$, it is common to add regularization penalties to the objective function. For a differentiable penalty $R : \Theta \to \R_+$, regularization weight $\lambda > 0$, and training set $S$, the training objective might be
\begin{align}
    \objective(\theta) &\defeq \emprisk(\theta; S) + \lambda R(\theta).
    \label{eq:training-objective}
\end{align}

In practice, we often approach a task by replacing its loss with another that is more amenable to training. For instance, in supervised classification, we might be tasked with learning under the 0/1 loss, which is an indicator of whether a prediction is correct (e.g.\ matches a ground-truth label), but we train by considering instead a surrogate loss (e.g.\ the logistic loss) that is more amenable to continuous optimization. When the surrogate loss bounds the original, achieving low loss under the surrogate implies low loss under the original. To distinguish the two, we say \vocab{error} to describe the original loss (e.g.\ 0/1), and we save \vocab{loss} to refer to the surrogate used in training.

\subsection{Algorithms}
\label{sec:setup-algos}
The dominant algorithms for training neural networks are based on \vocab{mini-batch stochastic gradient descent} \citep[SGD,][]{robbins1951stochastic, kiefer1952stochastic, rumelhart1986learning, bottou2008tradeoffs, lecun2015deep}.
Given an initial point $\theta_0 \in \Theta$, mini-batch SGD attempts to decrease the objective $\objective$ via the sequence of iterates
\begin{align*}
  \theta_t &\gets \theta_{t-1} - \eta_t g(\theta_{t-1}; B_t),
\end{align*}
where each $B_t$ is a random subset of training examples, the sequence $\{\eta_t\}$ of positive scalars is called the \vocab{learning rate}, and where, for any $\theta \in \Theta$ and $B \subset S$,
\begin{align}
    g(\theta; B) &\defeq \frac 1 {|B|} \sum_{z \in B} \grad \ell(\theta; z) + \lambda \grad R(\theta) .
    \label{eq:batch-stoch-grad}
\end{align}
When the examples $B$ are a uniformly random subset of training examples, $g(\theta; B)$ forms an unbiased estimate of the gradient of the objective $\objective$ that we call a \vocab{stochastic gradient}.
In our larger-scale experiments, when we sample subsequent batches $B_t$, we actually follow the common practice of cycling through permutations of the training set \citep{shamir2016without}.
The result of mini-batch SGD can be any of the iterates $\theta_t$ for which we estimate that $\risk(\theta_t)$ is low using a validation data set.

Variants of SGD commonly used with neural networks include SGD with momentum \citep{polyak1964some, rumelhart1986learning, SutskeverEtAl_icml2013}, Nesterov momentum \citep{nesterov1983method, SutskeverEtAl_icml2013}, RMSProp \citep{hinton2012rmsprop}, and Adam \citep{kingma2014adam}. All of these optimization procedures, or \vocab{optimizers}, interact with the training examples only by repeatedly computing stochastic gradients (Equation~\ref{eq:batch-stoch-grad}), so they support the same notion of batch size that we equate with the scale of data parallelism. In this work, we focus on the SGD, SGD with momentum, and Nesterov momentum optimizers. The latter two optimizers are configured by a learning rate $\{\eta_t\}$ and a scalar $\gamma \in (0, 1)$ that we call \vocab{momentum}. They define the iterates\footnote{These rules take slightly different forms across the literature and 
across library implementations. We present and use the update rules from the \texttt{MomentumOptimizer} class in TensorFlow~\citep{abadi2016tensorflow}.}
\begin{align*}
    \text{\underline{SGD with momentum}} \span & \text{\underline{Nesterov momentum}} \span \\
    v_{t+1} &\gets \gamma v_t + g(\theta_t; B_t) & v_{t+1} &\gets \gamma v_t + g(\theta_t; B_t) \\
    \theta_{t+1} &\gets \theta_t - \eta_t v_{t+1} & \theta_{t+1} &\gets \theta_t - \eta_t g(\theta_t; B_t) - \eta_t \gamma v_{t+1},
\end{align*}
given $v_0 = 0$ and an initial $\theta_0$.
Note that plain SGD can be recovered from either optimizer by taking $\gamma = 0$. The outcome of using these optimizers should therefore be no worse if than SGD, in any experiment, the momentum $\gamma$ is tuned across values including zero.

If we run SGD with momentum under a constant learning rate $\eta_t = \eta$, then, at a given iteration $t$, the algorithm computes
\begin{align*}
    \theta_{t+1}
    &= \theta_t - \eta v_{t+1}
     = \theta_0 - \eta \sum_{u=0}^t v_{u+1}
     = \theta_0 - \eta \sum_{u=0}^t \sum_{s=0}^u \gamma^{u-s} g(\theta_s; B_s).
\end{align*}
For any fixed $\tau \in \{0, \ldots, t\}$, the coefficient accompanying the stochastic gradient $g(\theta_\tau; B_\tau)$ in the above update is $\eta \sum_{u=\tau}^t \gamma^{u-\tau}$. We define the \vocab{effective learning rate}, $\eta^\text{eff}$ as the value of this coefficient at the end of training ($t=T$), in the limit of a large number of training steps ($T \rightarrow \infty$, while $\tau$ is held fixed):
\begin{align*}
    \eta^\text{eff}
    &= \lim_{T \to \infty} 
    \sum_{u=\tau}^T \eta \gamma^{u-\tau}
     = \frac \eta {1-\gamma}.
\end{align*}
Put intuitively, $\eta^\text{eff}$ captures the contribution of a given mini-batch gradient to the parameter values at the end of training.

\subsection{Additional Terminology in Experiments}
A \vocab{data-parallel implementation} of mini-batch SGD (or one of its variants) computes the summands of Equation~\ref{eq:batch-stoch-grad} in parallel and then synchronizes to coordinate their summation.

The models and algorithms in our experiments are modifiable by what we call \vocab{metaparameters}.\footnote{Sometimes called ``hyperparameters,'' but we prefer a different name so as not to clash with the notion of hyperparameters in Bayesian statistics.}
These include architectural choices, such as the number of layers in a neural network, and training parameters, such as learning rates $\{\eta_t\}$ and regularization weights $\lambda$.
When we use the term \vocab{model}, we typically assume that all architectural metaparameters have been set. In our experiments, we \vocab{tune} the training metaparameters by selecting the values that yield the best performance on a validation set.
We use the term \vocab{workload} to jointly refer to a data set, model, and training algorithm.

\section{Related Work}

In this section we review prior work related to our three main questions from Section~\ref{sec:scope}. First we review studies that considered the relationship between batch size and number of training steps (Questions~1 and~2), and then we review studies that considered the effects of batch size on solution quality (Question~3).

\subsection{Steps to Reach a Desired Out-Of-Sample Error}\label{sec:related-stt}

We broadly categorize the related work on this topic as either analytical or empirical in nature.

\subsubsection{Analytical Studies}\label{sec:related-stt-theory}

Convergence upper bounds from the theory of stochastic (convex) optimization can be specialized to involve terms dependent on batch size, so in this sense they comprise basic related work. These upper bounds arise from worst-case analysis, and moreover make convexity and regularity assumptions that are technically violated in neural network training, so whether they predict the actual observed behavior of our experimental workloads is an empirical question in its own right.

Given a sequence of examples drawn i.i.d.\ from a data source, an upper bound on the performance of SGD applied to $L$-Lipschitz convex losses is \citep{hazan2016oco, shalev2014understanding}
\begin{align}
  \objective(\theta_T) - \opt{\objective} &\le O\left(
    \sqrt{\frac {L^2} T} \right),
  \label{eq:sgd-lipschitz}
\end{align}
for any batch size. Here, $\objective$ is the objective function, $\opt{\objective}$ is its value at the global optimum, and $\theta_T$ denotes the final output of the algorithm supposing it took $T$ iterations.\footnote{Not necessarily the $T$\nth\ iterate, which may differ from $\theta_T$ if the algorithm averages its iterates.}
Meanwhile, when losses are convex and the objective is $H$-smooth, accelerated parallel mini-batch SGD enjoys the bound \citep{lan2012optimal}
\begin{align}
  \objective(\theta_T) - \opt{\objective} &\le O\left(
    \frac H {T^2} + \sqrt{\frac {L^2} {Tb}} \right),
\label{eq:asgd-smooth}
\end{align}
where $b$ is the batch size.

Compared to sequential processing without batching (i.e.\ a batch size of one), the bounds Equation~\ref{eq:sgd-lipschitz} and Equation~\ref{eq:asgd-smooth} offer two extremes, respectively:
\begin{enumerate}
    \item \textbf{No benefit:} Increasing the batch size $b$ does not change the number of steps to convergence, as per Equation~\ref{eq:sgd-lipschitz}.
    
    \item \textbf{A $b$-fold benefit:} The term in Equation~\ref{eq:asgd-smooth} proportional to $1/\sqrt{Tb}$ dominates the bound. Increasing the batch size $b$ by a multiplicative factor decreases the number of steps $T$ to a given achievable objective value by the same factor.
\end{enumerate}
In other words, under these simplifications, batching cannot hurt the asymptotic guarantees of steps to convergence, but it could be wasteful of examples.
The two extremes imply radically different guidance for practitioners, so the critical task of establishing a relationship between batch size and number of training steps remains one to resolve experimentally.

A few recent papers proposed analytical notions of a critical batch size:
a point at which a transition occurs from a $b$-fold benefit to no benefit.
Under assumptions including convexity, \citet{ma2017power} derived such a critical batch size, and argued that a batch size of one is optimal for minimizing the number of training epochs required to reach a given target error.
Under different assumptions, \citet{yin2017gradient} established a critical batch size and a pathological loss function that together exhibit a transition from a $b$-fold benefit to no benefit. Although they ran experiments with neural networks, their experiments were designed to investigate the effect of data redundancy and do not provide enough information to reveal the empirical relationship between batch size and number of training steps.
Focusing on linear least-squares regression, \citet{jain2018parallelizing} also derived a threshold batch size in terms of the operator norm of the objective's Hessian and a constant from a fourth-moment bound on example inputs.

To our knowledge, in all previous work that analytically derived a critical batch size, the thresholds defined are either (i) parameter-dependent, or (ii) specific to linear least-squares regression.
A critical batch size that depends on model parameters can change over the course of optimization; it is not a problem-wide threshold that can be estimated efficiently \emph{a priori}.
Focusing on least-squares has issues as well: while it sheds intuitive light on how batching affects stochastic optimization locally, the quantities defined inherently cannot generalize to the non-linear optimization setting of neural network training, both because the objective's Hessian is not constant across the space of parameters as it is in a quadratic problem, and more broadly because it is unclear whether the Hessian of the objective is still the correct analogue to consider.

\subsubsection{Empirical Studies}\label{sec:related-stt-empirical}

\citet{wilson2003general} investigated the relationship between batch size and training speed for plain mini-batch SGD. They found
that a simple fully connected neural network took more epochs to converge with larger batch sizes on a data set of 20,000 examples, and also that using a batch size equal to the size of the training set took more epochs to converge than a batch size of one on several small data sets of size $\le 600$.
However, their experimental protocol and assumptions limit the conclusions we can draw from their results.
One issue is that training time was measured to different out-of-sample errors for different batch sizes on the same data set.
To compare training speed fairly, the error goal should be fixed across all training runs being compared. Additionally, only four learning rates were tried for each data set, but quite often the best learning rate was at one of the two extremes and it appeared that a better learning rate might be found outside of the four possibilities allowed.
Finally, despite the contention of the authors, their results do not imply slower training with larger batch sizes in data-parallel training: for the most part, their larger batch size experiments took fewer training steps than the corresponding batch size one experiments.

In the last few years, increasingly specialized computing systems have spurred practitioners to try much larger batch sizes than ever before,
while increasingly promising results have driven hardware designers to create systems capable of even more data parallelism.
\citet{chen2016revisiting} used a pool of synchronized worker machines
to increase the effective batch size of mini-batch SGD. They demonstrated speedups in both wall time and steps to convergence for an Inception model \citep{szegedy2016rethinking} on ImageNet \citep{russakovsky2015imagenet} by scaling the effective batch size from 1,600 to 6,400. More recently, \citet{goyal2017accurate} showed that the number of training epochs could be held constant across a range of batch sizes to achieve the same validation error for ResNet-50 \citep{he2016deep} on ImageNet. Holding the number of training epochs constant is equivalent to scaling the number of training steps inversely with the batch size, and this reduction in training steps with increasing batch size produced nearly proportional wall time speedups on their hardware. Although this hints at a $b$-fold benefit regime in which increasing the batch size reduces the number of training steps by the same factor, the authors did not attempt to minimize the number of training steps (or epochs) required to reach the goal at each batch size separately. It is unclear whether any of the batch sizes that achieved the goal could do so in fewer steps than given, or how many steps the other batch sizes would have needed to achieve the same error goal.

Two studies performed concurrently with this work also investigated the relationship between batch size and training speed for neural networks. \citet{chen2018effect} provide experimental evidence of a problem-dependent critical batch size after which a $b$-fold benefit is no longer achieved for plain mini-batch SGD. They contend that wider and shallower networks have larger critical batch sizes, and while their empirical results are equivocal for this particular claim, they show that the threshold batch size can depend on aspects of both the data set and the model. Additionally, \citet{golmant2018computational} studied how three previously proposed heuristics for adjusting the learning rate as a function of batch size (linear scaling, square root scaling, and no scaling) affect the number of training steps required to reach a particular result. They found that if the learning rate is tuned for the the smallest batch size only, all three of these common scaling techniques break down for larger batch sizes and result in either (i) divergent training, or (ii) training that cannot reach the error goal within a fixed number of training epochs. They also describe a basic relationship between batch size and training steps to a fixed error goal, which is comprised of three regions: $b$-fold benefit initially, then diminishing returns, and finally no benefit for all batch sizes greater than a maximum useful batch size. However, their results are inconclusive because (i) not all model and data set pairs exhibit this basic relationship, (ii) it does not appear consistently across error goals, and (iii) the relationship is primarily evident in training error but not out-of-sample error.
These inconsistent results may be due to suboptimal pre-determined learning rates arising from the scaling rules, especially at larger batch sizes.
Finally, they also found that the maximum useful batch size depends on aspects of the model and the data set type, but not on the data set size. Since all their experiments use plain mini-batch SGD, their results are unable to reveal any effects from the choice of optimizer and might not generalize to other popular optimizers, such as SGD with momentum.

\subsection{Solution Quality}\label{sec:related-solution-quality}

The literature contains some seemingly conflicting claims about the effects of batch size on solution quality (out-of-sample error at the conclusion of training). Primarily, the debate centers on whether increasing the batch size incurs a cost in solution quality.
\citet{keskar2016large} argue that large batch\footnote{The term ``large batch'' is inherently ambiguous, and in this case accompanies experiments in \citet{keskar2016large} that only compare two absolute batch sizes per data set, rather than charting out a curve to its apparent extremes.} training converges to so-called ``sharp'' minima with worse generalization properties. However, \citet{DinhPBB17} show that a minimum with favorable generalization properties can be made, through reparameterization, arbitrarily sharp in the same sense.
\citet{lecun-98x} suggest that a batch size of one can result in better solutions because the noisier updates allow for the possibility of escaping from local minima in a descent algorithm. However, they also note that we usually stop training long before reaching any sort of critical point. \citet{hoffer2017train} argue that increasing the batch size need not degrade out-of-sample error at all, assuming training has gone on long enough. \citet{goyal2017accurate}, among others, tested batch sizes larger than those used in \citet{keskar2016large} without noticing any reduction in solution quality. Still, their results with yet larger batch sizes do not rule out the existence of a more sudden degradation once the batch size is large enough. Meanwhile, \citet{GoodfellowEtAlBook2016} state that small batches can provide a regularization effect such that they result in the best observed out-of-sample error, although in this case other regularization techniques might serve equally well.

Alas, the best possible out-of-sample error for a particular model and data set cannot be measured unconditionally due to practical limits on wall time and hardware resources, as well as practical limits on our ability to tune optimization metaparameters (e.g. the learning rate). An empirical study can only hope to measure solution quality subject to the budgets allowed for each model experiment, potentially with caveats due to limitations of the specific procedures for selecting the metaparameters. To the best of our knowledge, all published results handle the training budget issue in exactly one of three ways: by ignoring budgets (train to convergence, which is not always possible); by using a step budget (restrict the number of gradient descent updates performed); or by using an epoch budget (restrict number of training examples processed).\footnote{Of course, there are budgets in between an epoch budget and a step budget that might allow the possibility of trading off time, computation, and/or solution quality. For example, it may be possible to increase the number of training epochs and still take fewer steps to reach the same quality solution. However, we are not aware of work that emphasizes these budgets.}
Furthermore, while some published results tune the learning rate anew for each batch size, others tune for only a single batch size and use a preordained heuristic to set the learning rate for the remaining batch sizes (the most common heuristics are constant, square root, and linear learning rate scaling rules).
Tuning metaparameters at a single batch size and then heuristically adjusting them for others could clearly create a systematic advantage for trials at batch sizes near to the one tuned. All in all, the conclusions we can draw from previous studies depend on the budgets they assume and on how they select metaparameters across batch sizes. The following subsections attempt an investigation of their experimental procedures to this end.

\subsubsection{Studies That Ignore Budgets}

All studies in this section compared solution quality for different batch sizes after deeming their models to have converged. They determined training stopping time by using either manual inspection, convergence heuristics, or fixed compute budgets that they considered large enough to guarantee convergence.\footnote{As discussed further in Section \ref{sec:solution-quality}, we find that millions of training steps for small batch sizes, or thousands of epochs for large batch sizes, are required to saturate performance even for data sets as small and simple as MNIST. In our experiments, this corresponded to more than 25 hours of wall-time for each metaparameter configuration.}

\citet{keskar2016large} trained several neural network architectures on MNIST and CIFAR-10, each with two batch sizes, using the Adam optimizer and without changing the learning rate between batch sizes. They found that the larger batch size consistently achieved worse out-of-sample error after training error had ceased to improve.
However, all models used batch normalization \citep{ioffe2015batch} and presumably computed the batch normalization statistics using the full batch size. For a fair comparison between batch sizes, batch normalization statistics should be computed over the same number of examples or else the training objective differs between batch sizes \citep{goyal2017accurate}. Indeed, \citet{hoffer2017train} found that computing batch normalization statistics over larger batches can degrade solution quality, which suggests an alternative explanation for the results of \citet{keskar2016large}. Moreover, \citet{keskar2016large} reported that data augmentation eliminated the difference in solution quality between small and large batch experiments.

\citet{smith2018bayesian} trained a small neural network on just 1,000 examples sampled from MNIST with two different batch sizes, using SGD with momentum and without changing the learning rate between batch sizes. They observed that the larger batch size overfit more than the small batch size resulting in worse out-of-sample error, but this gap was mitigated by applying $L_2$~regularization \citep[][figures 3 and 8]{smith2018bayesian}. They also compared a wider range of batch sizes in experiments that either (i) used a step budget without changing the learning rate for each batch size \citep[][figures 4 and 6]{smith2018bayesian}, or (ii) varied the learning rate and used a step budget that was a function of the learning rate \citep[][figure 5]{smith2018bayesian}.
Instead, we focus on the case where the learning rate and batch size are chosen independently.

\citet{breuel2015benchmarking,breuel2015effects} trained a variety of neural network architectures on MNIST with a range of batch sizes, using the SGD and SGD with momentum optimizers with a range of learning rates and momentum values. They found that batch size had no effect on solution quality for LSTM networks \citep{breuel2015benchmarking}, but found that larger batch sizes achieved worse solutions for fully connected and convolutional networks, and that the scale of the effect depended on the activation function in the hidden and output layers \citep{breuel2015effects}.

Finally, \citet{chen2016revisiting} observed no difference in solution quality when scaling the batch size from 1,600 to 6,400 for an Inception model on ImageNet when using the RMSProp optimizer and a heuristic to set the learning rate for each batch size.

\subsubsection{Studies with Step Budgets}

\citet{hoffer2017train} trained neural networks with two different batch sizes on several image data sets. They found that, by computing batch normalization statistics over a fixed number of examples per iteration (``ghost batch normalization''), and by scaling the learning rate with the square root of the batch size instead of some other heuristic, the solution quality arising from the larger batch size was as good as or better than the smaller batch size. However, the largest batch size used was 4,096, which does not rule out an effect appearing at still larger batch sizes, as suggested by the work of \citet{goyal2017accurate}. Moreover, it remains open whether their proposed learning rate heuristic extends to arbitrarily large batch sizes, or whether it eventually breaks down for batch sizes sufficiently far from the base batch size.

\subsubsection{Studies with Epoch Budgets}

An epoch budget corresponds to fixing the total number of per-example gradient computations, but, in an idealized data-parallel implementation of SGD, it also corresponds to a step (or even wall time) budget that scales inversely with the batch size. With an epoch budget, a larger batch size can only achieve the same solution quality as a smaller batch size if it achieves perfect scaling efficiency (a $b$-fold reduction in steps from increasing the batch size, as described in Section \ref{sec:related-stt-theory}).

\citet{masters2018revisiting} show that after a critical batch size depending on the model and data set, solution quality degrades with increasing batch size \emph{when using a fixed epoch budget}.
Their results effectively show a limited region of $b$-fold benefit for those model and data set pairs when trained with SGD, although they did not investigate whether this critical batch size depends on the optimizer used, and they did not consider more than one epoch budget for each problem. We reproduced a subset of their experiments and discuss them in Section~\ref{sec:discussion}.

\citet{goyal2017accurate} recently popularized a linear learning rate scaling heuristic for training the ResNet-50 model using different batch sizes. Using this heuristic, a 90 epoch budget, and SGD with momentum without adjusting or tuning the momentum, they increased the batch size from 64 to 8,192 with no loss in accuracy. However, their learning rate heuristic broke down for even larger batch sizes. Inspired by these results, a sequence of follow-up studies applied additional techniques to further increase the batch size while still achieving the same accuracy and using the same 90 epoch budget. These follow-on studies \citep{CodreanuEtAl2017,you2017imagenet,AkibaEtAl2017} confirm that the best solution quality for a given batch size will also depend on the exact optimization techniques used.

There are several additional papers \citep{lin2018don,devarakonda2017adabatch,golmant2018computational} with experiments relevant to solution quality that used an epoch budget, tuned the learning rate for the smallest batch size, and then used a heuristic to choose the learning rate for all larger batch sizes.
For instance, \citet{devarakonda2017adabatch} and \citet{lin2018don} used linear learning rate scaling and \citet{golmant2018computational} tried constant, square root, and linear learning rate scaling heuristics.
All of them concluded that small batch sizes have superior solution quality to large batch sizes with a fixed epoch budget, for various notions of ``small'' and ``large.'' This could just as easily be an artifact of the learning rate heuristics, and a possible alternative conclusion is that these heuristics are limited (as heuristics often are).

\section{Experiments and Results}
\label{sec:methods_and_experiments}

The primary quantity we measure is the number of steps needed to first reach a desired out-of-sample error, or \textit{steps to result}. 
To measure steps to result, we used seven image and text data sets with training set sizes ranging from 45,000 to 26 billion examples. Table~\ref{table:datasets} summarizes these data sets and Appendix~\ref{appendix:datasets} provides the full details. We chose six families of neural network to train on these data sets. For MNIST and Fashion MNIST, we chose a simple fully connected neural network and a simple convolutional neural network (CNN). For CIFAR-10, we chose the ResNet-8 model without batch normalization, partly to compare our results to \citet{masters2018revisiting}, and partly to have a version of ResNet without batch normalization. For ImageNet, we chose ResNet-50, which uses batch normalization and residual connections, and VGG-11, which uses neither. For Open Images, we chose ResNet-50. For LM1B, we chose the Transformer model and an LSTM model. For Common Crawl, we chose the Transformer model. Table~\ref{table:models} summarizes these models and Appendix~\ref{appendix:models} provides the full details.

\begin{table}[t]
\begin{center}
\begin{small}
\begin{tabular}{ |l|l|l|l|l|l| }
\hline
\textbf{Data Set} & \textbf{Type} & \textbf{Task} & \textbf{Size} & \textbf{Evaluation Metric} \\
\hline
MNIST & Image & Classification & 55,000 & Classification error \\  
\hline
Fashion MNIST & Image & Classification & 55,000 & Classification error \\ 
\hline
CIFAR-10 & Image & Classification & 45,000 & Classification error \\ 
\hline
ImageNet & Image & Classification & 1,281,167 & Classification error \\ 
\hline
Open Images & Image & Classification (multi-label) & 4,526,492 & Average precision \\ 
\hline
LM1B & Text & Language modeling & 30,301,028 & Cross entropy error \\
\hline
Common Crawl & Text & Language modeling & ${\sim}25.8$ billion & Cross entropy error \\ 
\hline
\end{tabular}
\end{small}

\caption{Summary of data sets. Size refers to the number of examples in the training set, which we measure in sentences for text data sets. See Appendix~\ref{appendix:datasets} for full details.}
\label{table:datasets}

\end{center}
\end{table}

\begin{table}[t]
\begin{center}
\begin{small}
\begin{tabular}{ |l|l|l|l|l| }
\hline
\textbf{Model Class}  & \textbf{Sizes} & \textbf{Optimizers} & \textbf{Data Sets} & \textbf{Learning rate}\\
                &               &                   &               & \textbf{schedule}  \\
\hline
Fully Connected & Various       & SGD               & MNIST         & Constant      \\
\hline
Simple CNN      & Base          & SGD               & MNIST         & Constant      \\
                & Narrow        & Momentum          & Fashion MNIST &               \\
                & Wide          & Nesterov mom. &               &               \\
\hline 
ResNet          & ResNet-8      & SGD               & CIFAR-10      & Linear decay  \\
                &               & Nesterov mom. &               &               \\
\cline{2-5}
                & ResNet-50     & Nesterov mom. & ImageNet      & Linear decay  \\
                &               &                   & Open Images   &               \\
\hline
VGG             & VGG-11        & Nesterov mom. & ImageNet      & Linear decay  \\
\hline
Transformer     & Base          & SGD               & LM1B          & Constant      \\
                & Narrow and shallow & Momentum     & Common crawl  &               \\
                & Shallow       & Nesterov mom. &               &               \\
                & Wide          &                   &               &               \\
\hline
LSTM            & ---           & Nesterov mom. & LM1B          & Constant      \\
\hline
\end{tabular}
\end{small}

\caption{Summary of models. See Appendix~\ref{appendix:models} for full details.}
\label{table:models}

\end{center}
\end{table}

Measuring steps to result requires a particular value of out-of-sample error to be chosen as the goal. Ideally, we would select the best achievable error for each task and model, but since validation error is noisy, the best error is sometimes obtained unreliably. Moreover, for some workloads, the validation error continues to improve steadily beyond the maximum practical training time. Therefore, we generally tried to select the best validation error that we could achieve reliably within a practical training time.

Table~\ref{table:models} also shows the learning rate schedule we used for each model and data set. Learning rate schedules are often used to accelerate neural network training, but finding the best schedule is an optimization problem in its own right \citep{wu2018understanding}. Instead, researchers typically choose from a range of common learning rate functions based on validation performance and individual preference.
While most schedules decay the learning rate monotonically over training, some researchers also ``warm-up'' the learning rate at the start of training \citep[e.g.][]{he2016deep}, particularly when training with large batch sizes \citep{goyal2017accurate}.
We ran experiments with both constant learning rates and with learning rate decay.
We used decay for ResNet-8, ResNet-50, and VGG-11, which significantly reduced training time for those models.
We selected our decay function by running an extensive set of experiments with ResNet-50 on ImageNet (see Appendix~\ref{appendix:lr_decay} for details). We chose linear decay because it performed at least as well as all other schedules we tried, while also being the simplest and requiring only two additional metaparameters. In experiments that used linear decay, we specified metaparameters $(\eta_0, \alpha, T)$ such that the learning rate decayed linearly from $\eta_0$ to $\eta_T = \alpha \eta_0$. That is, the learning rate at step $t$ is given by
\begin{equation*}
    \eta_t = \begin{cases}
      \eta_0 - (1- \alpha) \eta_0 \frac{t}{T} & \text{if } t \le T, \\
      \alpha \eta_0 & \text{if } t > T.
    \end{cases}
\end{equation*}

Steps to result depends on the training metaparameters, and, for a given task and model, each batch size might have a different metaparameter configuration that minimizes steps to result.
In all experiments, we independently tuned the metaparameters at each batch size, including the initial learning rate $\eta_0$ and, when learning rate decay was used, the decay schedule ($\alpha, T$). Also, unless otherwise specified, we used the Nesterov momentum optimizer \citep{SutskeverEtAl_icml2013} and tuned the momentum $\gamma$.\footnote{For LSTM for LM1B, we used a fixed value of $\gamma=0.99$. We chose this value based on initial experiments and validated that tuning $\gamma$ did not significantly affect the results for batch sizes 256, 1,024, or 4,096.}
Tuning anew for each batch size is extremely important since otherwise we would not be measuring steps to result as a function of batch size, rather we would be measuring steps to result as a function of batch size and the specific values of the learning rate and other metaparameters.
We used quasi-random search \citep{BousquetEtAl_LDS_2017} to tune the metaparameters with equal budgets of non-divergent\footnote{We discarded trials with a divergent training loss, which occurred when the learning rate was too high.} trials for different batch sizes. We selected metaparameter search spaces by hand based on preliminary experiments. The exact number of non-divergent trials needed to produce stable results depends on the search space, but 100 trials seemed to suffice in our experiments.\footnote{We used 100 non-divergent trials for all experiments except Transformer Shallow on LM1B with SGD, Transformer on Common Crawl, and LSTM on LM1B, for which we used 50 trials each.
} If the optimal trial occurred near the boundary of the search space, or if the goal validation error was not achieved within the search space, we repeated the search with a new search space. We measured steps to result for each batch size by selecting the metaparameter trial that reached the goal validation error in the fewest number of steps.

\subsection{Steps to Result Depends on Batch Size in a Similar Way Across Problems}\label{sec:stt-basic-relationship}

To get a sense of the basic empirical relationship, we measured the number of steps required to reach a goal validation error as a function of batch size across several different data sets and models (Figure~\ref{fig:stt-problems}). In all cases, as the batch size grows, there is an initial period of \textbf{perfect scaling} ($b$-fold benefit, indicated with a dashed line on the plots) where the steps needed to achieve the error goal halves for each doubling of the batch size. However, for all problems, this is followed by a region of \textbf{diminishing returns} that eventually leads to a regime of \textbf{maximal data parallelism} where additional parallelism provides no benefit whatsoever. In other words, for any given problem and without making strong assumptions about learning rates or other optimizer parameters, we can achieve both extremes suggested by theory (see Section~\ref{sec:related-stt-theory}).
\textit{A priori}, it is not obvious that every workload in our experiments should exhibit perfect scaling at the smallest batch sizes instead of immediately showing diminishing returns. 

\newcommand{\threecolfigwidth}{0.32\textwidth}
\newcommand{\capshift}{-6mm}
\newcommand{\lineshift}{3mm}
\begin{figure}
    \centering
    \begin{subfigure}[b]{\threecolfigwidth}
        \includegraphics[width=\textwidth]{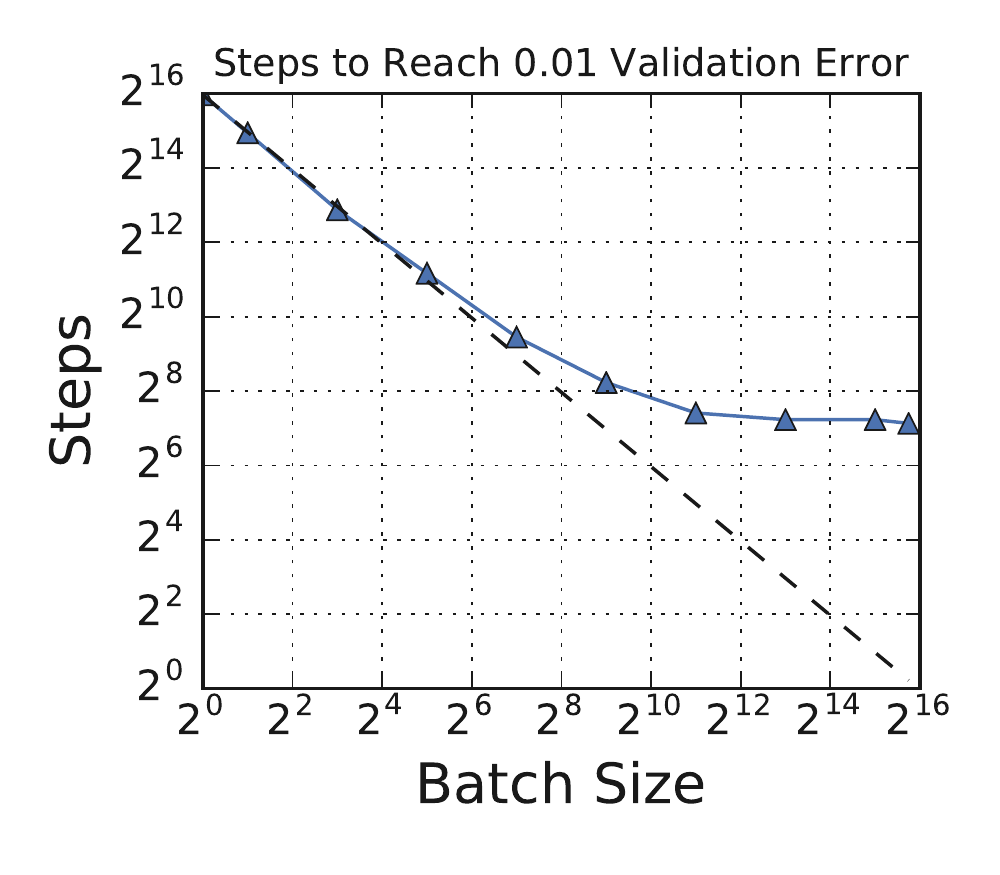}
        \vspace*{\capshift}
        \caption{Simple CNN on MNIST}
        \label{fig:stt-problems-cnn-mnist}
    \end{subfigure}
    \begin{subfigure}[b]{\threecolfigwidth}
        \includegraphics[width=\textwidth]{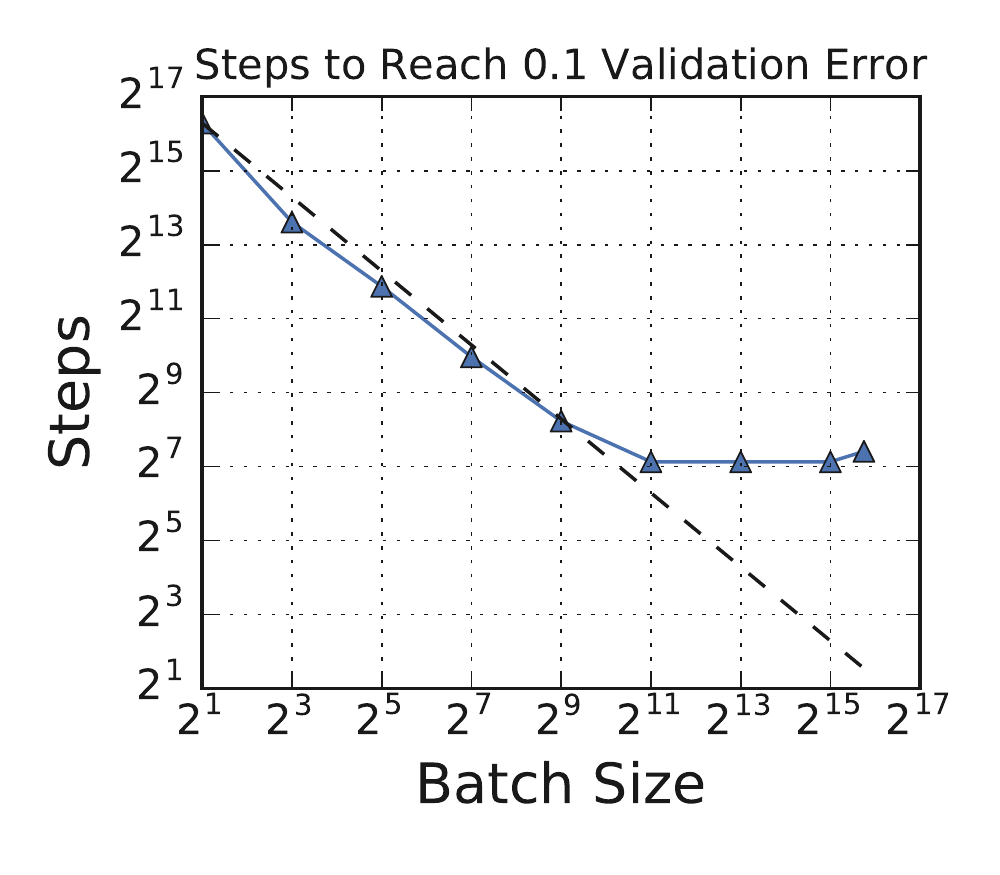}
        \vspace*{\capshift}
        \caption{Simple CNN on Fashion MNIST}
        \label{fig:stt-problems-cnn-fmnist}
    \end{subfigure}
    \begin{subfigure}[b]{\threecolfigwidth}
        \includegraphics[width=\textwidth]{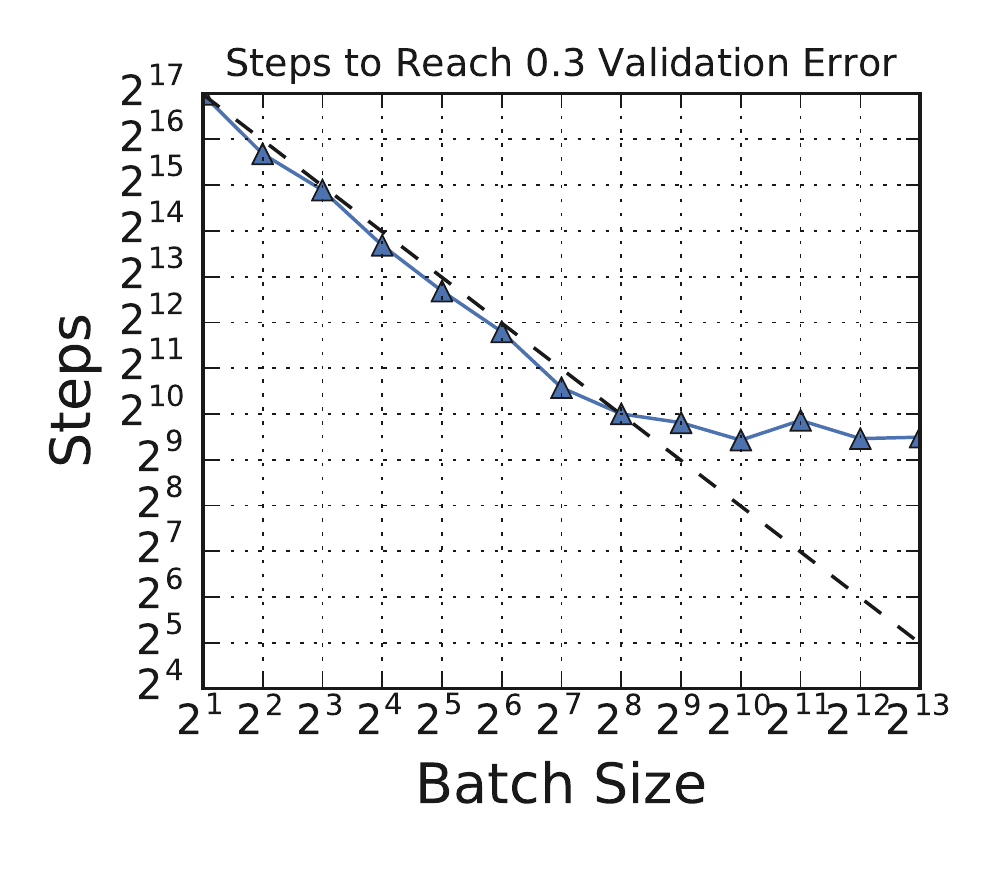}
        \vspace*{\capshift}
        \caption{ ResNet-8 on CIFAR-10}
        \label{fig:stt-problems-resnet-cifar}
    \end{subfigure} \\
    \vspace*{\lineshift}
    \begin{subfigure}[b]{\threecolfigwidth}
        \includegraphics[width=\textwidth]{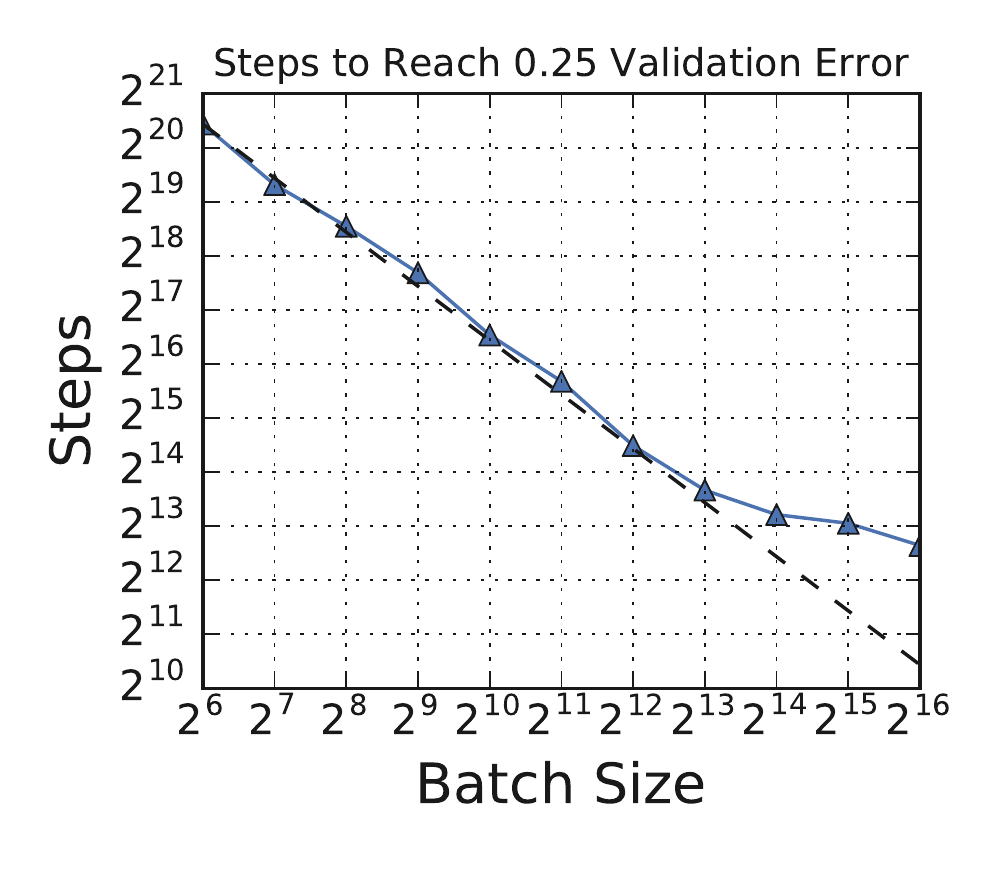}
        \vspace*{\capshift}
        \caption{ResNet-50 on ImageNet}
        \label{fig:stt-problems-resnet-imagenet}
    \end{subfigure}
    \begin{subfigure}[b]{\threecolfigwidth}
        \includegraphics[width=\textwidth]{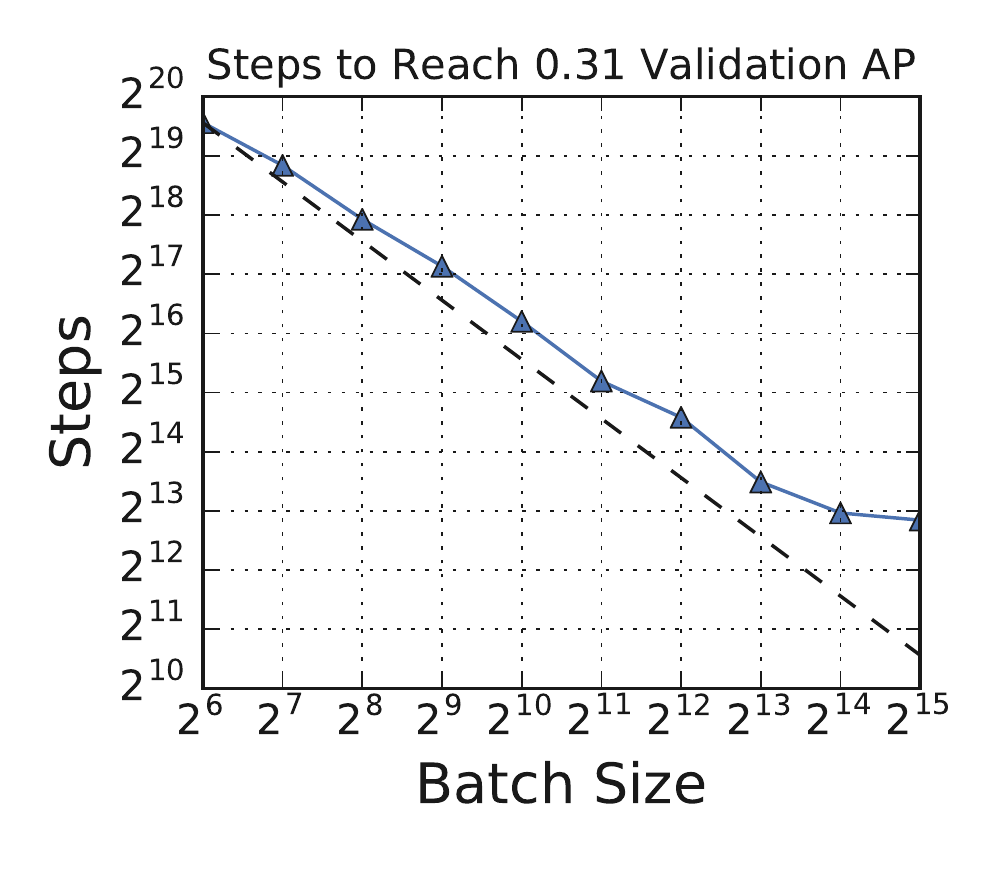}
        \vspace*{\capshift}
        \caption{ ResNet-50 on Open Images}
        \label{fig:stt-problems-resnet-oi}
    \end{subfigure}
    \begin{subfigure}[b]{\threecolfigwidth}
        \includegraphics[width=\textwidth]{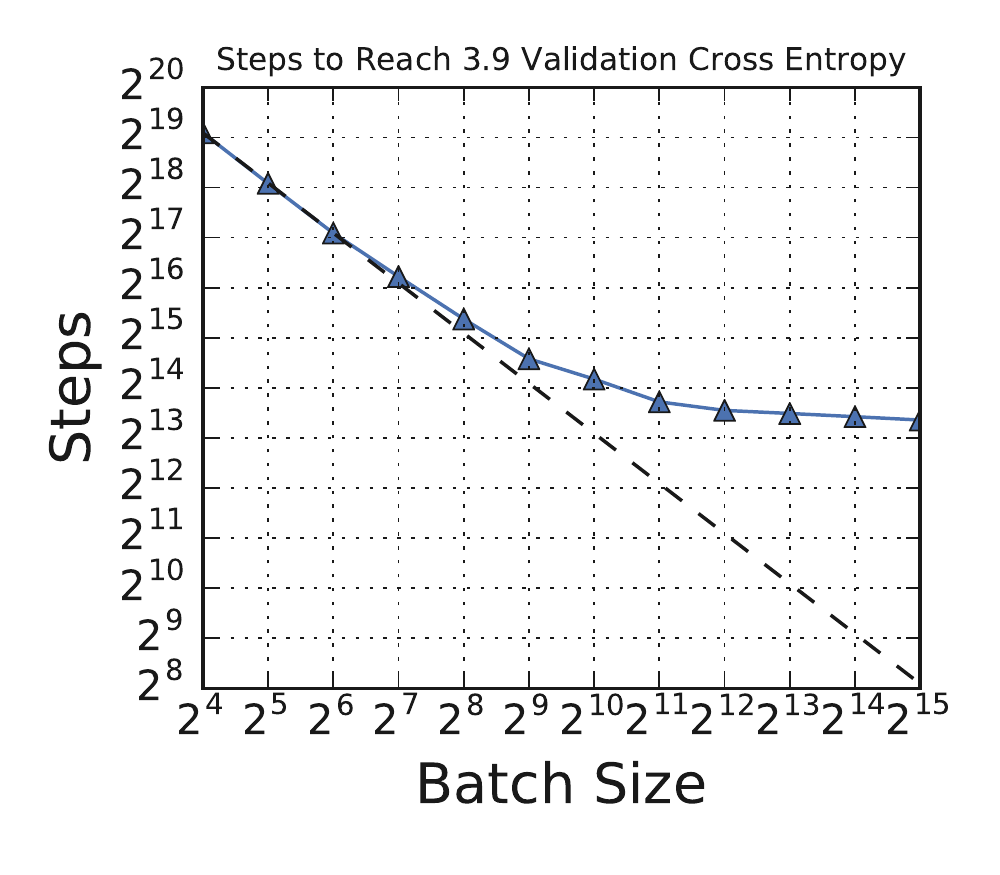}
        \vspace*{\capshift}
        \caption{ Transformer on LM1B}
        \label{fig:stt-problems-transformer-lm1b}
    \end{subfigure}\\
    \vspace*{\lineshift}
    \begin{subfigure}[b]{\threecolfigwidth}
        \includegraphics[width=\textwidth]{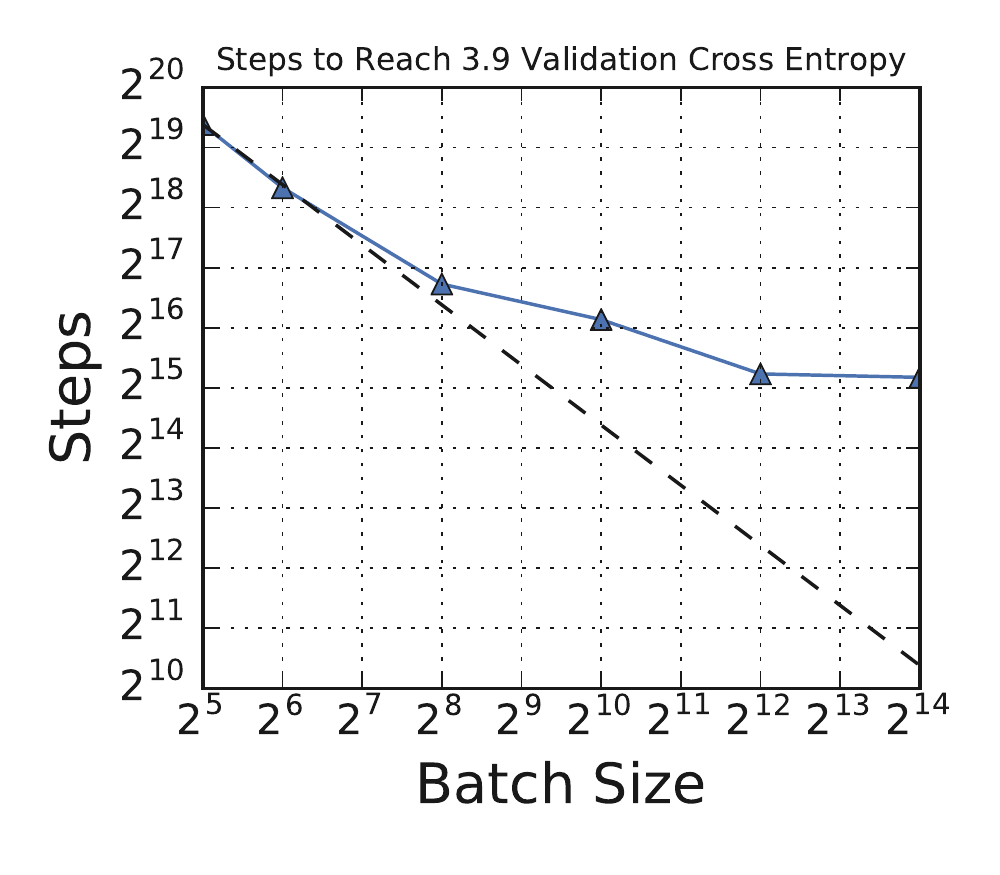}
        \vspace*{\capshift}
        \caption{Transformer on Common Crawl}
        \label{fig:stt-problems-transformer-cc}
    \end{subfigure}
    \begin{subfigure}[b]{\threecolfigwidth}
        \includegraphics[width=\textwidth]{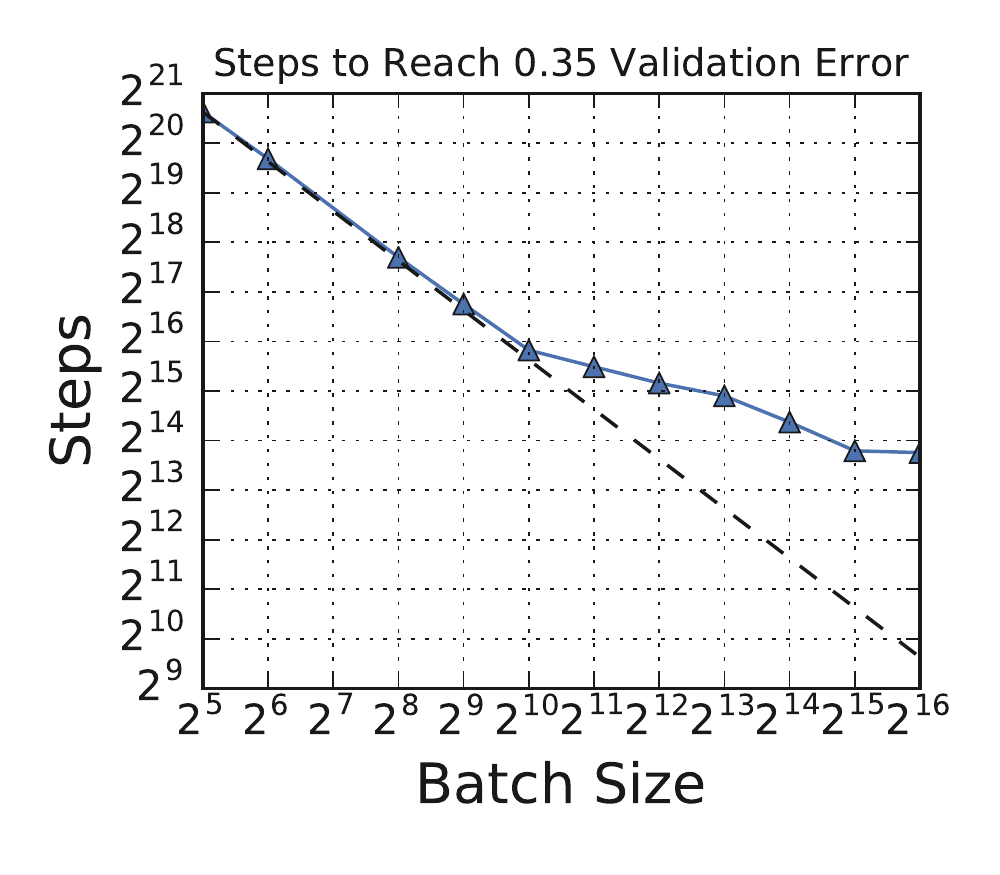}
        \vspace*{\capshift}
        \caption{ VGG-11 on ImageNet}
        \label{fig:stt-problems-vgg11-imagenet}
    \end{subfigure}
        \begin{subfigure}[b]{\threecolfigwidth}
        \includegraphics[width=\textwidth]{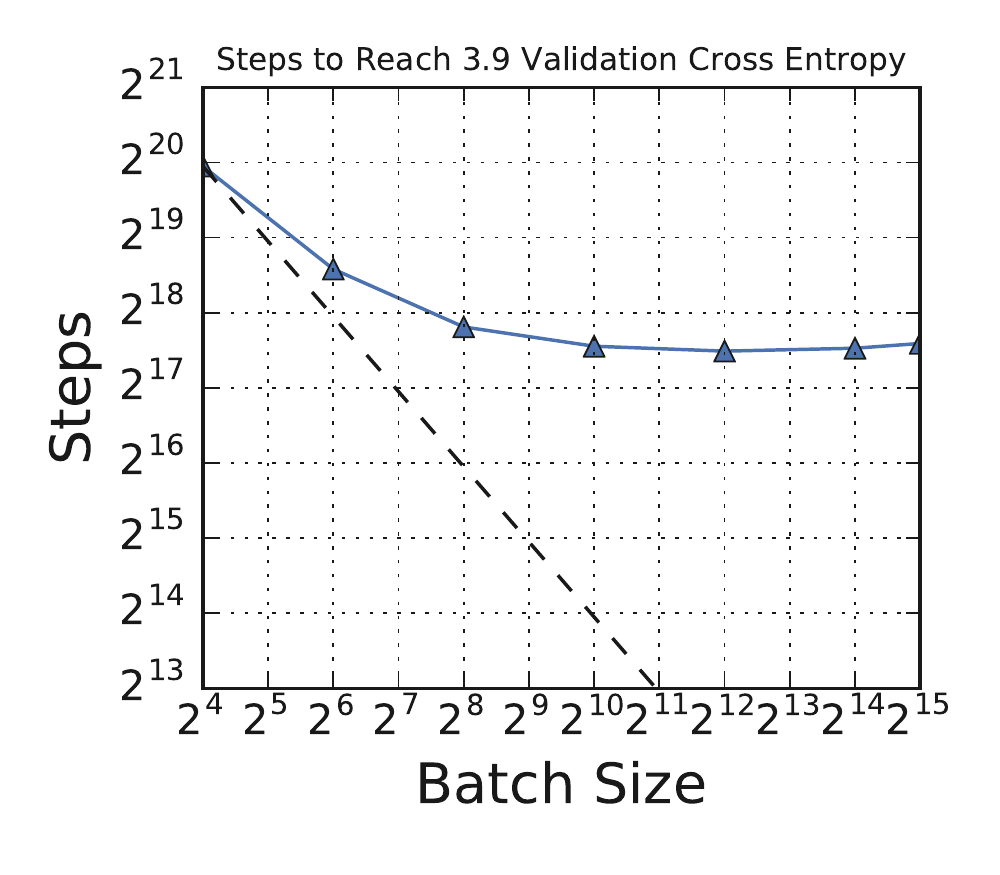}
        \vspace*{\capshift}
        \caption{LSTM on LM1B}
        \label{fig:stt-problems-lstm-lm1b}
    \end{subfigure}
    \caption{
    \textbf{The relationship between steps to result and batch size has the same characteristic form for all problems.} In all cases, as the batch size grows, there is an initial period of \textbf{perfect scaling} (indicated with a dashed line) where the steps needed to achieve the error goal halves for each doubling of the batch size. Then there is a region of \textbf{diminishing returns} that eventually leads to a region of \textbf{maximal data parallelism} where additional parallelism provides no benefit whatsoever. AP denotes average precision (see Appendix~\ref{appendix:datasets}).}
    \label{fig:stt-problems}
\end{figure}

\subsection{Validating Our Measurement Protocol}\label{sec:validating-protocol}

\begin{figure}
    \centering
    \begin{subfigure}[b]{\threecolfigwidth}
        \includegraphics[width=\textwidth]{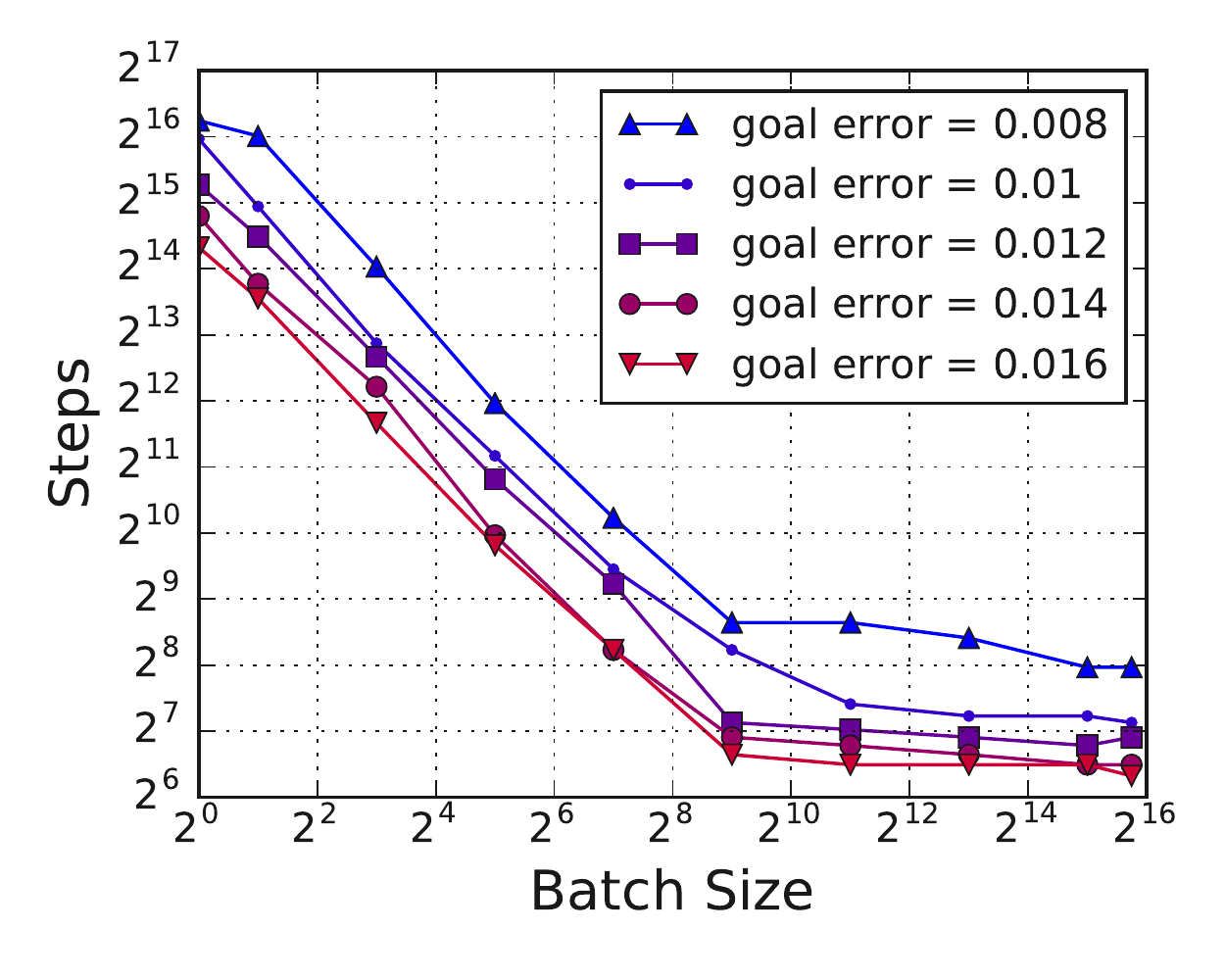}
        \vspace*{\capshift}
        \caption{Simple CNN on MNIST}
        \label{fig:cnn-mnist-multiple}
    \end{subfigure}    
    \begin{subfigure}[b]{\threecolfigwidth}
        \includegraphics[width=\textwidth]{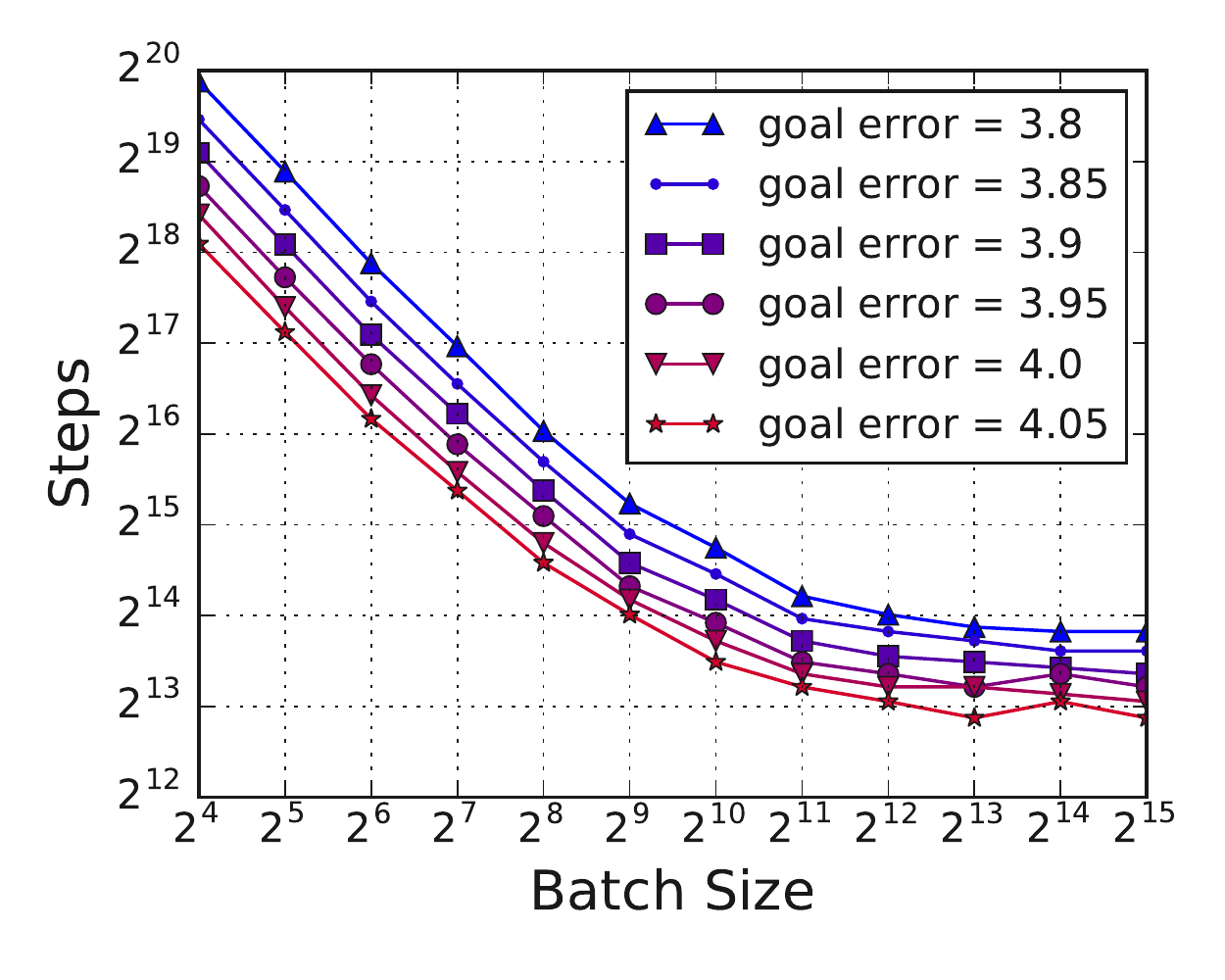}
        \vspace*{\capshift}
        \caption{Transformer on LM1B}
        \label{fig:transformer-lm1b-multiple}
    \end{subfigure}
    \begin{subfigure}[b]{\threecolfigwidth}
        \includegraphics[width=\textwidth]{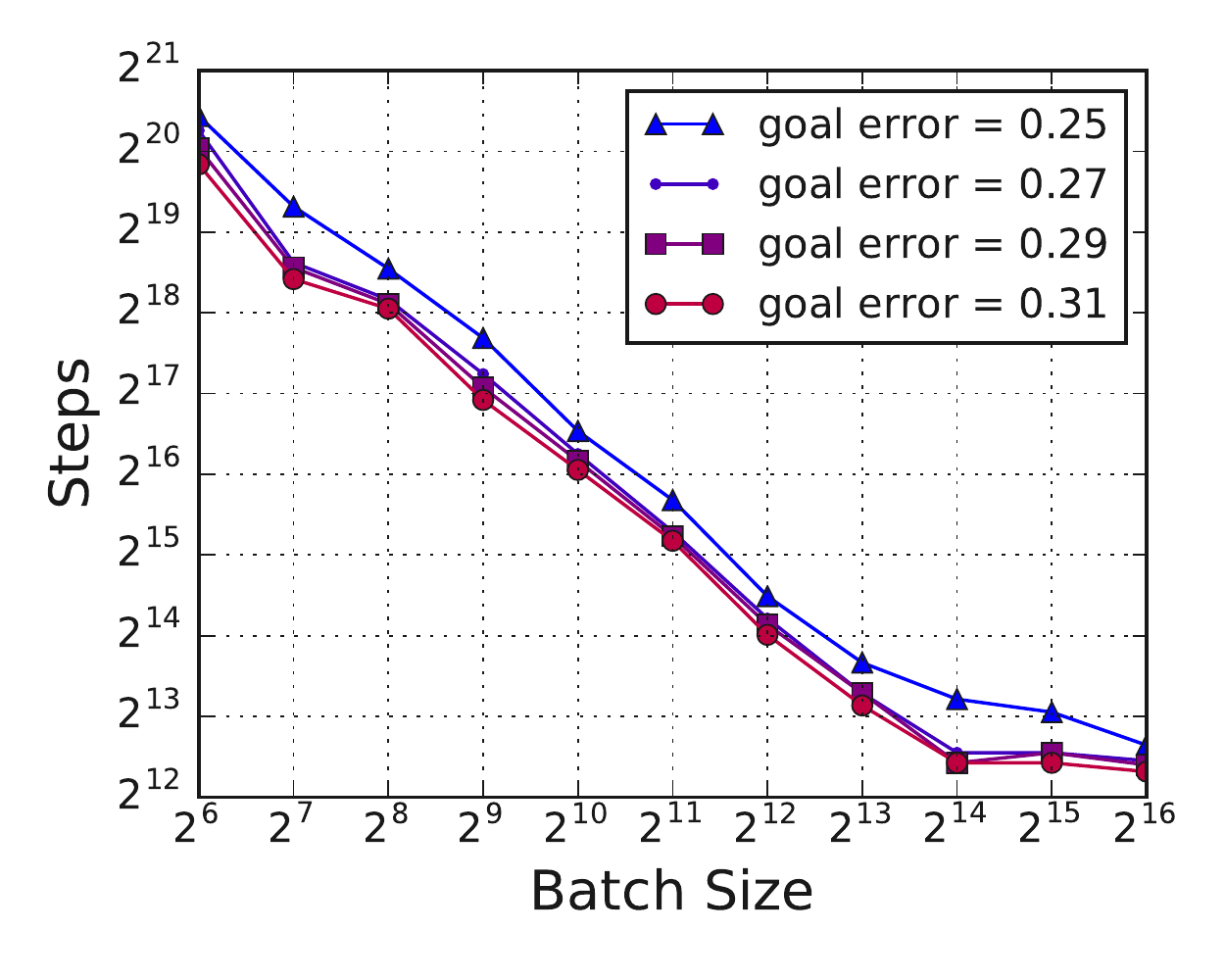}
        \vspace*{\capshift}
        \caption{ResNet-50 on ImageNet}
        \label{fig:resnet-multiple}
    \end{subfigure}
    \caption{
    \textbf{Steps-to-result plots have a similar form for different (nearby) performance goals.} The transition points between the three regions (perfect scaling, diminishing returns, and maximal data parallelism) are nearly the same.}
    \label{fig:multi_target}
\end{figure}

If the curves in Figure~\ref{fig:stt-problems} were very sensitive to the goal validation error, then measuring the steps needed to reach our particular choice of the goal would not be a meaningful proxy for training speed. For small changes in the goal validation error, we do not care about vertical shifts as long as the transition points between the three scaling regions remain relatively unchanged. Figure~\ref{fig:multi_target} shows that varying the error goal only vertically shifts the steps-to-result curve, at least for modest variations centered around a good absolute validation error. Furthermore, although we ultimately care about out-of-sample error, if our plots looked very different when measuring the steps needed to reach a particular \emph{training} error, then we would need to include both curves when presenting our results. However, switching to training error does not change the plots much at all (see Figure~\ref{fig:stt-train-valid} in the Appendix).

Our experiments depend on extensive metaparameter tuning for the learning rate, momentum, and, where applicable, the learning rate schedule. For each experiment, we verified our metaparameter search space by checking that the optimal trial was not too close to a boundary of the space. See Figures~\ref{fig:hparams-validation-transformer} and~\ref{fig:hparams-validation-resnet} in the Appendix for examples of how we verified our search spaces. 

\subsection{Some Models Can Exploit Much Larger Batch Sizes Than Others}\label{sec:stt-different-models}

\begin{figure}
    \centering
    \begin{subfigure}[b]{0.445\textwidth}
        \includegraphics[width=\textwidth]{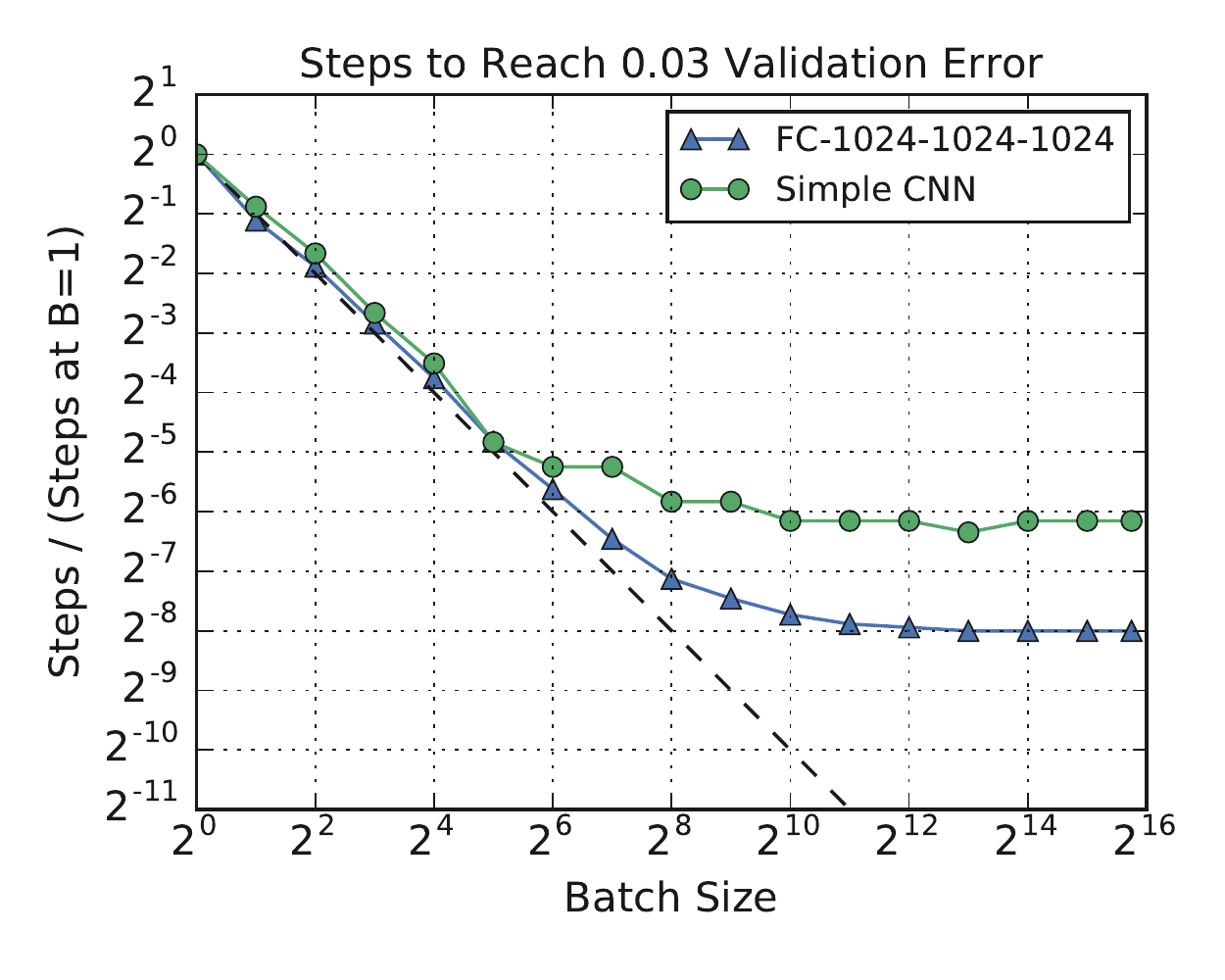}
        \vspace*{\capshift}
        \caption{Fully Connected vs Simple CNN on MNIST}
        \label{fig:stt-multiple-models-mnist-fc-cnn}
    \end{subfigure}
    \begin{subfigure}[b]{0.445\textwidth}
        \includegraphics[width=\textwidth]{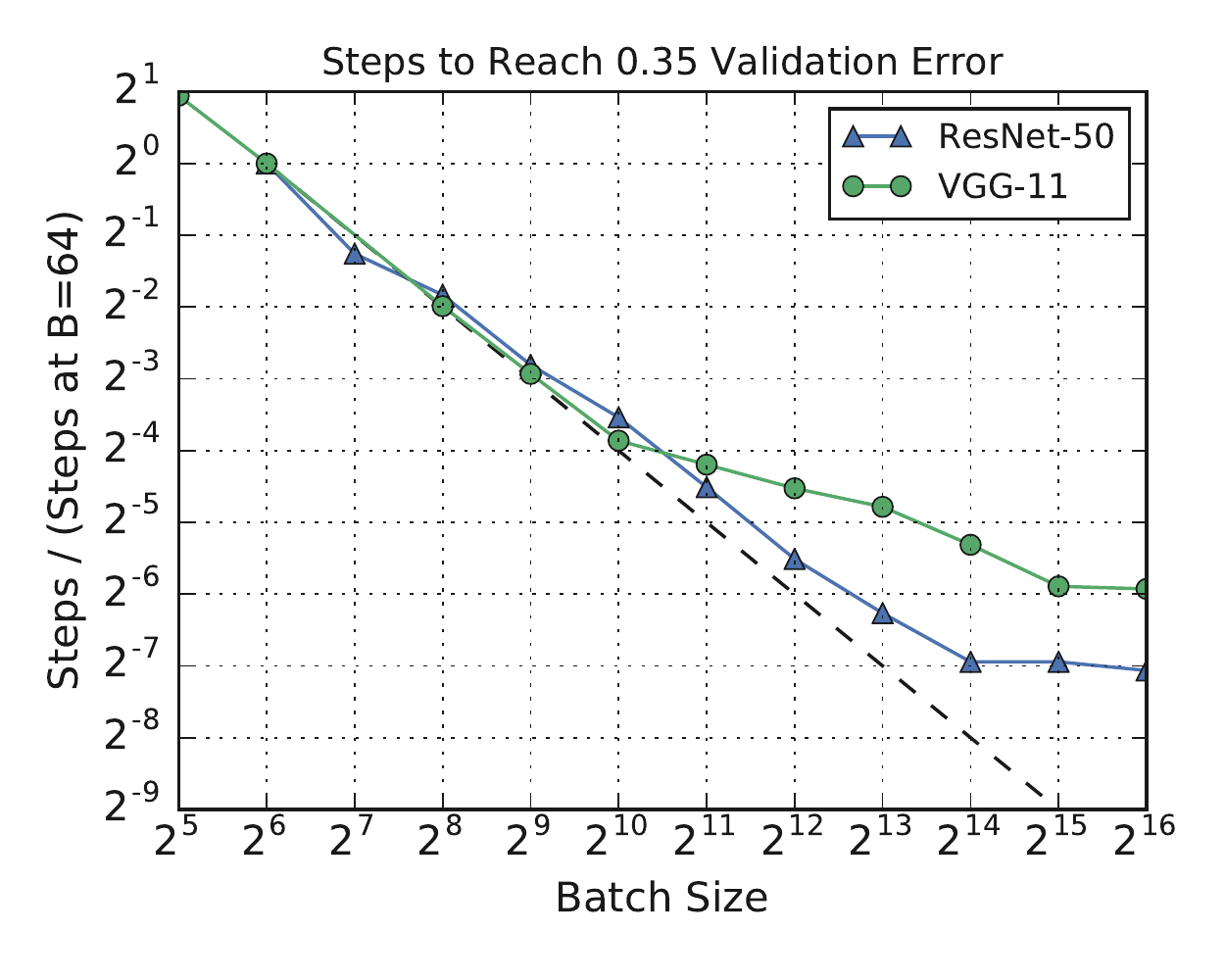}
        \vspace*{\capshift}
        \caption{ResNet-50 vs VGG-11 on ImageNet}
        \label{fig:stt-multiple-models-imagenet}
    \end{subfigure}\\
    \vspace*{\lineshift}
    \begin{subfigure}[b]{0.445\textwidth}
        \includegraphics[width=\textwidth]{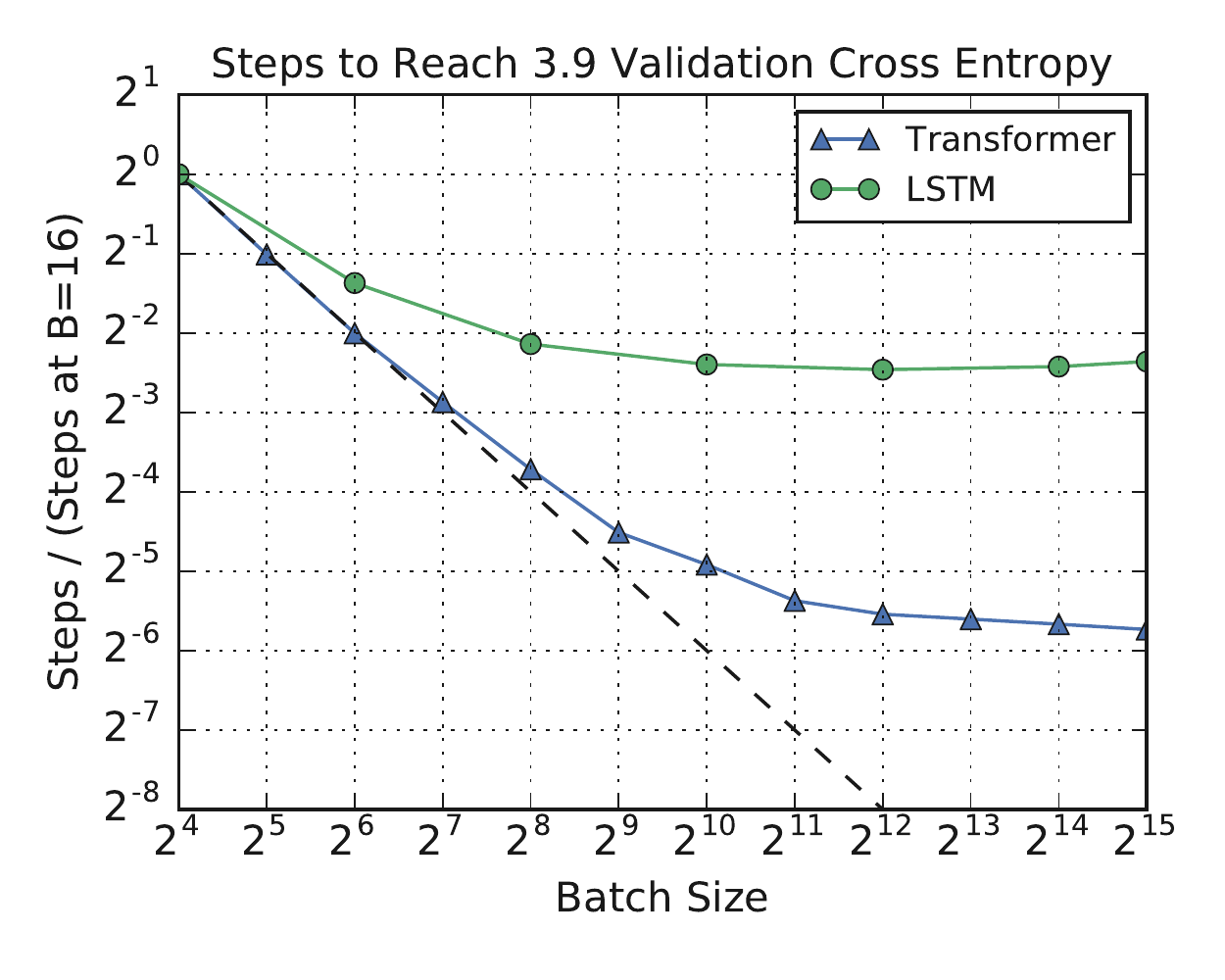}
        \vspace*{\capshift}
        \caption{Transformer vs LSTM on LM1B}
        \label{fig:stt-multiple-models-lm1b}
    \end{subfigure}
    \begin{subfigure}[b]{0.445\textwidth}
        \includegraphics[width=\textwidth]{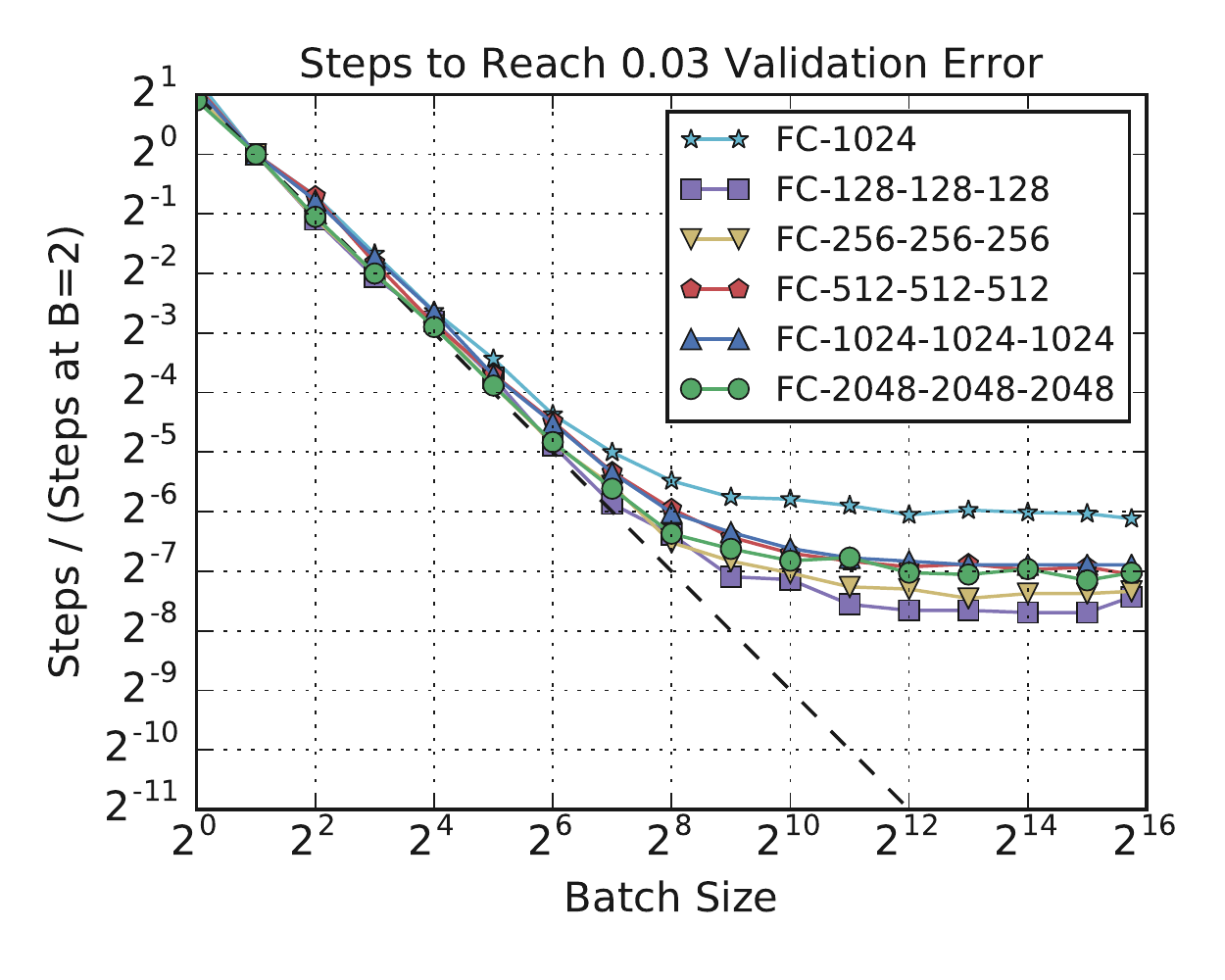}
        \vspace*{\capshift}
        \caption{Fully Connected sizes on MNIST}
        \label{fig:stt-multiple-models-fc}
    \end{subfigure}\\
    \vspace*{\lineshift}
    \begin{subfigure}[b]{0.445\textwidth}
        \includegraphics[width=\textwidth]{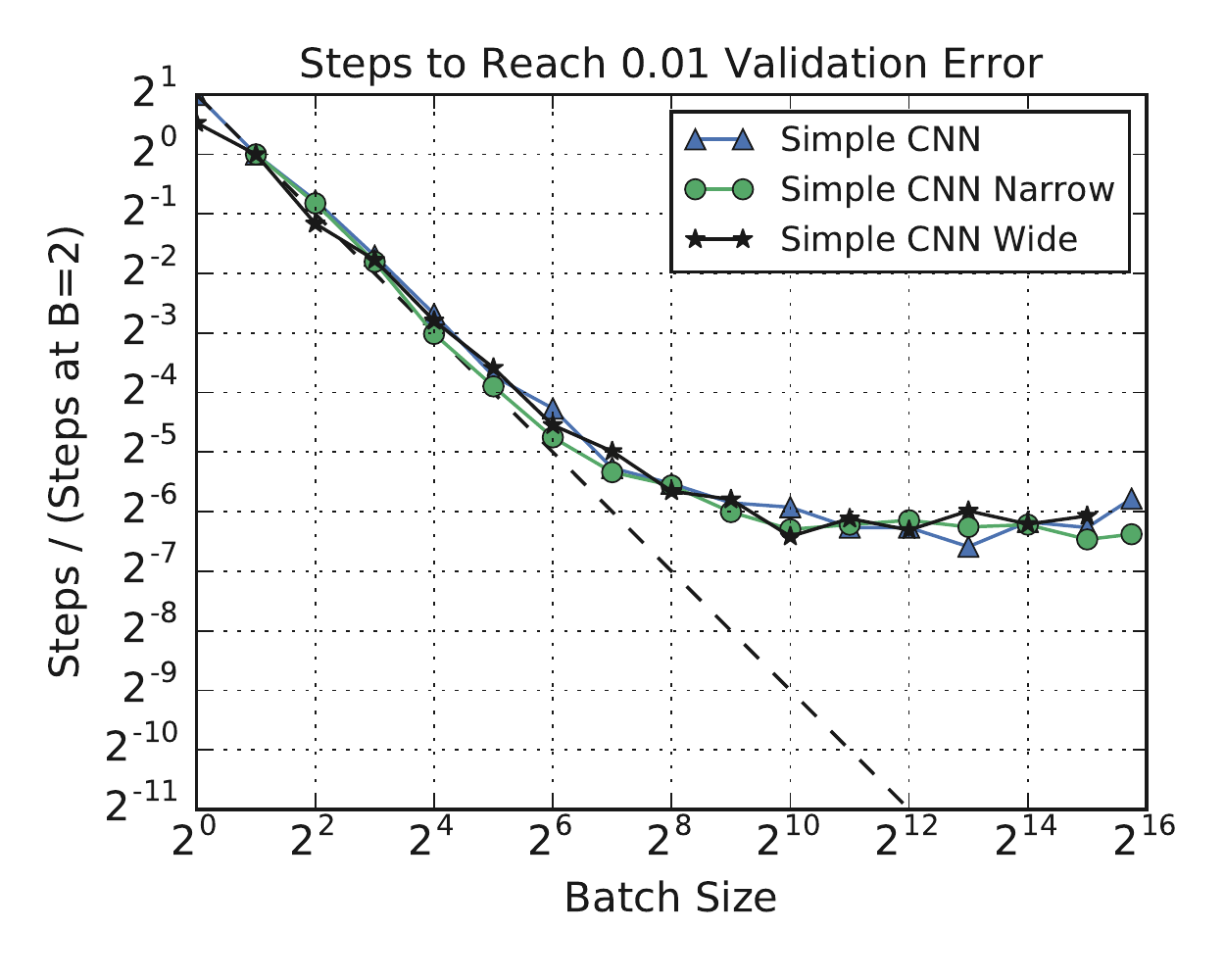}
        \vspace*{\capshift}
        \caption{Simple CNN sizes on MNIST}
        \label{fig:stt-multiple-models-cnn}
    \end{subfigure}
    \begin{subfigure}[b]{0.445\textwidth}
        \includegraphics[width=\textwidth]{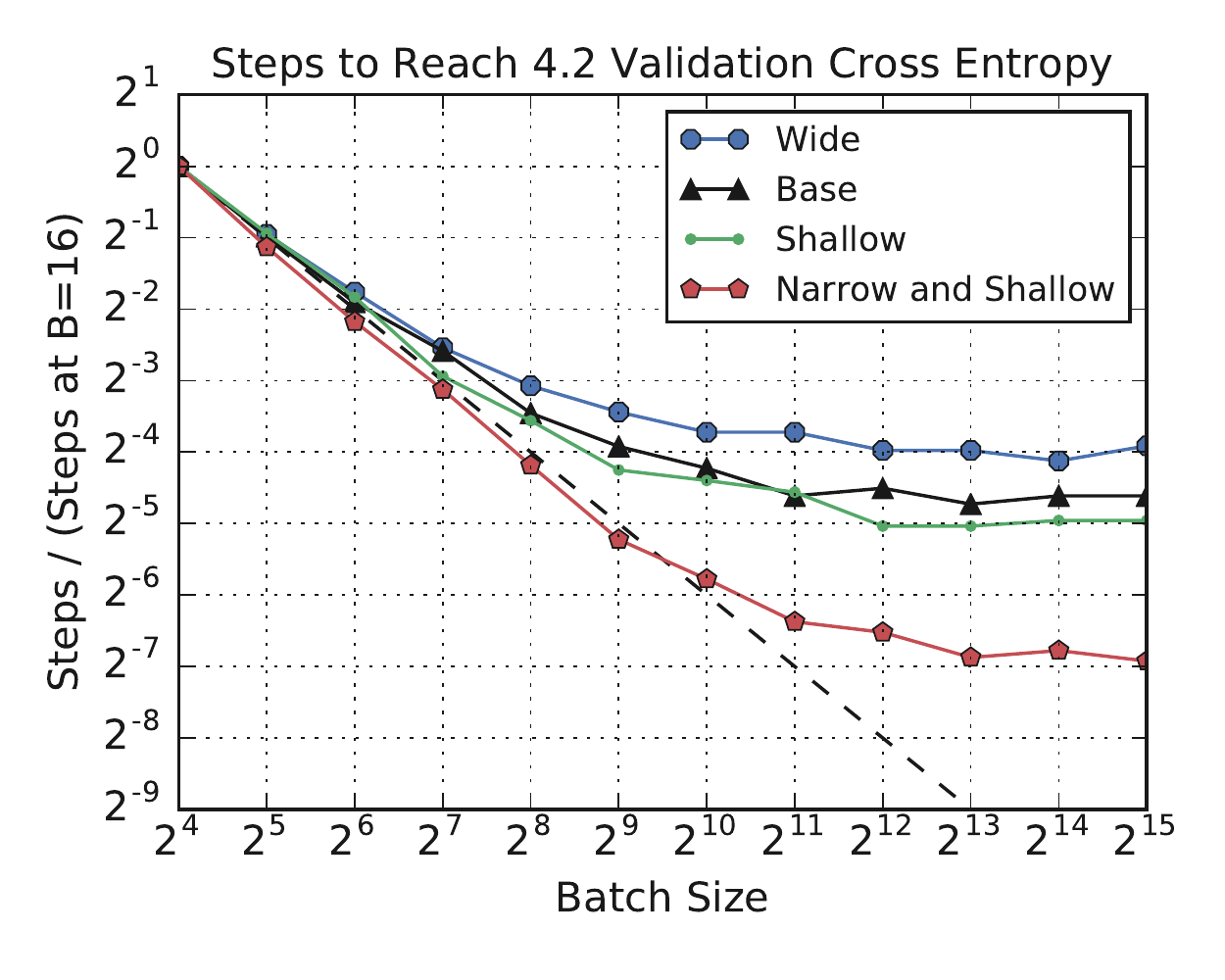}
        \vspace*{\capshift}
        \caption{Transformer sizes on LM1B}
        \label{fig:stt-multiple-model-sizes-lm1b}
    \end{subfigure}
    \caption{\textbf{Some models can exploit much larger batch sizes than others.}
    Figures~\ref{fig:stt-multiple-models-mnist-fc-cnn}-\ref{fig:stt-multiple-models-lm1b} show that some model architectures can exploit much larger batch sizes than others on the same data set.
    Figures~\ref{fig:stt-multiple-models-fc}-\ref{fig:stt-multiple-model-sizes-lm1b} show that varying the depth and width can affect a model's ability to exploit larger batches, but not necessarily in a consistent way across different model architectures.
    All MNIST models in this Figure used plain mini-batch SGD, while all other models used Nesterov momentum.
    The goal validation error for each plot was chosen to allow all model variants to achieve that error. Figure~\ref{fig:stt-multiple-models-raw} in the Appendix contains these plots without the $y$-axis normalized.}
    \label{fig:stt-multiple-models}
\end{figure}

We investigated whether some models can make more use of larger batches than others by experimenting with different models while keeping the data set and optimizer fixed. We explored this question in two ways: (i) by testing completely different model architectures on the same data set, and (ii) by varying the size (width and depth) of a model within a particular model family. Since the absolute number of steps needed to reach a goal validation error depends on the model, the steps to result vs.\ batch size curves for each model generally appear at different vertical offsets from each other. Since we primarily care about the locations of the perfect scaling, diminishing returns, and maximal data parallelism regions, we normalized the $y$-axis of each plot by dividing by the number of steps needed to reach the goal for a particular batch size and data set. This normalization corresponds to a vertical shift of each curve (on log-scale plots), and makes it easier to compare different models. Appendix~\ref{appendix:additional-plots} contains all plots in this section without the $y$-axis normalized.

Figures~\ref{fig:stt-multiple-models-mnist-fc-cnn}--\ref{fig:stt-multiple-models-lm1b} show that the model architecture significantly affects the relationship between batch size and the number of steps needed to reach a goal validation error. In Figure~\ref{fig:stt-multiple-models-mnist-fc-cnn}, the curve for the Fully Connected model flattens later than for the Simple CNN model on MNIST (although in this case the Simple CNN model can ultimately achieve better performance than the Fully Connected model). In Figure~\ref{fig:stt-multiple-models-imagenet}, the curve for ResNet-50 flattens much later than the curve for VGG-11, indicating that ResNet-50 can make better use of large batch sizes on this data set. Unlike ResNet-50, VGG-11 does not use batch normalization or residual connections. Figure~\ref{fig:stt-multiple-models-lm1b} shows that Transformer can make better use of large batch sizes than LSTM on LM1B.

Figures~\ref{fig:stt-multiple-models-fc}--\ref{fig:stt-multiple-model-sizes-lm1b} show that varying the depth and width can affect a model's ability to exploit larger batches, but not necessarily in a consistent way across different model architectures. In Figure~\ref{fig:stt-multiple-models-fc}, the regions of perfect scaling, diminishing returns, and maximum useful batch size do not change much when the width is varied for the Fully Connected model on MNIST, although the shallower model seems less able to exploit larger batches than the deeper models. This contrasts with the findings of \citet{chen2018effect}, although they changed width and depth simultaneously while keeping the number of parameters fixed. For Simple CNN on MNIST, the relationship between batch size and steps to a goal validation error seems not to depend on width at all (Figure~\ref{fig:stt-multiple-models-cnn-raw} in the Appendix shows that the curves are the same even when the $y$-axis is not normalized). However, in Figure~\ref{fig:stt-multiple-model-sizes-lm1b}, the curves for \textit{narrower} Transformer models on LM1B flatten later than for wider Transformer models, while the depth seems to have less of an effect. Thus, reducing width appears to allow Transformer to make more use of larger batch sizes on LM1B.

\subsection{Momentum Extends Perfect Scaling to Larger Batch Sizes, but Matches Plain SGD at Small 
Batch Sizes}

We investigated whether some optimizers can make better use of larger batches than others by experimenting with plain SGD, SGD with momentum, and Nesterov momentum on the same model and data set. Since plain SGD is a special case of both Nesterov momentum and SGD with momentum (with $\gamma = 0$ in each case), and since we tune $\gamma$ in all experiments, we expect that experiments with either of these optimizers should do no worse than plain SGD at any batch size. However, it is not clear \textit{a priori} whether momentum 
optimizers should outperform SGD, either by taking fewer training steps
or by extending the perfect scaling region to larger batch sizes.

\begin{figure}
    \centering
    \begin{subfigure}[b]{\threecolfigwidth}
        \includegraphics[width=\textwidth]{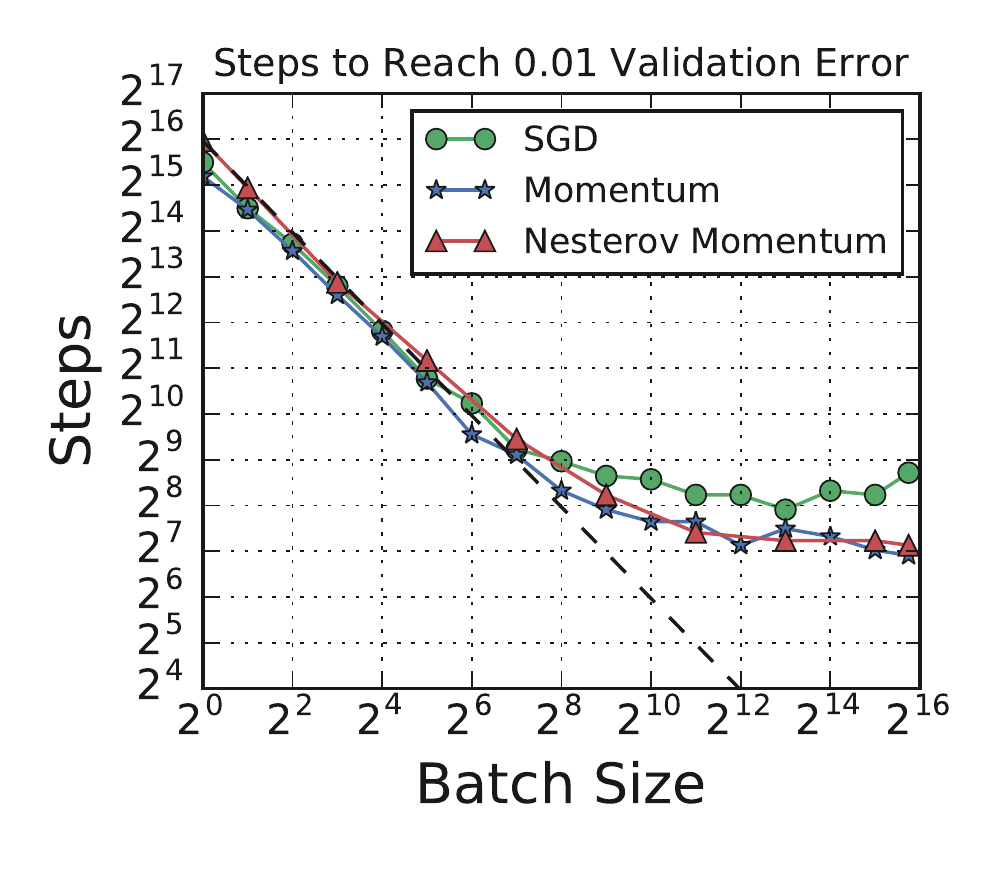}
        \vspace*{\capshift}
        \caption{Simple CNN on MNIST}
        \label{fig:stt-optimizers-cnn-mnist}
    \end{subfigure}    
    \begin{subfigure}[b]{\threecolfigwidth}
        \includegraphics[width=\textwidth]{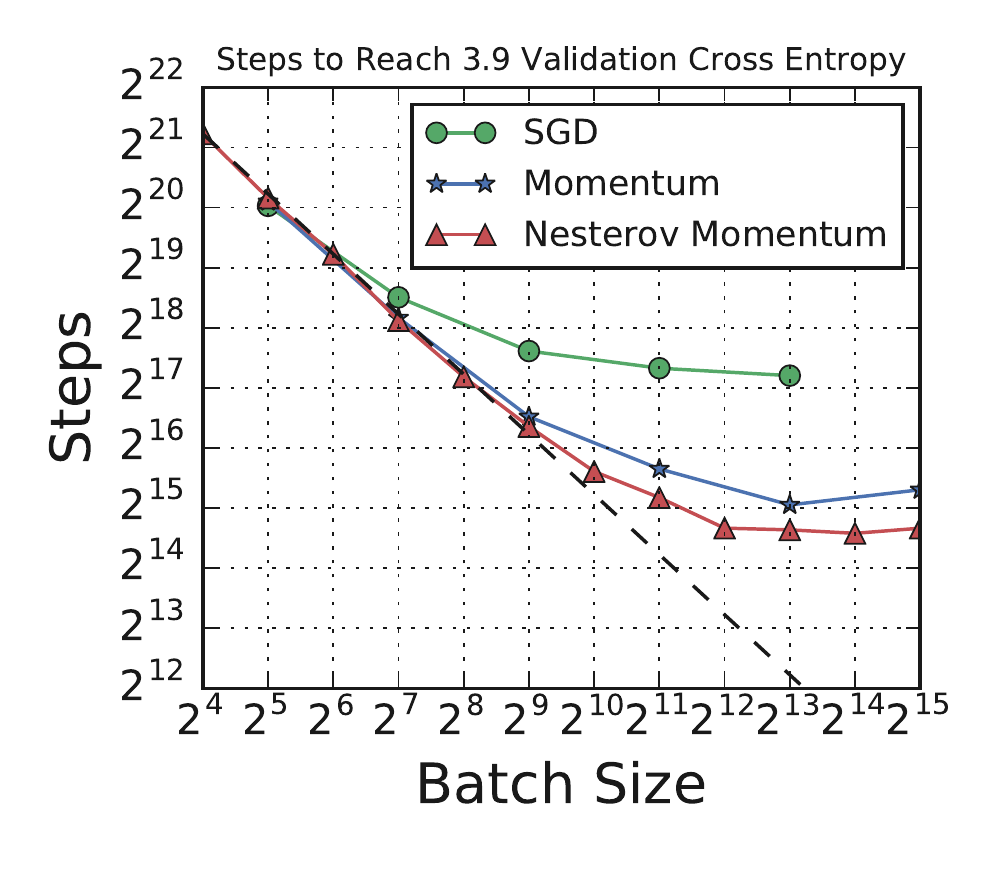}
        \vspace*{\capshift}
        \caption{Transformer Shallow on LM1B}
        \label{fig:stt-optimizers-transformer-lm1b}
    \end{subfigure}
    \begin{subfigure}[b]{\threecolfigwidth}
        \includegraphics[width=\textwidth]{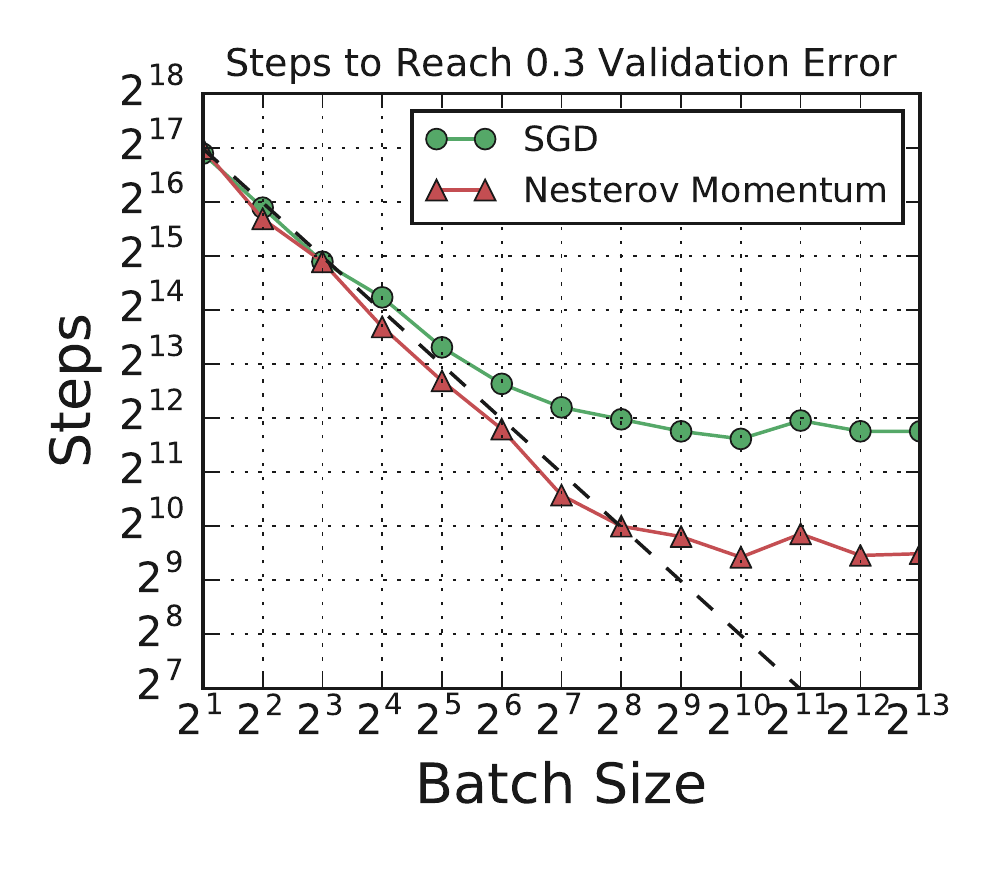}
        \vspace*{\capshift}
        \caption{ResNet-8 on CIFAR-10}
        \label{fig:stt-optimizers-resnet-cifar}
    \end{subfigure}
    \caption{\textbf{Momentum extends perfect scaling to larger batch sizes, but matches plain SGD at small batch sizes.} Nesterov momentum and SGD with momentum can both significantly reduce the absolute number of training steps to reach a goal validation error, and also significantly extend the perfect scaling region and thus better exploit larger batches than plain mini-batch SGD.}
    \label{fig:stt-optimizers}
\end{figure}

Figure~\ref{fig:stt-optimizers} shows that Nesterov momentum and SGD with momentum can both 
extend the perfect scaling region beyond that achieved by SGD, 
and thus can significantly reduce the number of training steps required to reach a goal validation error at larger batch sizes. 
However, at batch sizes small enough that all optimizers are within their perfect scaling region, 
momentum optimizers perform identically to SGD without momentum. 
Though initially surprising, 
this identical performance at small batch sizes is consistent with observations made in \citet{kidambi2018insufficiency}. 
In our experiments, we did not see a large difference between Nesterov momentum and SGD with momentum---Nesterov momentum appears to scale slightly better for Transformer on LM1B, but both perform about equally well for Simple CNN on MNIST.

\subsection{The Data Set Matters, at Least Somewhat}
\label{sec:stt-datasets}

\begin{figure}
    \centering
    \begin{subfigure}[b]{\threecolfigwidth}
    \includegraphics[width=\textwidth]{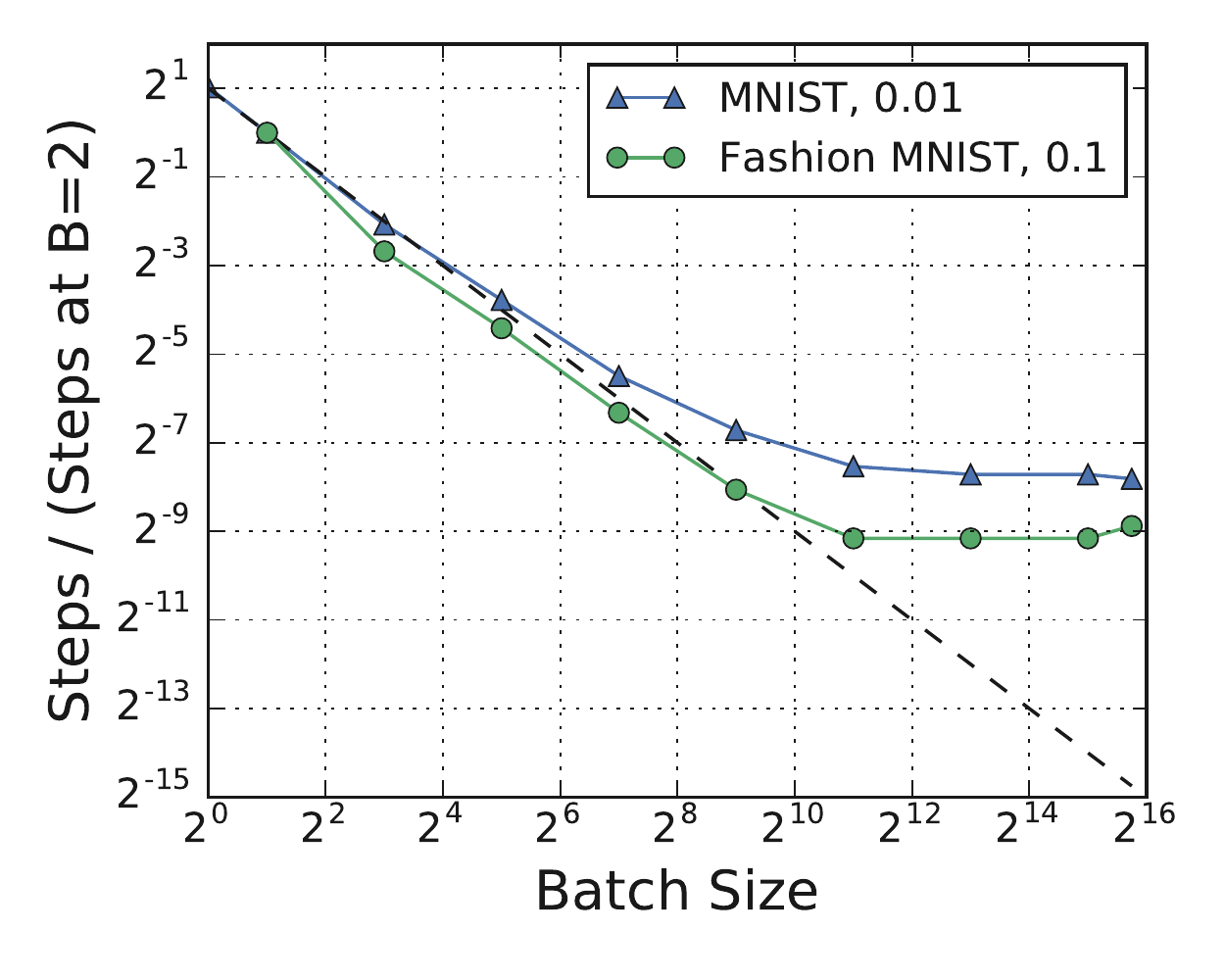}
    \vspace*{\capshift}
    \caption{Simple CNN on different data sets}
    \label{fig:stt-mnist-and-fashion}
    \end{subfigure}
    \begin{subfigure}[b]{\threecolfigwidth}
        \includegraphics[width=\textwidth]{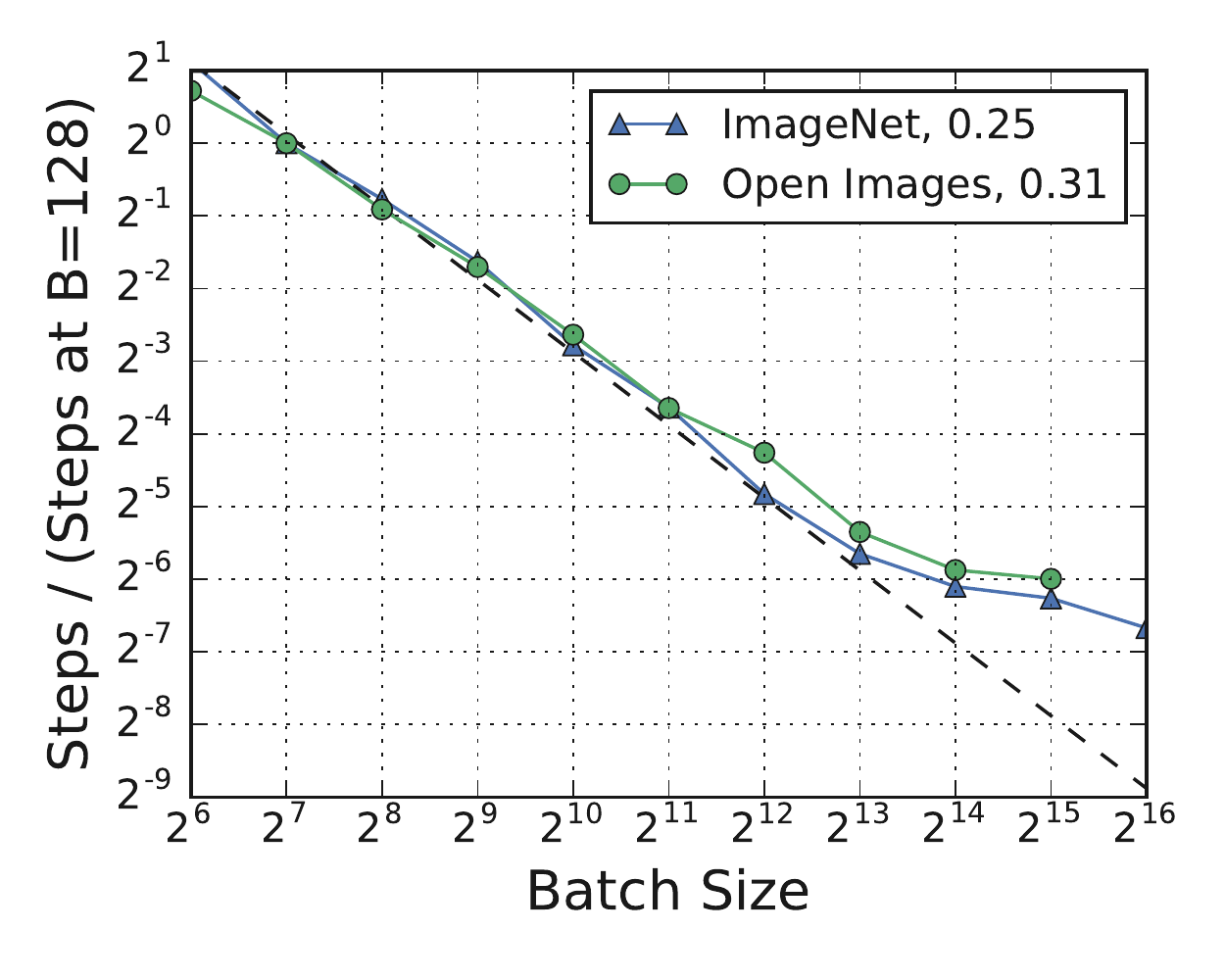}
        \vspace*{\capshift}
        \caption{ResNet-50 on different data sets}
        \label{fig:stt-resnet-imagenet-oi}
    \end{subfigure}
    \begin{subfigure}[b]{\threecolfigwidth}
        \includegraphics[width=\textwidth]{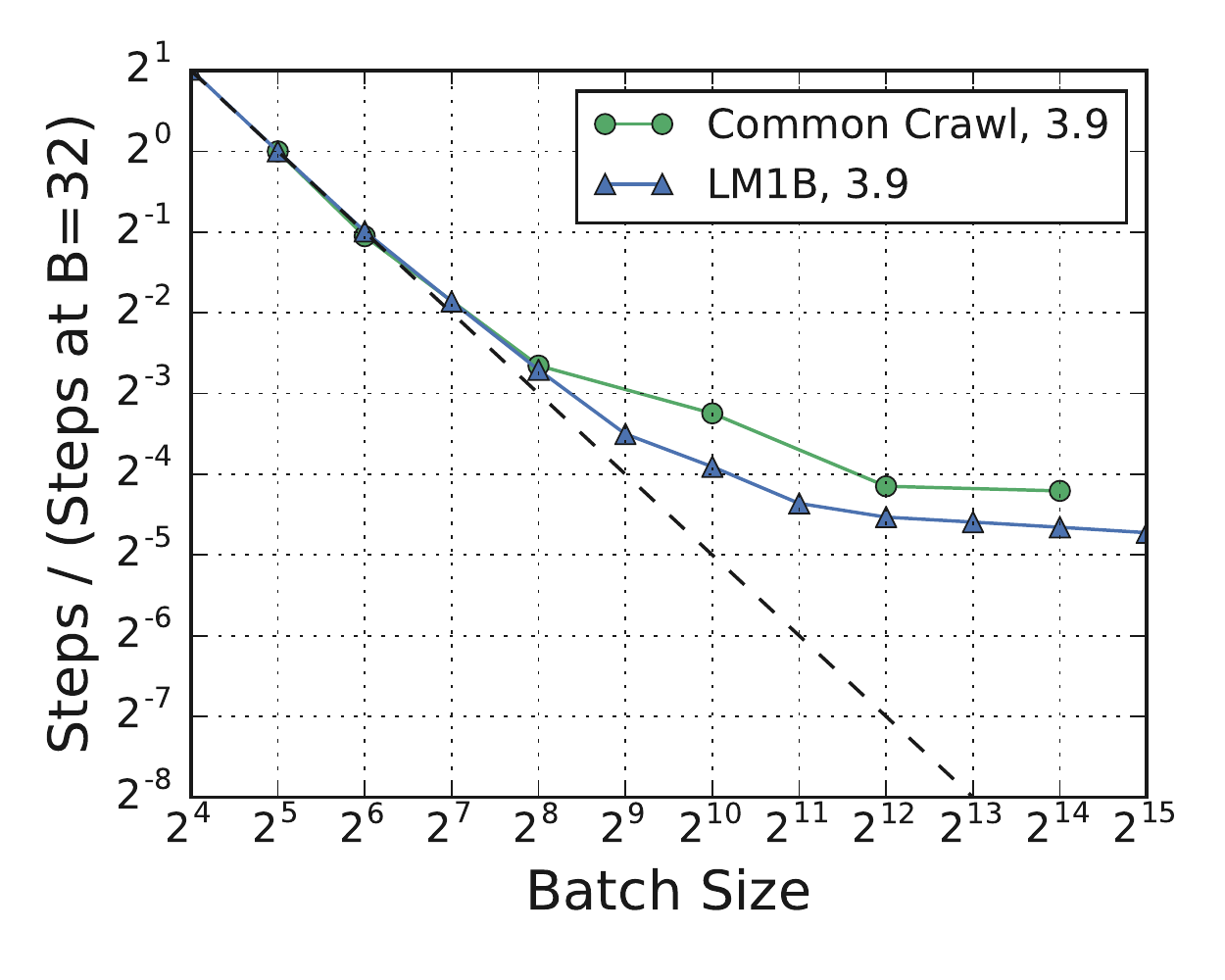}
        \vspace*{\capshift}
        \caption{Transformer on different data sets}
        \label{fig:stt-transformer-lm1b-cc}
    \end{subfigure}
    \caption{\textbf{The data set can influence the maximum useful batch size}. For the data sets shown in this plot, these differences are not simply as straightforward as larger data sets making larger batch sizes more valuable. Appendix~\ref{appendix:metrics} describes the evaluation metric used for each data set, and the plot legends show the goal metric value for each task. Figure~\ref{fig:stt-datasets-raw} in the Appendix contains these plots without the $y$-axis normalized.}
    \label{fig:stt-datasets}
\end{figure}

\begin{figure}
    \centering
    \begin{subfigure}[b]{0.49\textwidth}
        \includegraphics[width=\textwidth]{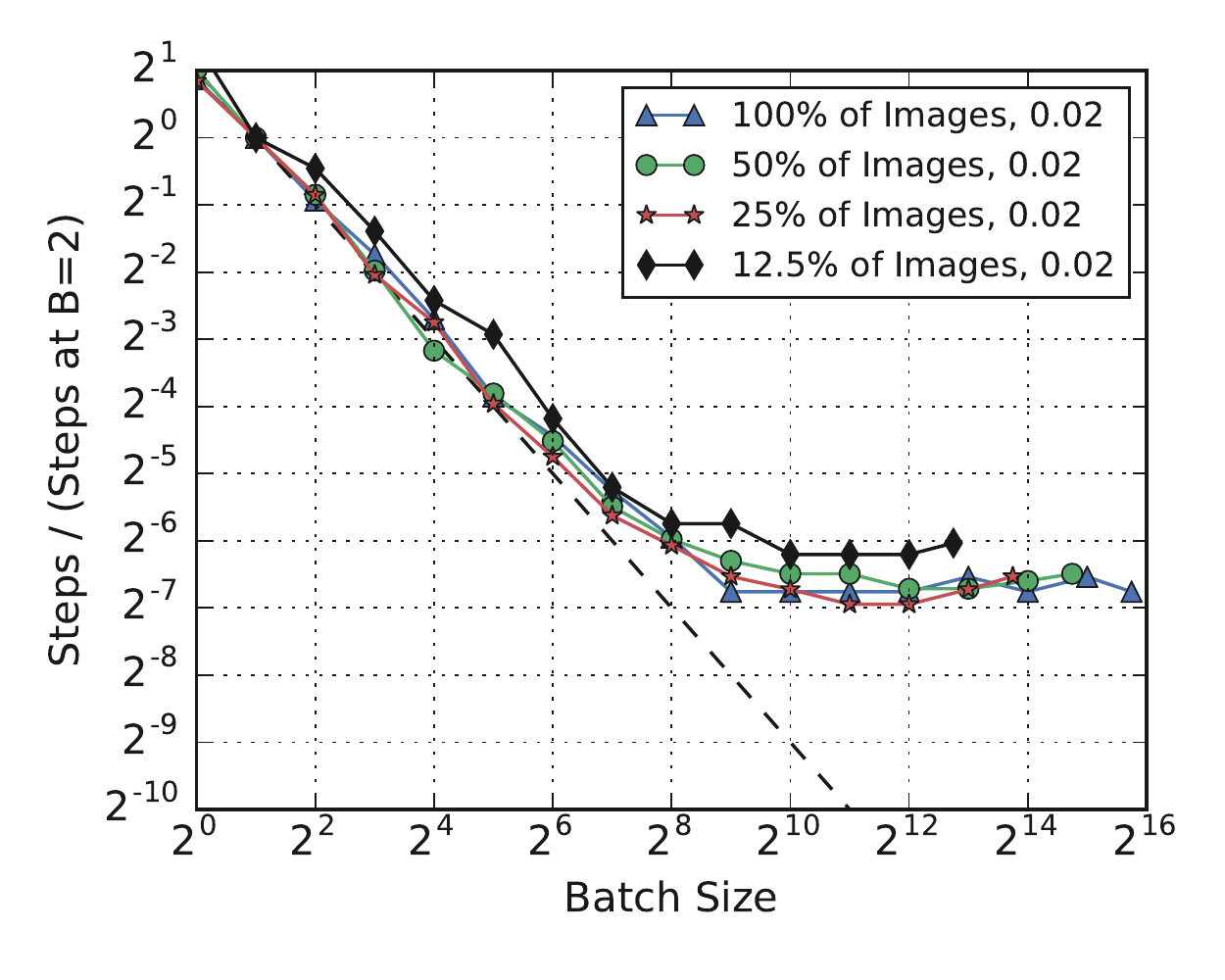}
        \vspace*{\capshift}
        \caption{Simple CNN on MNIST subsets}
        \label{fig:stt-mnist-size}
    \end{subfigure}
    \begin{subfigure}[b]{0.49\textwidth}
        \includegraphics[width=\textwidth]{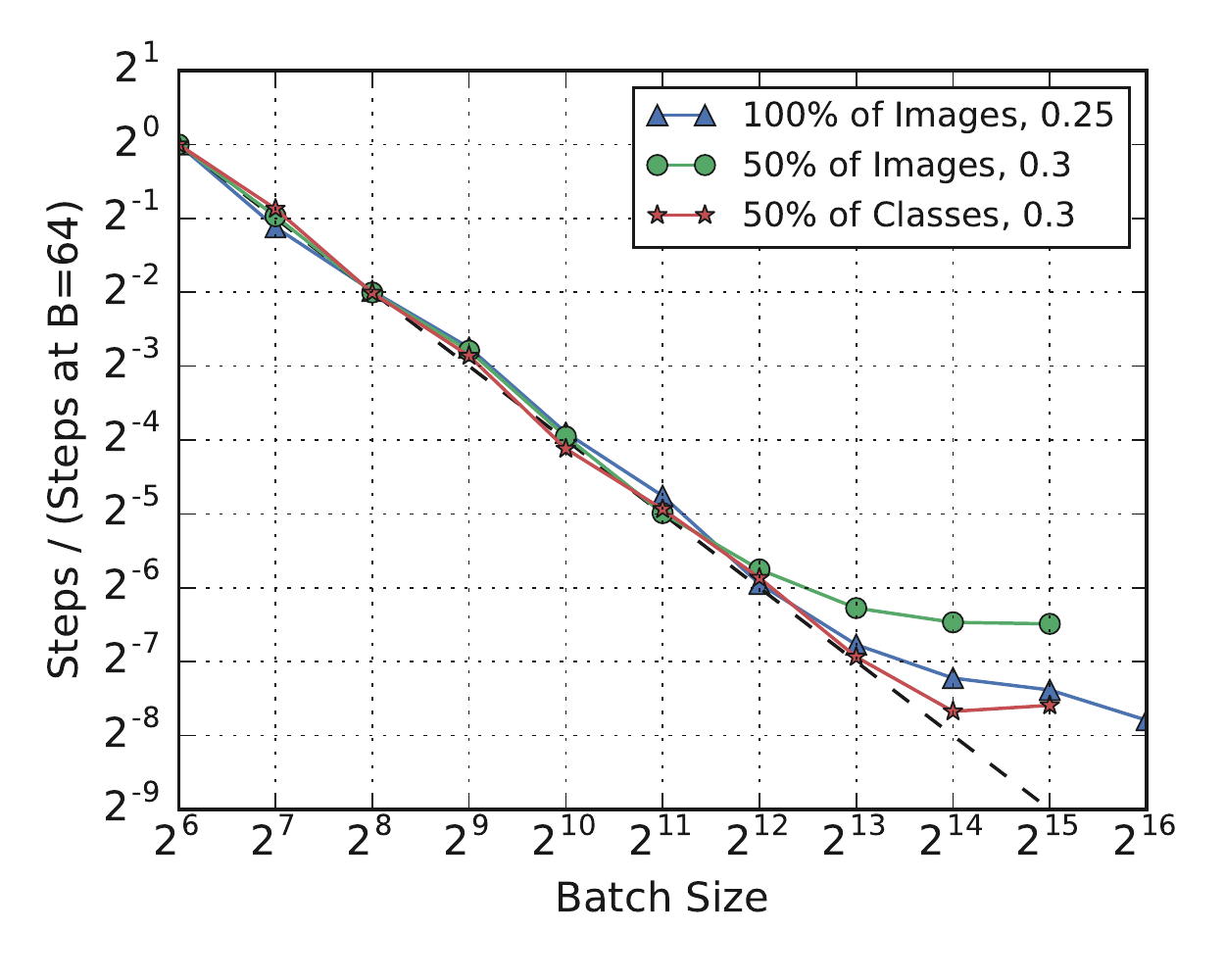}
        \vspace*{\capshift}
        \caption{ResNet-50 on ImageNet subsets}
        \label{fig:stt-resnet-half}
    \end{subfigure}
    \caption{\textbf{Investigating the effect of data set size.} At least for MNIST, any effect of subset size on the maximum useful batch size is extremely small or nonexistent. For ImageNet, the random subset of half the images deviates from perfect scaling sooner than the full data set, but the curve for the subset with half the classes is very close to the curve for the full data set and, if anything, deviates from perfect scaling later. Appendix~\ref{appendix:metrics} describes the evaluation metric used for each data set, and the plot legends show the goal metric value for each task. Figure~\ref{fig:stt-dataset-sizes-raw} in the Appendix contains these plots without the $y$-axis normalized.
    }
    \label{fig:stt-dataset-sizes}
\end{figure}

We investigated whether properties of the data set make some problems able to exploit larger batch sizes than others by experimenting with different data sets while keeping the model and optimizer fixed. We approached this in two ways: (i) by testing the same model on completely different data sets, and (ii) by testing the same model on different subsets of the same data set. We normalized the $y$-axis of all plots in this section in the same way as Section~\ref{sec:stt-different-models}. Appendix~\ref{appendix:additional-plots} contains all plots in this section without the $y$-axis normalized.

Figure~\ref{fig:stt-datasets} shows that changing the data set can affect the relationship between batch size and the number of steps needed to reach a goal validation error. Figure~\ref{fig:stt-mnist-and-fashion} shows that Fashion MNIST deviates from perfect scaling at a slightly larger batch size than MNIST for the Simple CNN model. Figure~\ref{fig:stt-resnet-imagenet-oi} shows that ImageNet and Open Images are extremely similar in how well ResNet-50 can make use of larger batch sizes, although, if anything, ImageNet might make slightly better use of larger batch sizes. Figure~\ref{fig:stt-transformer-lm1b-cc} shows that LM1B scales slightly better with increasing batch size than Common Crawl for Transformer. Since Fashion MNIST is the same size as MNIST, Open Images is larger than ImageNet, and Common Crawl is far larger than LM1B, these differences are not simply as straightforward as larger data sets making larger batch sizes more valuable. 

To disentangle the effects from changes to the distribution and changes to the number of examples, we generated steps to result vs batch size plots for different random subsets of MNIST (Figure~\ref{fig:stt-mnist-size}) and ImageNet (Figure~\ref{fig:stt-resnet-half}). For MNIST, we selected subsets of different sizes, while for ImageNet, we selected a random subset of half the images and a similar sized subset that only includes images from half of the classes. At least on MNIST, any effect on the maximum useful batch size is extremely small or nonexistent. For ImageNet, Figure~\ref{fig:stt-resnet-half} shows that the random subset of half the images deviates from perfect scaling sooner than the full data set, but the curve for the subset with half the classes is very close to the curve for the full data set and, if anything, deviates from perfect scaling later, even though it contains roughly the same number of images as the random subset.

\subsection{Regularization Can Be More Helpful at Some Batch Sizes Than Others}\label{sec:label-smoothing}

\begin{figure}
    \centering
    \begin{subfigure}[b]{0.49\textwidth}
        \includegraphics[width=\textwidth]{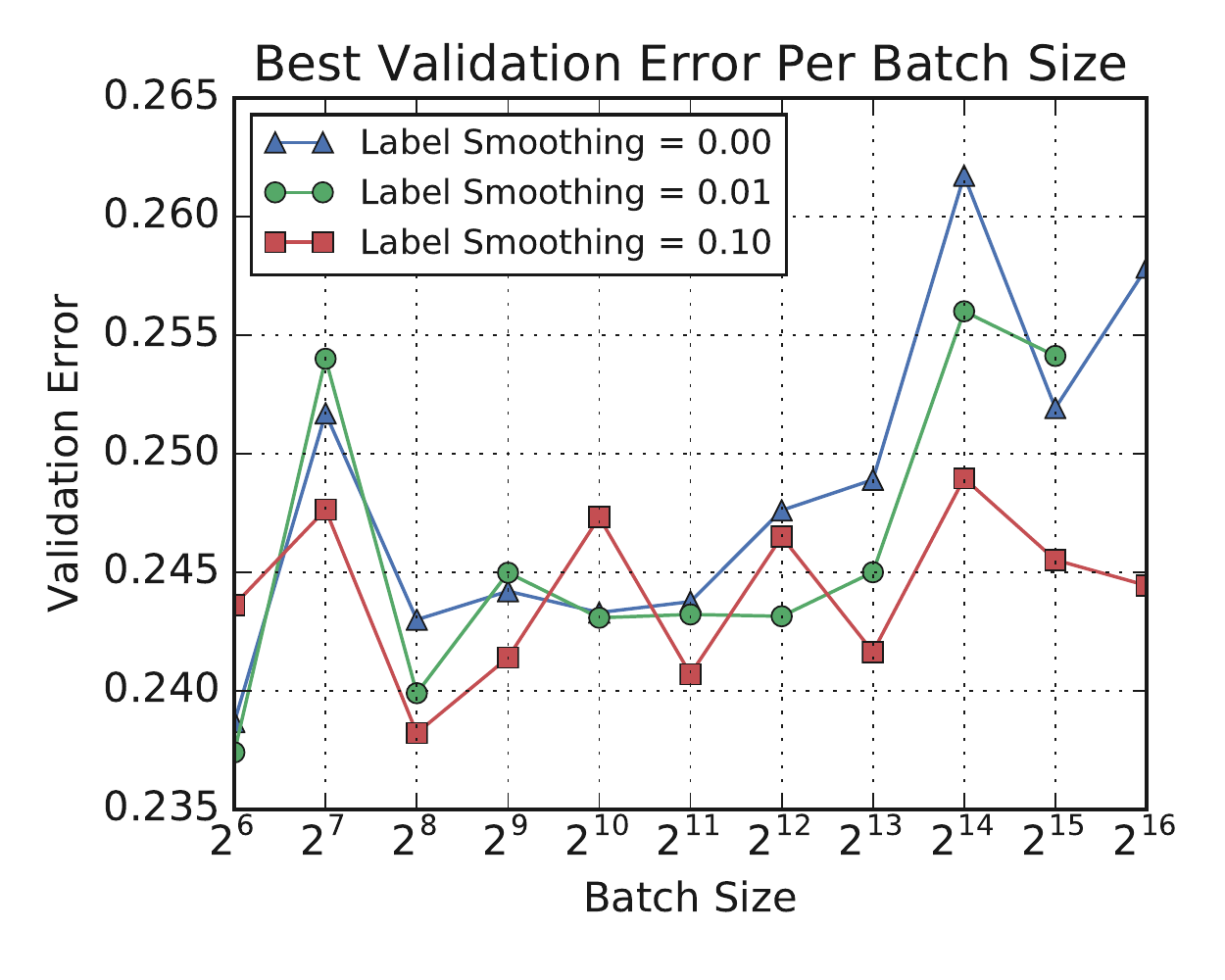}
        \vspace*{\capshift}
        \caption{Label smoothing benefits larger batch sizes, but has no apparent effect for smaller batch sizes.}
        \label{fig:label-smoothing-solution-quality}
    \end{subfigure}
    \begin{subfigure}[b]{0.49\textwidth}
        \includegraphics[width=\textwidth]{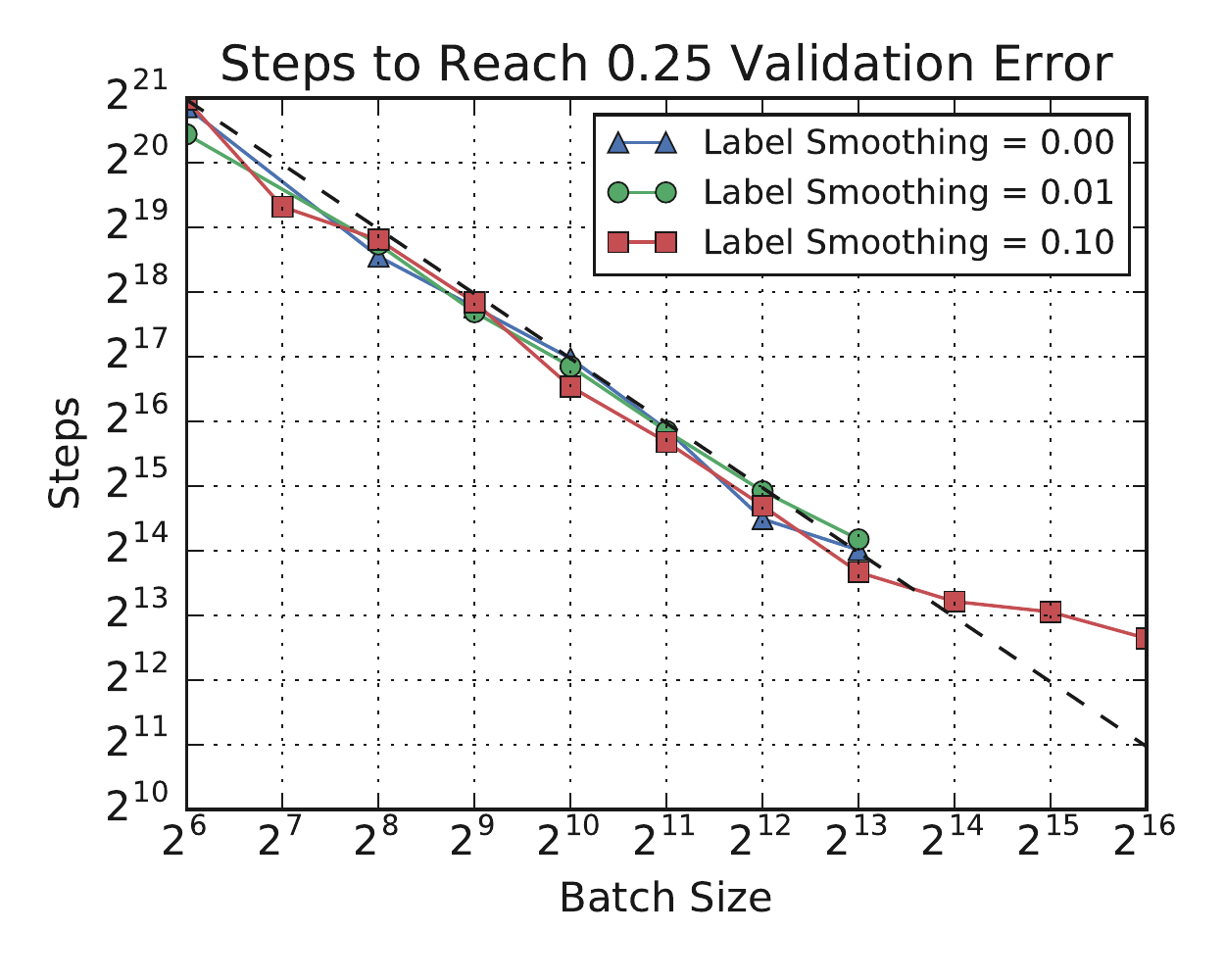}
        \vspace*{\capshift}
        \caption{Label smoothing has no apparent effect on training speed, provided the goal error is achieved.}
        \label{fig:label-smoothing-stt}
    \end{subfigure}
    \caption{\textbf{Regularization can be more helpful at some batch sizes than others.} Plots are for ResNet-50 on ImageNet. Each point corresponds to a different metaparameter tuning trial, so the learning rate, Nesterov momentum, and learning rate schedule are independently chosen for each point. The training budget is fixed for each batch size, but varies between batch sizes.}
    \label{fig:label-smoothing}
\end{figure}

We used label smoothing \citep{szegedy2016rethinking} to regularize training in our experiments with ResNet-50 on ImageNet. Without label smoothing, we could not achieve our goal validation error rate of 0.25 with batch sizes greater than $2^{14}$ within our training budget. With a fixed compute budget for each batch size, label smoothing improved the error by as much as one percentage point at large batch sizes, while having no apparent effect at small batch sizes (Figure~\ref{fig:label-smoothing-solution-quality}). Meanwhile, if multiple choices for the label smoothing metaparameter achieved the goal within the training budget, then label smoothing did not change the number of steps needed (Figure~\ref{fig:label-smoothing-stt}).

We confirmed that label smoothing reduced overfitting at large batch sizes for ResNet-50 on ImageNet (see Figure~\ref{fig:label-smoothing-training-curves} in the Appendix). This is consistent with the idea that noise from small batch training is a form of implicit regularization \citep[e.g.][]{GoodfellowEtAlBook2016}. However, although our results show that \emph{other forms of regularization can serve in place of this noise}, it might be difficult to select and tune other forms of regularization for large batch sizes. For example, we unsuccessfully tried to control overfitting with larger batch sizes by increasing the $L_2$ weight penalty and by applying additive Gaussian gradient noise before we obtained good results with label smoothing.

Finally, we also tried label smoothing with Simple CNN on MNIST and Fashion MNIST, and found that it generally helped \textit{all} batch sizes, with no consistent trend of helping smaller or larger batch sizes more (see Figure~\ref{fig:label-smoothing-mnists} in the Appendix), perhaps because these data sets are sufficiently small and simple that overfitting is an issue at all batch sizes.

\subsection{
The Best Learning Rate and Momentum Vary with Batch Size
}\label{sec:effective-lr}

\begin{figure}
    \centering
    \begin{subfigure}[b]{\threecolfigwidth}
        \includegraphics[width=\textwidth]{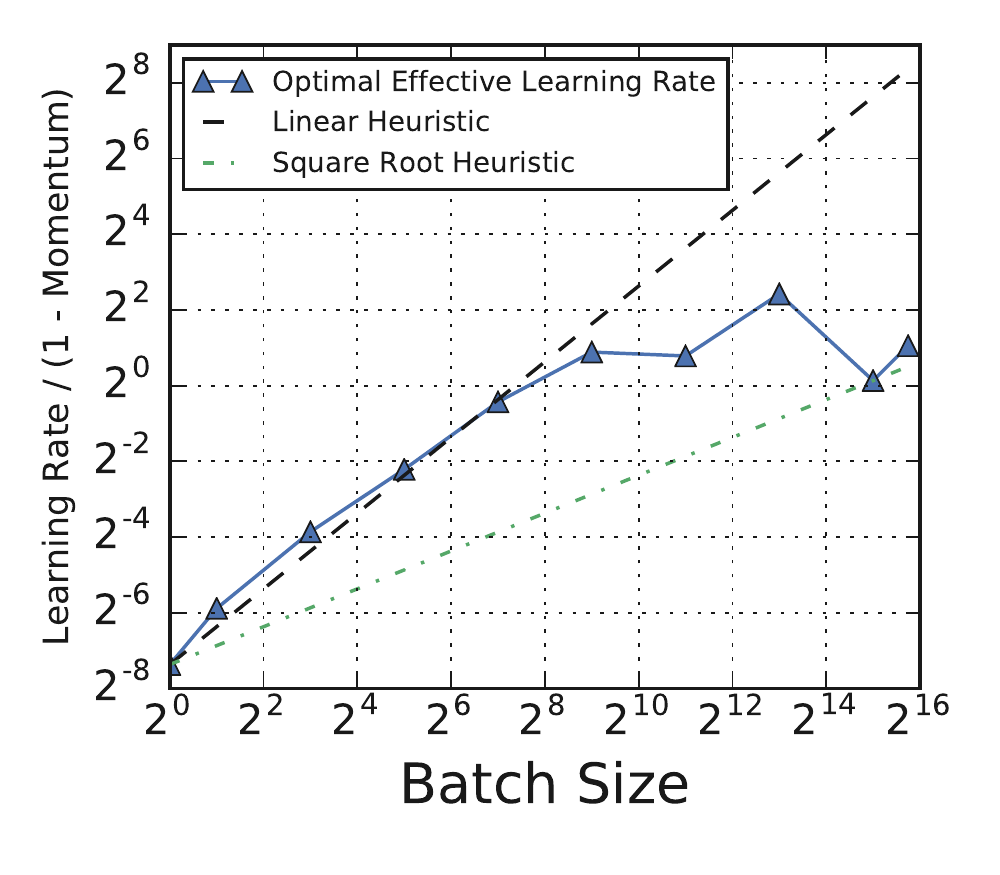}
        \vspace*{\capshift}
        \caption{Simple CNN on MNIST}
        \label{fig:hparams-lr-cnn-mnist}
    \end{subfigure}
    \begin{subfigure}[b]{\threecolfigwidth}
        \includegraphics[width=\textwidth]{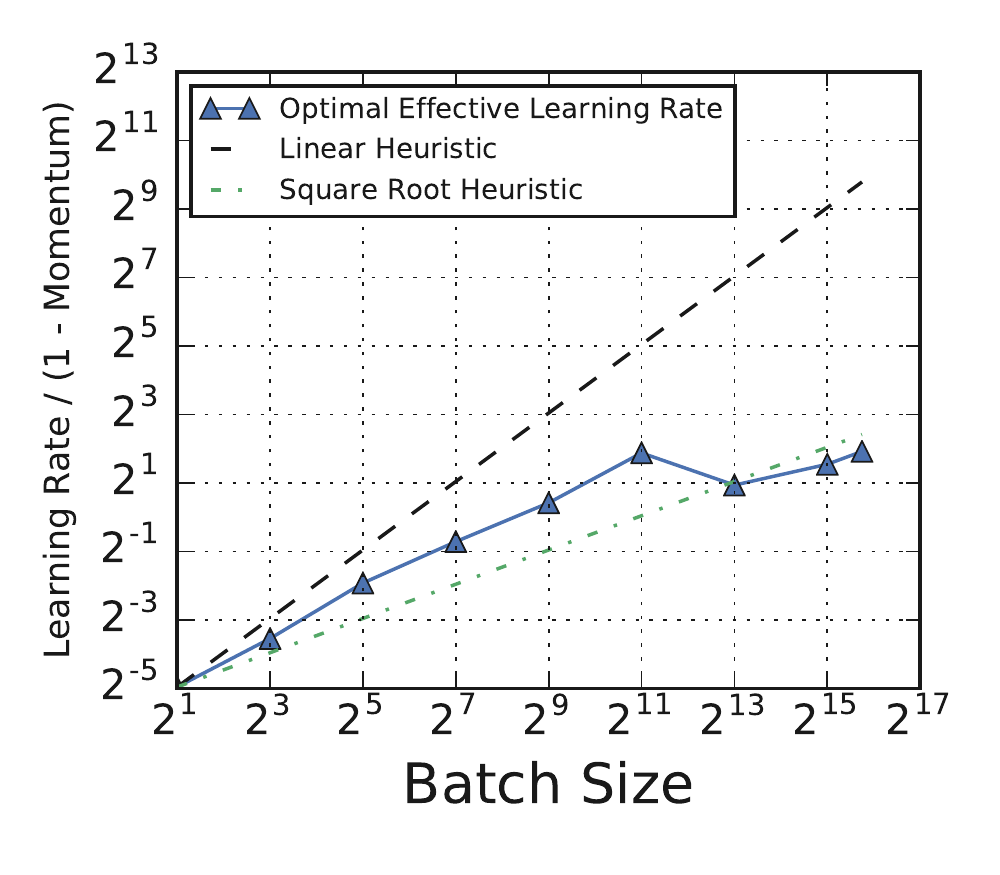}
        \vspace*{\capshift}
        \caption{Simple CNN on Fashion MNIST}
        \label{fig:hparams-lr-cnn-fmnist}
    \end{subfigure}
    \begin{subfigure}[b]{\threecolfigwidth}
        \includegraphics[width=\textwidth]{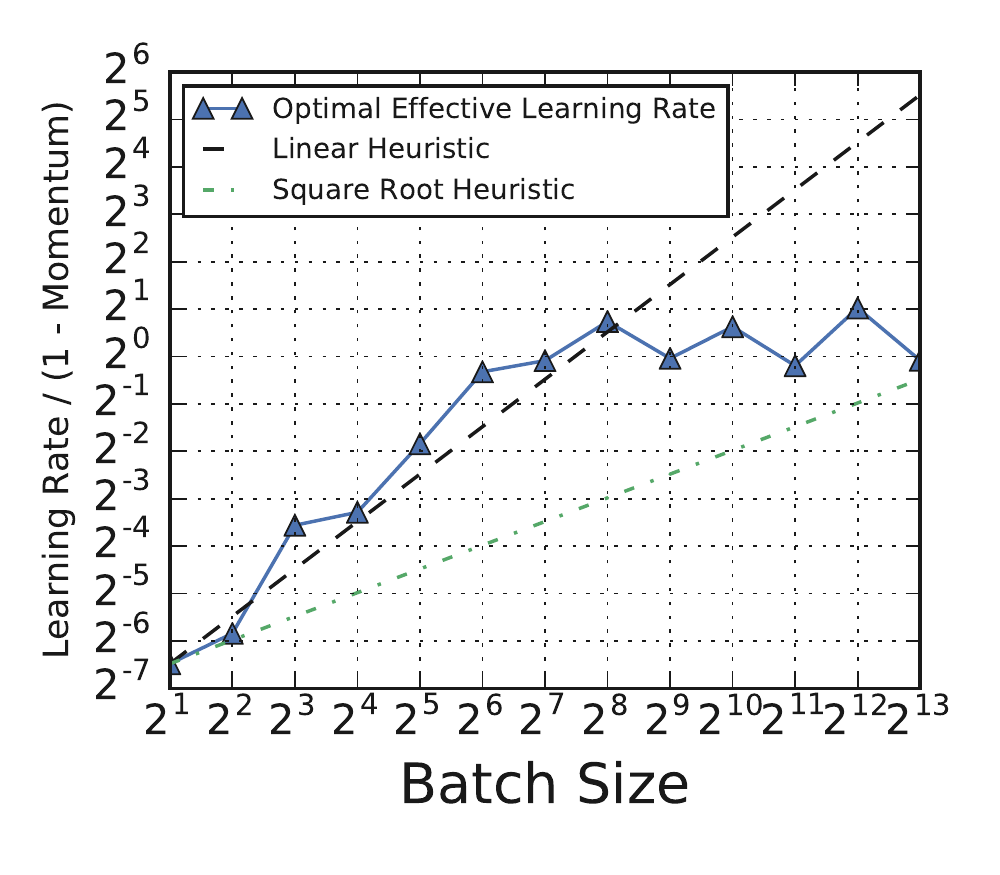}
        \vspace*{\capshift}
        \caption{ ResNet-8 on CIFAR-10}
        \label{fig:hparams-lr-resnet-cifar}
    \end{subfigure} \\
    \vspace*{\lineshift}
    \begin{subfigure}[b]{\threecolfigwidth}
        \includegraphics[width=\textwidth]{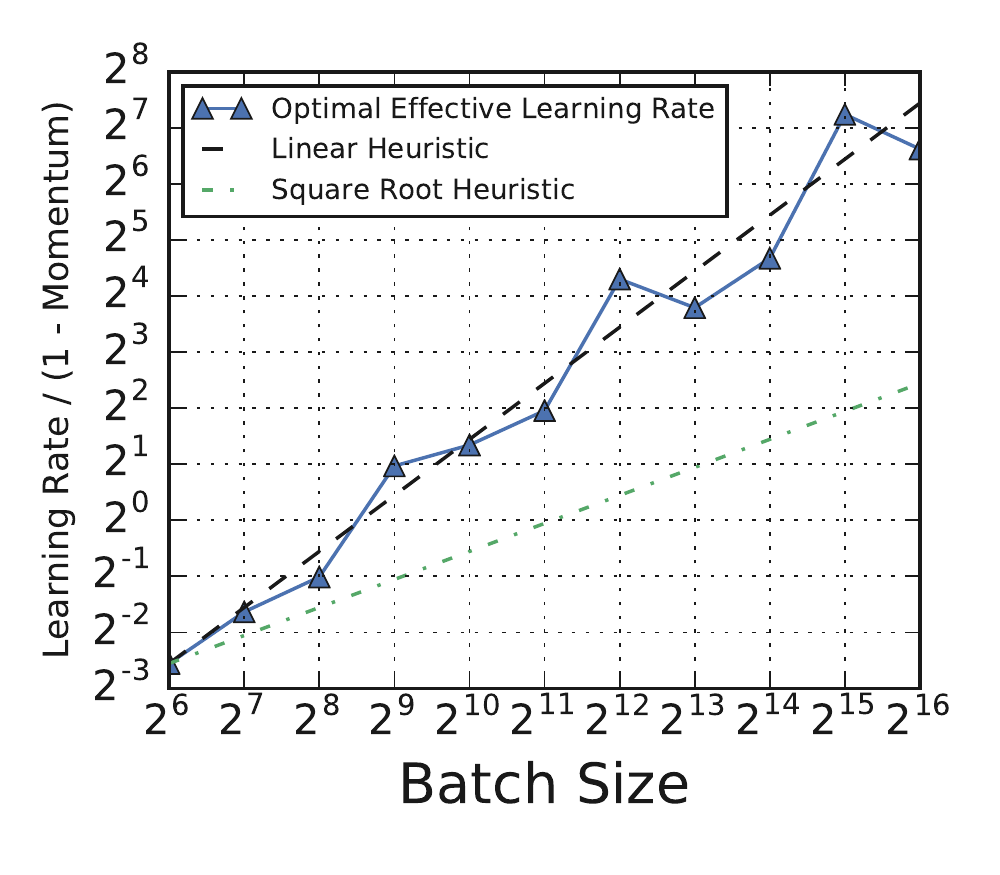}
        \vspace*{\capshift}
        \caption{ResNet-50 on ImageNet}
        \label{fig:hparams-lr-resnet-imagenet}
    \end{subfigure}
    \begin{subfigure}[b]{\threecolfigwidth}
        \includegraphics[width=\textwidth]{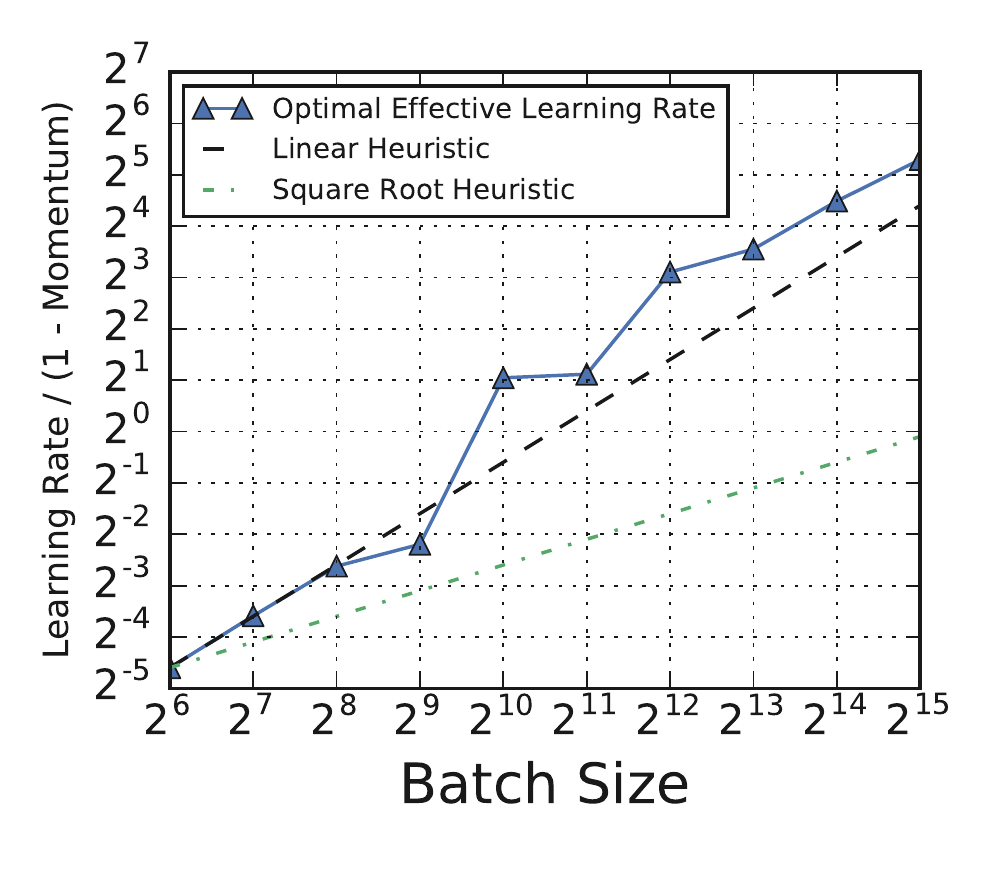}
        \vspace*{\capshift}
        \caption{ ResNet-50 on Open Images}
        \label{fig:hparams-lr-resnet-oi}
    \end{subfigure}
    \begin{subfigure}[b]{\threecolfigwidth}
        \includegraphics[width=\textwidth]{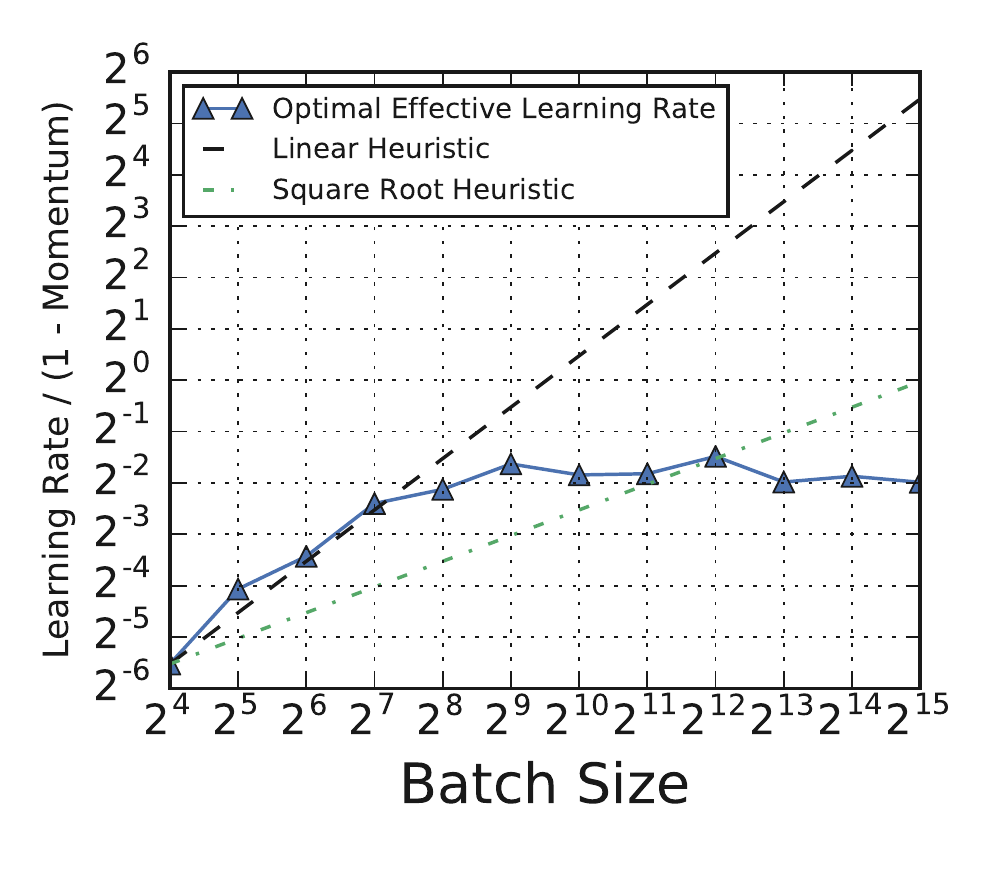}
        \vspace*{\capshift}
        \caption{ Transformer on LM1B}
        \label{fig:hparams-lr-transformer-lm1b}
    \end{subfigure}\\
    \vspace*{\lineshift}
    \begin{subfigure}[b]{\threecolfigwidth}
        \includegraphics[width=\textwidth]{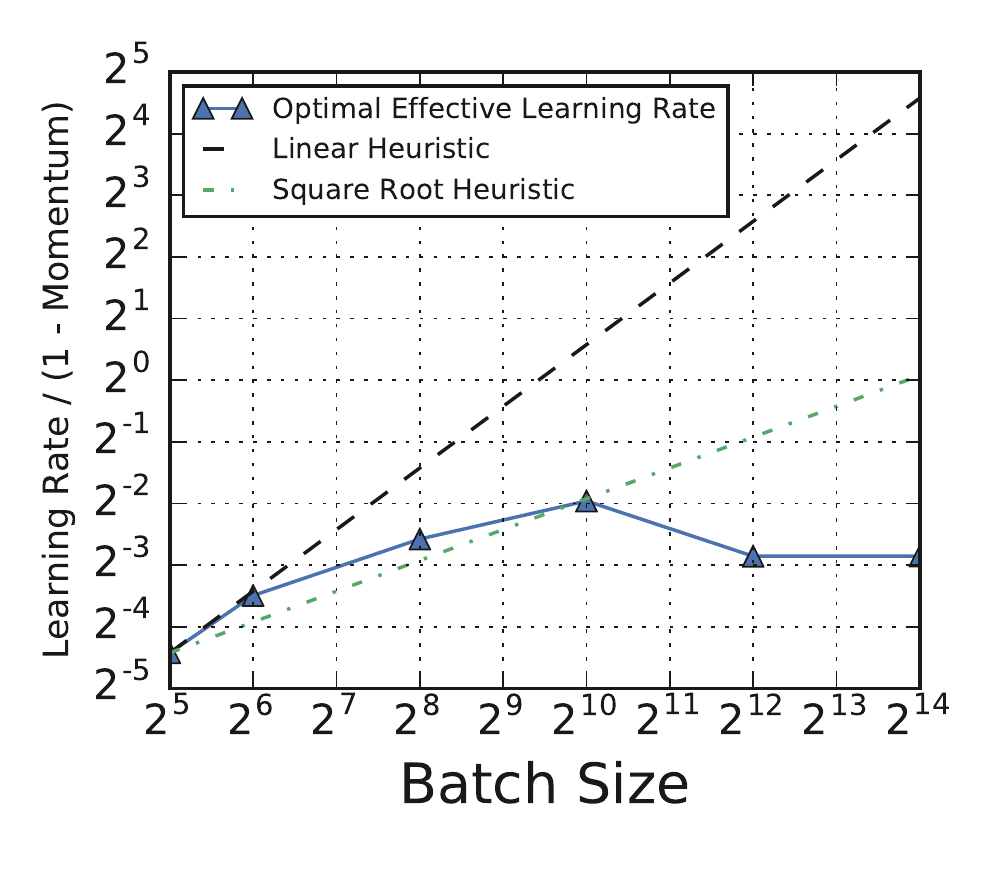}
        \vspace*{\capshift}
        \caption{Transformer on Common Crawl}
        \label{fig:hparams-lr-transformer-cc}
    \end{subfigure}
    \begin{subfigure}[b]{\threecolfigwidth}
        \includegraphics[width=\textwidth]{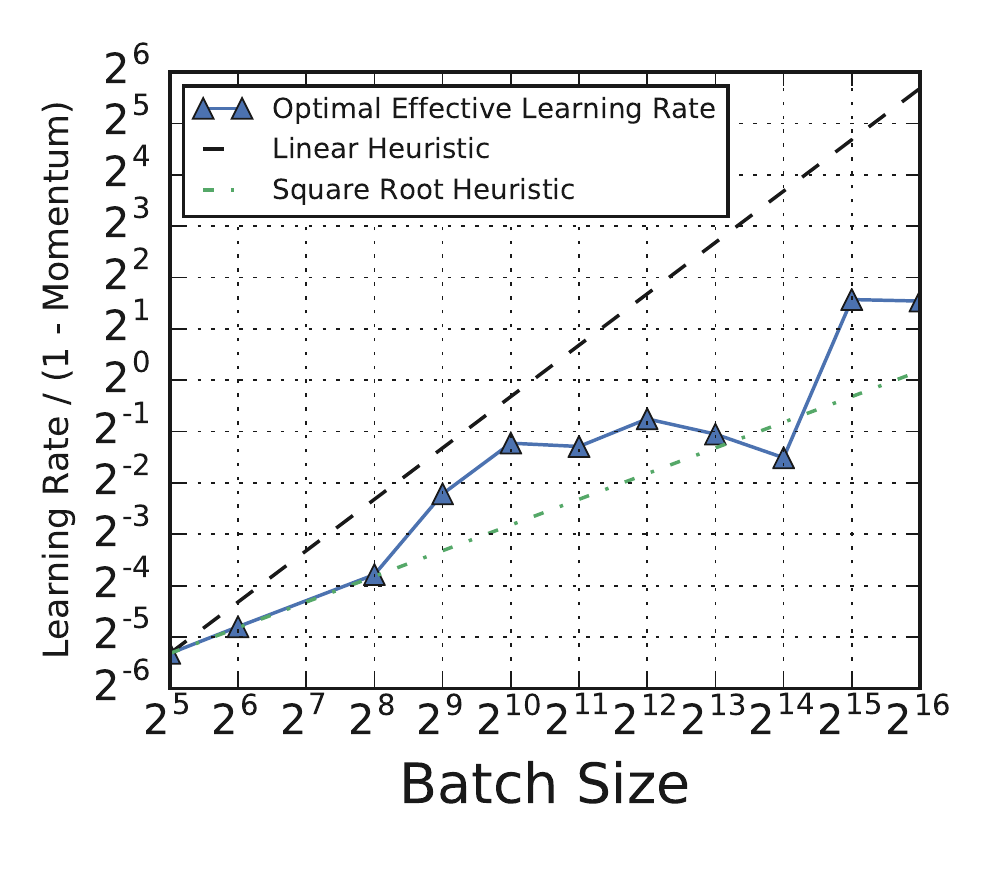}
        \vspace*{\capshift}
        \caption{ VGG-11 on ImageNet}
        \label{fig:hparams-lr-vgg11-imagenet}
    \end{subfigure}
        \begin{subfigure}[b]{\threecolfigwidth}
        \includegraphics[width=\textwidth]{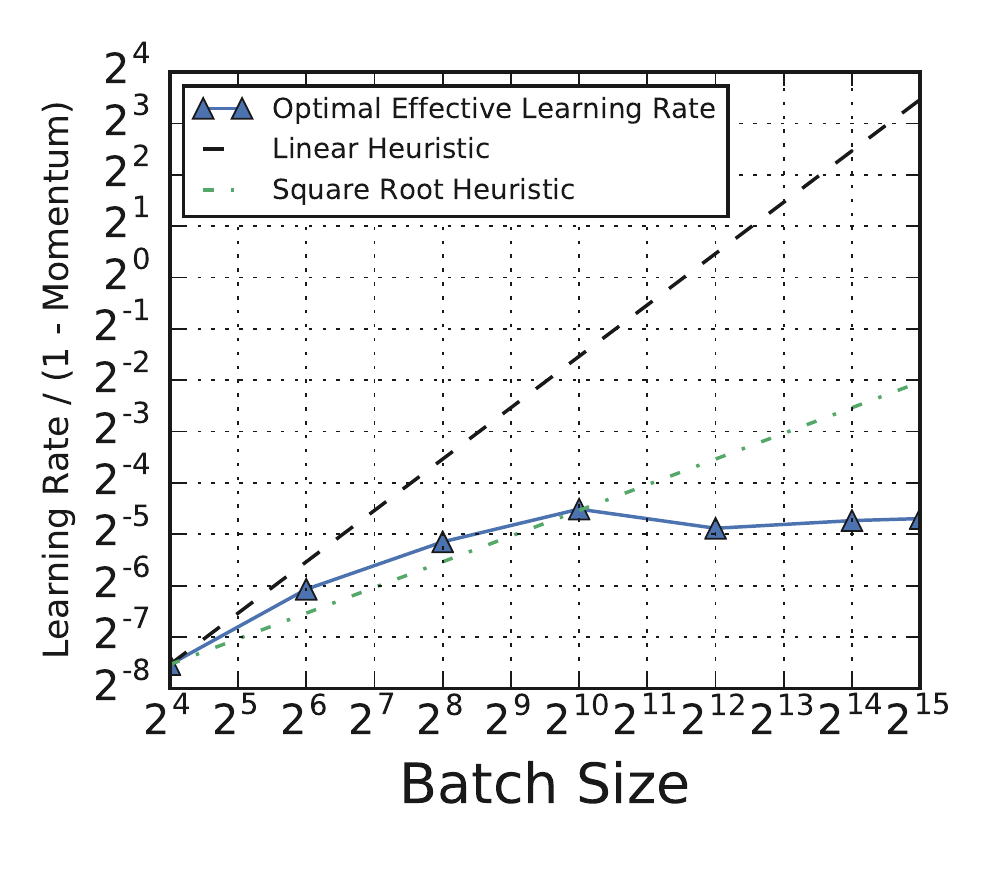}
        \vspace*{\capshift}
        \caption{LSTM on LM1B}
        \label{fig:hparams-lr-lstm-lm1b}
    \end{subfigure}
    \caption{\textbf{Optimal effective learning rates do not always follow linear or square root scaling heuristics.} Effective learning rates correspond to the trial that reached the goal validation error in the fewest training steps (see Figure~\ref{fig:stt-problems}). For models that used learning rate decay schedules (ResNet-8, ResNet-50, VGG-11), plots are based on the initial learning rate. See Figures~\ref{fig:raw-lr} and~\ref{fig:momentum} in the Appendix for separate plots of the optimal learning rate and momentum.}
    \label{fig:effective-lr}
\end{figure}

Across all problems we considered, the effective learning rate ($\eta^\text{eff}$; see Section~\ref{sec:setup-algos}) that minimized the number of training steps to a goal validation error tended to increase with increasing batch size (Figure~\ref{fig:effective-lr}). However, it did not always follow either a linear or square root scaling heuristic, despite the popularity of these rules of thumb. In some cases, the optimal effective learning rate even decreased for larger batch sizes. We also found that the best effective learning rate should be chosen by jointly tuning the learning rate and momentum, rather than tuning only the learning rate. For example, the optimal way to scale the effective learning rate for Transformer was to increase the momentum while decreasing the learning rate or holding it constant (see Figures~\ref{fig:raw-lr} and~\ref{fig:momentum} in the Appendix). This is a refinement to past prescriptions that only change the learning rate while keeping the momentum fixed.

\newcommand{\hparamsensitivitywidth}{0.99\textwidth}
\newcommand{\hparamsensitivitycapshift}{-1mm}
\begin{figure}
    \centering
    \begin{subfigure}[b]{\textwidth}
        \includegraphics[width=\hparamsensitivitywidth]{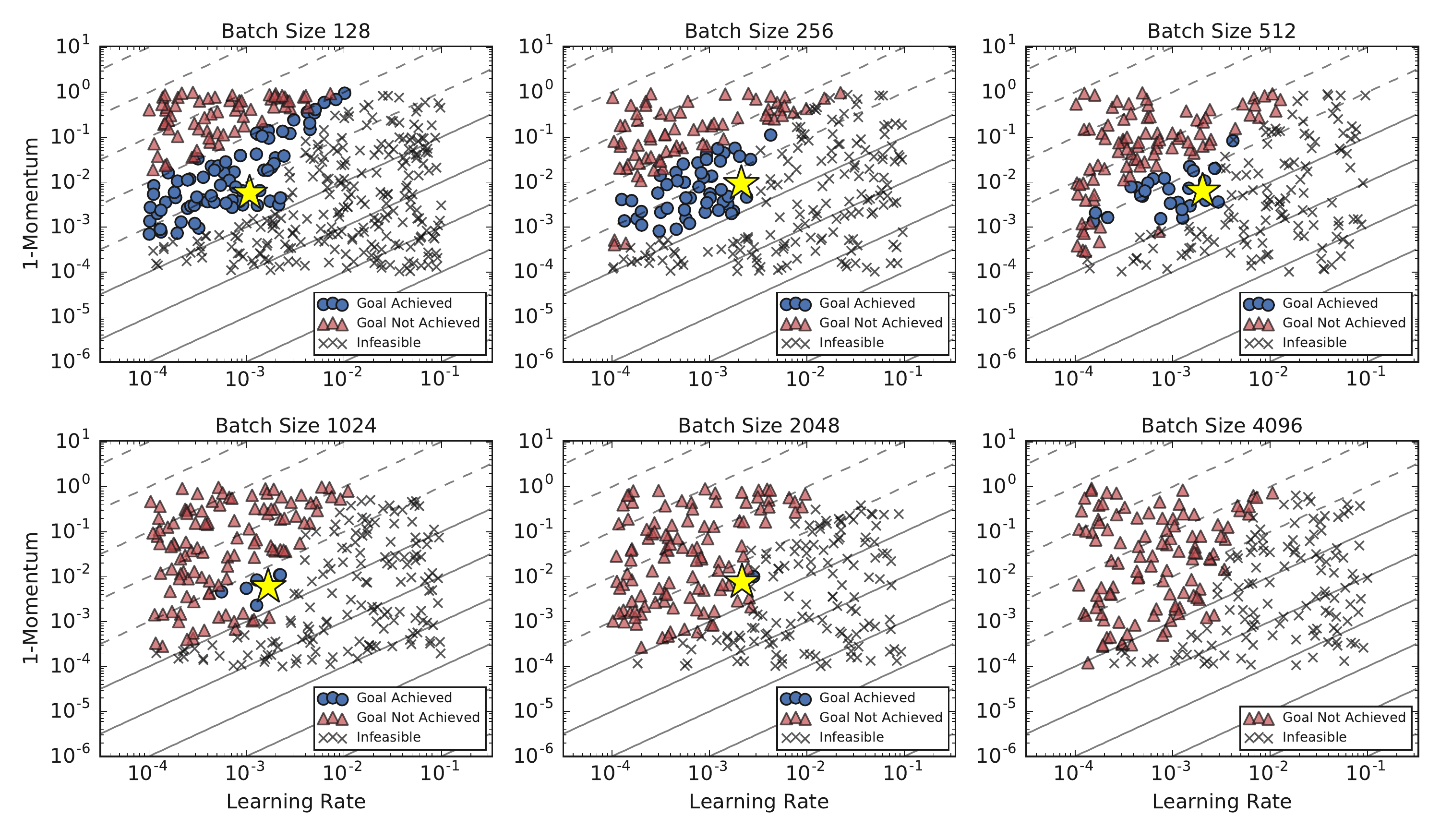}
        \vspace*{\hparamsensitivitycapshift}
        \caption{\small Transformer on LM1B with a training budget of one epoch.}
        \label{fig:hparams-sensitivity-transformer-epoch-budget}
    \end{subfigure}\\
    \vspace*{4mm}
    \begin{subfigure}[b]{\textwidth}
        \includegraphics[width=\hparamsensitivitywidth]{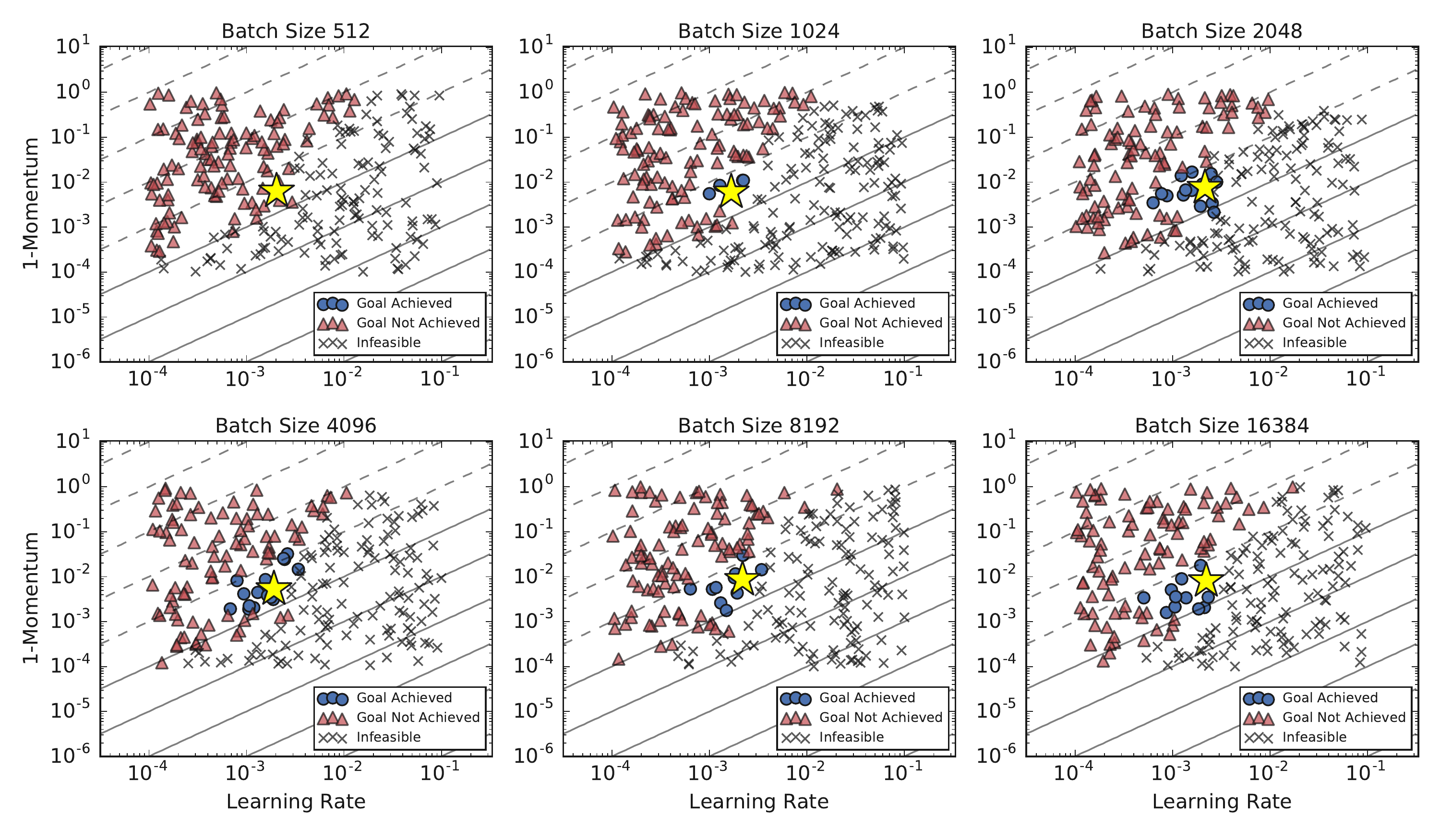}
        \vspace*{\hparamsensitivitycapshift}
        \caption{\small Transformer on LM1B with a training budget of 25,000 steps.}
        \label{fig:hparams-sensitivity-transformer-step-budget}
    \end{subfigure}
    
    \caption{
    \textbf{With increasing batch size, the region in metaparameter space corresponding to rapid training in terms of epochs becomes smaller, while the region in metaparameter space corresponding to rapid training in terms of step-count grows larger.}
    Yellow stars are the trials that achieved the goal in the fewest number of steps. Contours indicate the effective learning rate $\eta^\text{eff} = \frac{\eta}{1 - \gamma}$. Infeasible trials are those that resulted in divergent training.
    }
    \label{fig:hparams-sensitivity-transformer}
\end{figure}

\begin{figure}
    \centering
    \includegraphics[width=\textwidth]{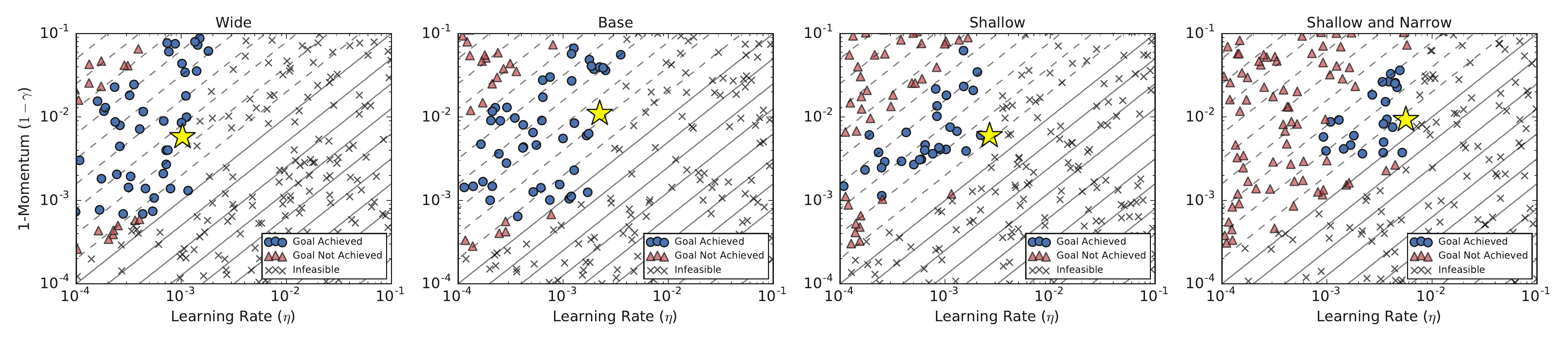}
    \caption{\textbf{Smaller models have larger stable learning rates for Transformer on LM1B.}
    Plots are for different sizes of Transformer on LM1B with a batch size of 1024, a goal validation cross entropy error of 4.2, and a training budget of 50,000 steps. Contours indicate the effective learning rate $\eta^\text{eff} = \frac{\eta}{1 - \gamma}$. Infeasible trials are those that resulted in divergent training.}
    \label{fig:hparams-sensitivity-transformer-size}
\end{figure}

We further investigated the relationship between learning rate, momentum, and training speed by examining our metaparameter search spaces for different batch sizes and model sizes. 
For this analysis, we used Transformer on LM1B with Nesterov momentum because the metaparameter search spaces are consistent between all batch and model sizes, and can be easily visualized because they consist only of the constant learning rate $\eta$ and the momentum $\gamma$. We observe the following behaviors:
\begin{itemize}
    \item  With increasing batch size, the region in metaparameter space corresponding to rapid training in terms of epochs becomes smaller \citep[Figure~\ref{fig:hparams-sensitivity-transformer-epoch-budget}, consistent with the findings of][]{breuel2015effects}, while the region in metaparameter space corresponding to rapid training in terms of step-count grows larger (Figure~\ref{fig:hparams-sensitivity-transformer-step-budget}, although it eventually plateaus for batch sizes in the maximal data parallelism regime).
    Thus, with a fixed error goal and in a setting where training epochs are constrained (e.g.\ a compute budget), it may become more challenging to choose good values for the metaparameters with increasing batch size. Conversely, with a fixed error goal and in a setting where training steps are constrained (e.g.\ a wall-time budget), it may become easier to choose good values for the metaparameters with increasing batch size.

    \item The metaparameters yielding the fastest training are typically on the edge of the feasible region of the search space (Figure~\ref{fig:hparams-sensitivity-transformer}). In other words, small changes in the optimal metaparameters might make training diverge. This behavior may pose a challenge for metaparameter optimization techniques, such as Gaussian Process approaches, that assume a smooth relationship between metaparameter values and model performance. It could motivate techniques such as learning rate warm-up that enable stability at larger eventual learning rates, since the maximum stable learning rate depends on the current model parameters. 
    We did not observe the same behavior for ResNet-50 on ImageNet. Figure~\ref{fig:hparams-error-vs-lr} in the Appendix shows the results for a range of effective learning rates near the optimum for ResNet-50 on ImageNet and Transformer on LM1B.

    \item Smaller models have larger stable learning rates (Figure~\ref{fig:hparams-sensitivity-transformer-size}). This is consistent with recent work predicting that the largest stable learning rate is inversely proportional to layer width \citep{karakida2018universal}.

\end{itemize}

\subsection{Solution Quality Depends on Compute Budget More Than Batch Size}
\label{sec:solution-quality}

\newcommand{\solqualfigwidth}{0.42\textwidth}
\newcommand{\solqualcapshift}{-3mm}
\newcommand{\solquallineshift}{1mm}
\newcommand{\solqualhspace}{3mm}
\begin{figure}
    \centering
    \begin{tabular}{
        >{\centering\arraybackslash}m{3in}
        >{\centering\arraybackslash}m{3in}
        }
    \toprule
    Step budget & Epoch budget \\
    \midrule
    \begin{subfigure}[b]{\textwidth}
        \centering
        \includegraphics[width=\solqualfigwidth]{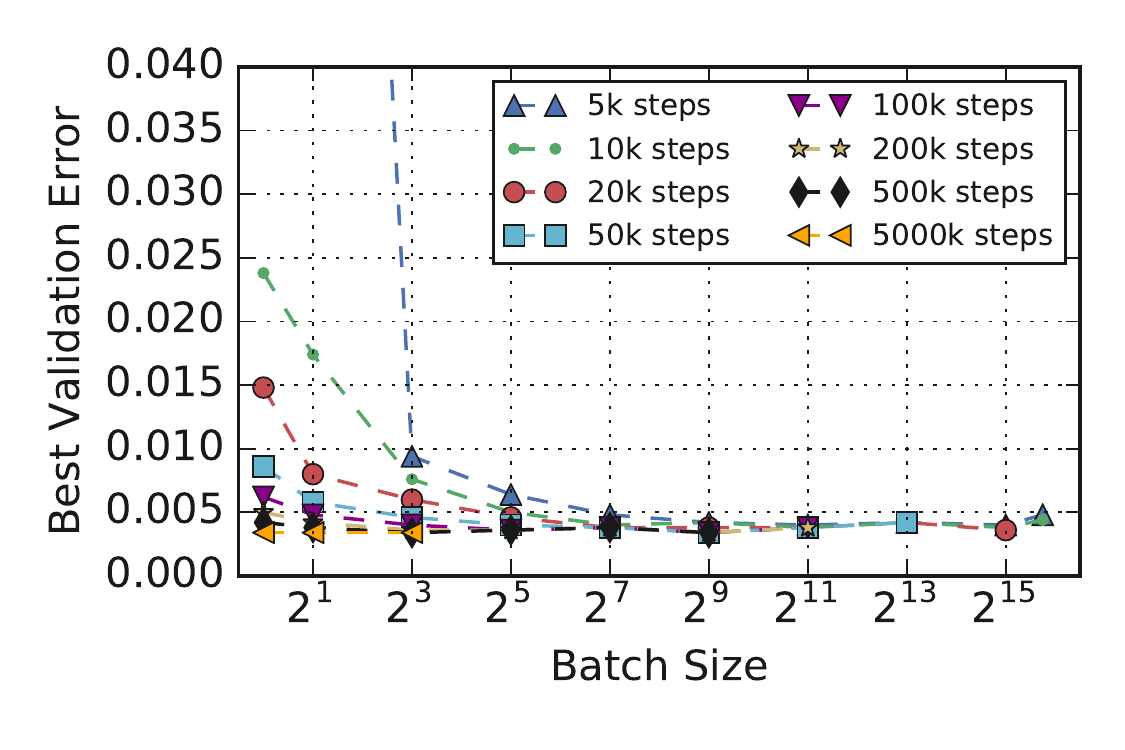}
        \hspace*{\solqualhspace}
        \includegraphics[width=\solqualfigwidth]{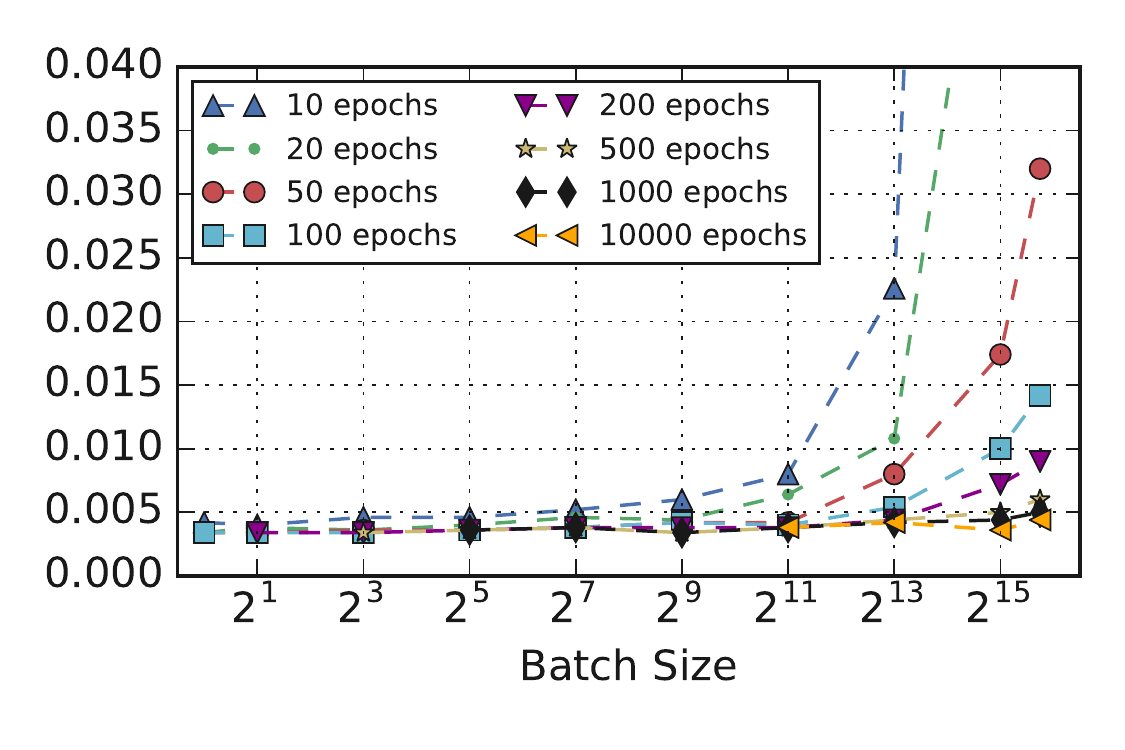}
        \vspace*{\solqualcapshift}
        \caption{\small Simple CNN on MNIST}\label{fig:sol-qual-mnist}
    \end{subfigure}\\
    \vspace*{\solquallineshift}
    \begin{subfigure}[b]{\textwidth}
        \centering
        \includegraphics[width=\solqualfigwidth]{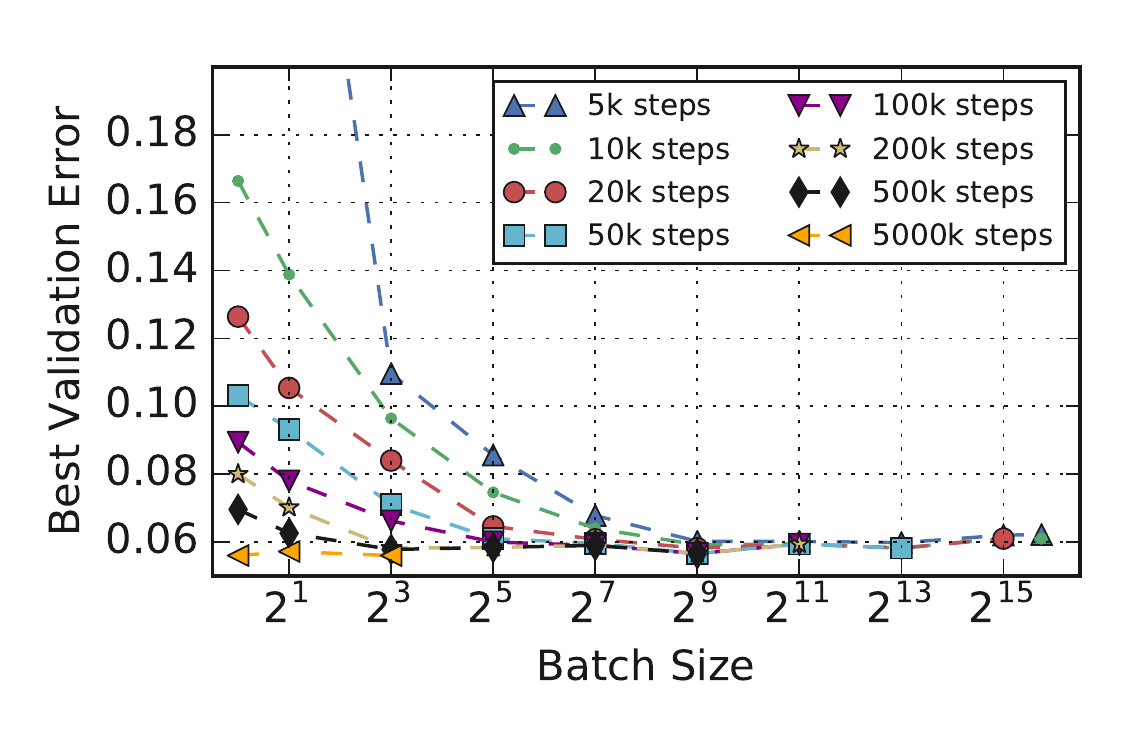}
        \hspace*{\solqualhspace}
        \includegraphics[width=\solqualfigwidth]{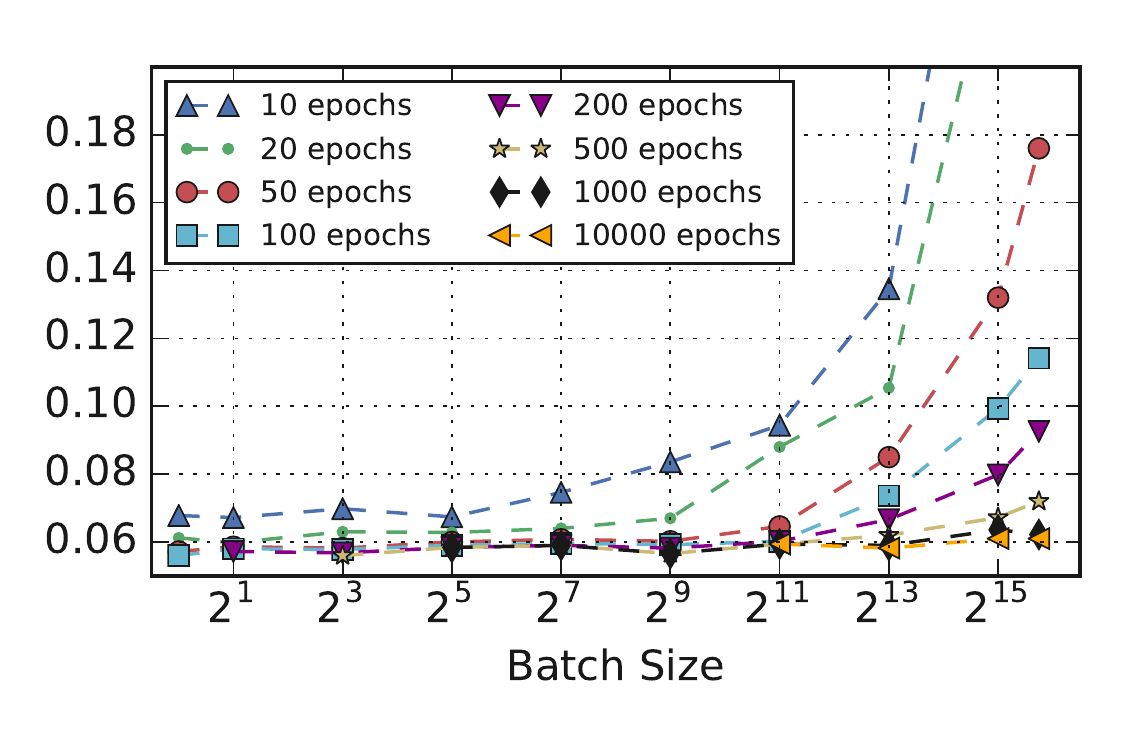}
        \vspace*{\solqualcapshift}
        \caption{\small Simple CNN on Fashion MNIST}\label{fig:sol-qual-fashion-mnist}
    \end{subfigure}\\
    \vspace*{\solquallineshift}
    \begin{subfigure}[b]{\textwidth}
        \centering
        \includegraphics[width=\solqualfigwidth]{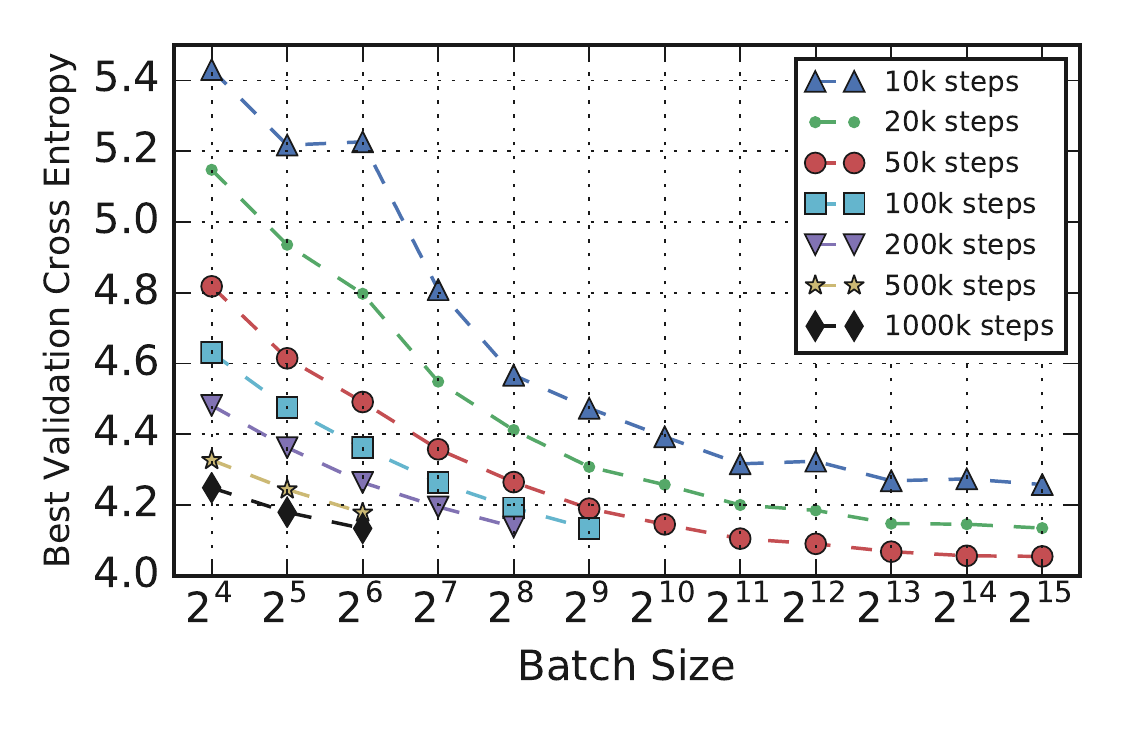}
        \hspace*{\solqualhspace}
        \includegraphics[width=\solqualfigwidth]{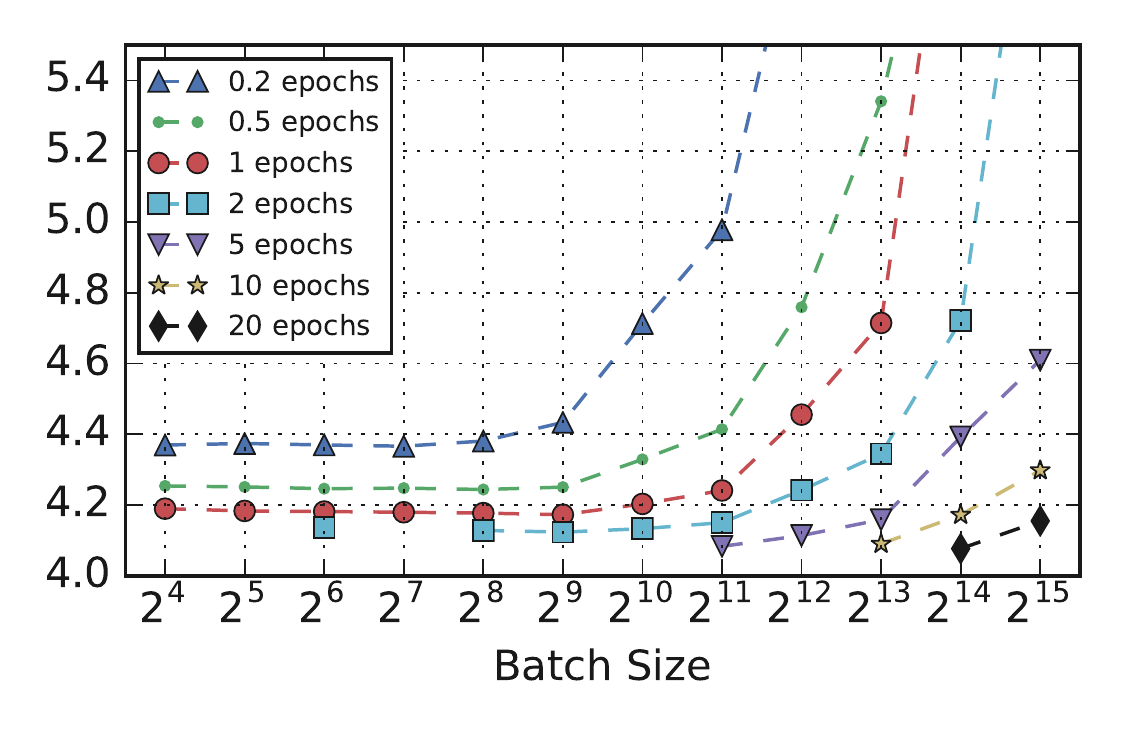}
        \vspace*{\solqualcapshift}
        \caption{\small Transformer (narrow and shallow) on LM1B}\label{fig:sol-qual-transformer-small}
    \end{subfigure}\\
    \vspace*{\solquallineshift}
    \begin{subfigure}[b]{\textwidth}
        \centering
         \includegraphics[width=\solqualfigwidth]{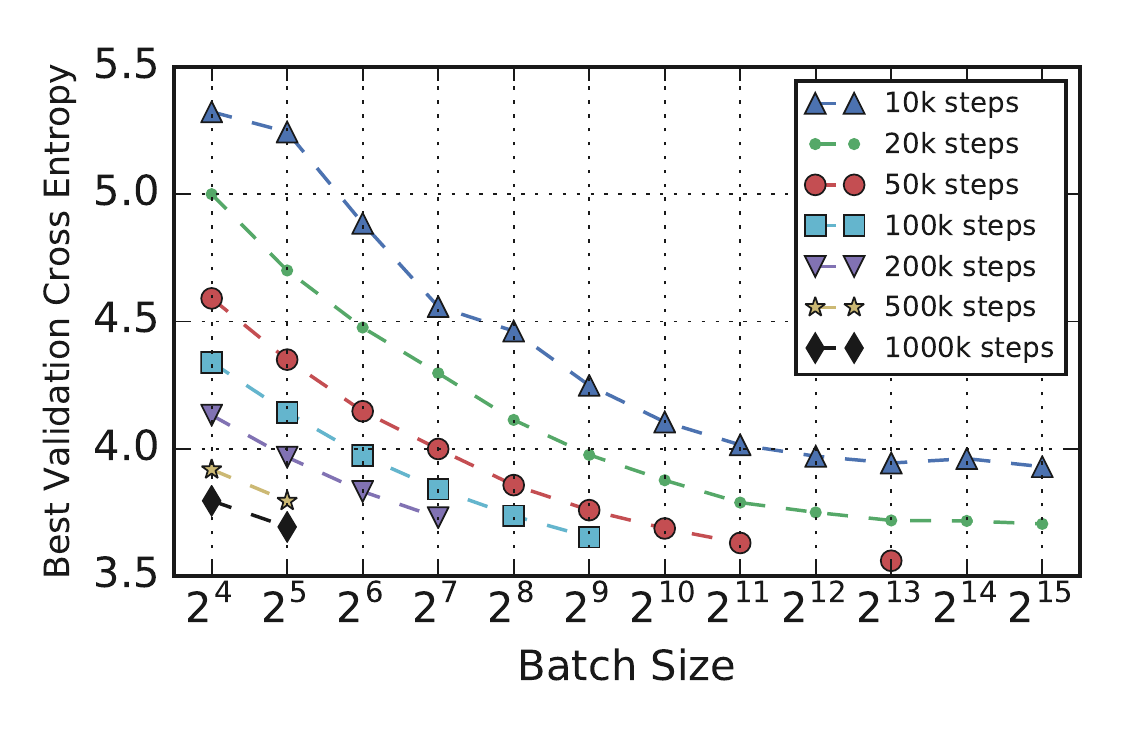}
         \hspace*{\solqualhspace}
         \includegraphics[width=\solqualfigwidth]{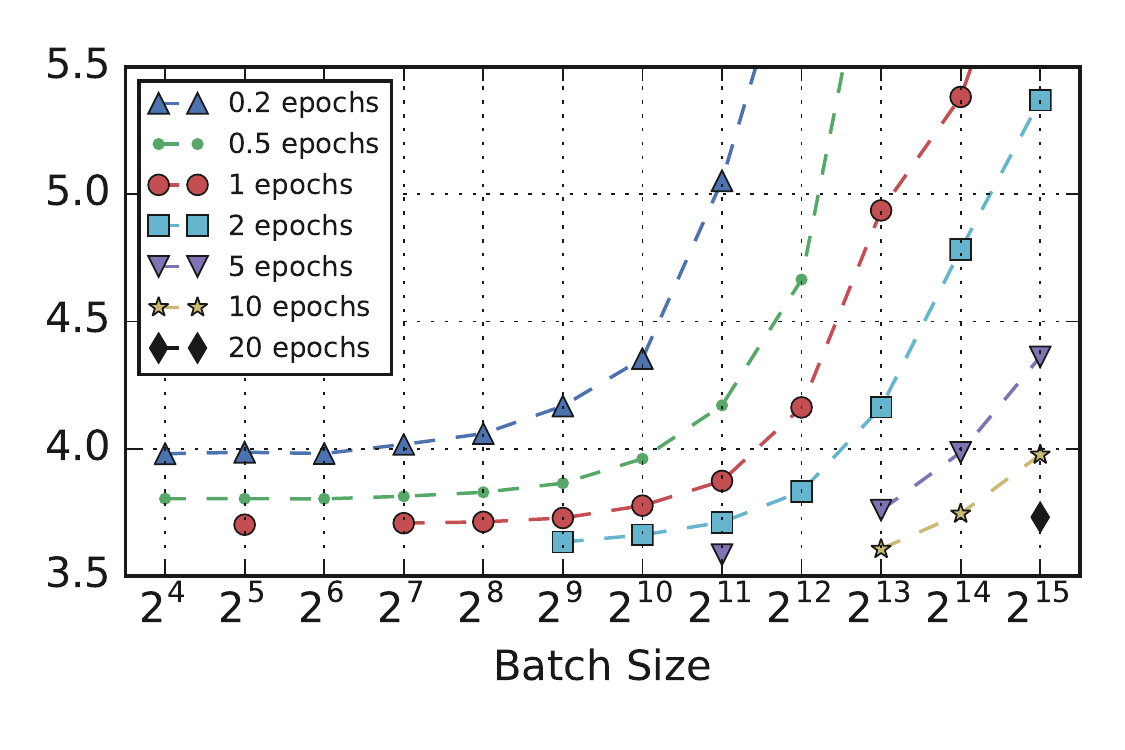}
         \vspace*{\solqualcapshift}
        \caption{\small Transformer (base) on LM1B}\label{fig:sol-qual-transformer-base}
    \end{subfigure}
    \end{tabular}
    \caption{
    \textbf{Validation error depends on compute budget more than batch size.} Plots show the best validation error subject to budgets of training steps (left column) or training epochs (right column). Step budgets favor large batch sizes, while epoch budgets favor small batch sizes.}
    \label{fig:sol-qual}
\end{figure}

We investigated the relationship between batch size and out-of-sample error for Simple CNN on MNIST and Fashion MNIST, and for two sizes of Transformer on LM1B. For each task, we ran a quasi-random metaparameter search over the constant learning rate $\eta$ and Nesterov momentum $\gamma$. For MNIST and Fashion MNIST, we also added label smoothing and searched over the label smoothing parameter in $\{0, 0.1\}$ to mitigate any confounding effects of overfitting (see Section~\ref{sec:label-smoothing}). We ran 100 metaparameter trials for each batch size with a large practical wall-time budget.

To disentangle the effects of the batch size from the compute budget, we compared batch sizes subject to budgets of either training steps or training epochs.
For each batch size and compute budget, we found the model checkpoint that achieved the best validation accuracy across all metaparameter trials, and across all training steps that fell within the compute budget. Figure~\ref{fig:sol-qual} shows the validation error for these best-validation-error checkpoints, as a function of batch size, for a range of compute budgets.
We observe that, subject to a budget on training steps, larger batch sizes achieve better out-of-sample error than smaller batch sizes, but subject to a budget on training epochs, smaller batch sizes achieve better out-of-sample error than larger batch sizes.
These observations are likely explained by the observations that, for a fixed number of training steps, larger batch sizes train on more data, while for a fixed number of epochs, smaller batch sizes perform more training steps. 

The workloads in Figure~\ref{fig:sol-qual} represent two distinct modes of neural network training. For the small MNIST and Fashion MNIST data sets, we used training budgets that would saturate (or almost saturate) performance at each batch size. In other words, out-of-sample error cannot be improved by simply increasing the budget, with caveats due to practical limitations on our ability to find optimal values for the metaparameters. Figures~\ref{fig:sol-qual-mnist} and~\ref{fig:sol-qual-fashion-mnist} show that differences in maximum performance between batch sizes on these data sets are very small (see Figures~\ref{fig:sol-qual-zoom-mnist} and~\ref{fig:sol-qual-zoom-fmnist} in the Appendix for zoomed versions of these plots). We cannot rule out that any differences at this magnitude are due to noise from metaparameter choices and training stochasticity. Thus, for these workloads at least, the effect of batch size on solution quality is either very small or nonexistent.
On the other hand, we cannot saturate performance with Transformer on LM1B within a practical training time. In this case, Figures~\ref{fig:sol-qual-transformer-small} and~\ref{fig:sol-qual-transformer-base} show that the best error is simply achieved by the largest compute budget. 

Taken together, these observations suggest that in practice \textit{the relevant question is not which batch size leads to the best performance, but rather how compute budget varies as a function of batch size.}
Although we tried our best to saturate performance with MNIST and Fashion MNIST, we found that it took millions of training steps for small batch sizes, and thousands of epochs for large batch sizes, even for data sets as small and simple as these. Indeed, despite sampling 100 metaparameter configurations per batch size and training for up to 25 hours per configuration, it is still not certain whether we truly saturated performance at the smallest and largest batch sizes (see Figures~\ref{fig:sol-qual-zoom-mnist} and~\ref{fig:sol-qual-zoom-fmnist} in the Appendix). Thus, the regime of saturated performance is of limited practical concern for most workloads---the compute budget required to saturate performance is likely beyond what a practitioner would typically use. For realistic workloads, practitioners should be most concerned with identifying the batch size at which they can most efficiently apply their compute.

\section{Discussion}
\label{sec:discussion}

Our goals in measuring the effects of data parallelism on neural network training were twofold: first, we hoped to produce actionable advice for practitioners, and second, we hoped to understand the utility of building systems capable of very high degrees of data parallelism. Our results indicate that, for idealized data parallel hardware, there is a universal relationship between training time and batch size, but there is dramatic variation in how well different workloads can make use of larger batch sizes. Across all our experiments, increasing the batch size initially reduced the number of training steps needed proportionally. However, depending on the workload, this perfect scaling regime ended anywhere from a batch size of $2^4$ to a batch size of $2^{13}$. As batch size increases beyond the perfect scaling regime, there are diminishing returns (where increasing the batch size by a factor of $k$ only reduces the number of training steps needed by a factor less than $k$) that end with a maximum useful batch size (where increasing the batch size no longer changes the number of training steps needed). Once again, the maximum useful batch size is extremely problem-dependent and varied between roughly $2^9$ and $2^{16}$ in our experiments. Other workloads may have the region of perfect scaling end at batch sizes even smaller or larger than the range we observed, as well as having even smaller or larger maximum useful batch sizes. 

On the one hand, the possibility that perfect scaling can extend to batch sizes beyond $2^{13}$ for some workloads is good news for practitioners because it suggests that efficient data-parallel systems can provide extremely large speedups for neural network training. On the other hand, the wide variation in scaling behavior across workloads is bad news because any given workload might have a maximum useful batch size well below the limits of our hardware. Moreover, for a new workload, measuring the training steps needed as a function of batch size and confirming the boundaries of the three basic scaling regimes requires expensive experiments. In this work, we have only described how to retrospectively predict the scaling behavior by tuning the optimization metaparameters for every batch size. Although \citet{golmant2018computational} also described the same basic scaling behavior we found, in their experiments the relationship did not appear consistently across problems, across error goals, or in out-of-sample error. In light of our own results, the heuristics they assumed for adjusting the learning rate as a function of batch size are the likely cause of these inconsistencies, but this explanation only drives home the inconvenience of having to carefully tune at every new batch size. We were unable to find reliable support for any of the previously proposed heuristics for adjusting the learning rate as a function of batch size. Thus we are forced to recommend that practitioners tune all optimization parameters anew when they change the batch size or they risk masking the true behavior of the training procedure.

If the scaling behavior of workloads with respect to batch size has a simple dependence on properties of the workload, then we might be able to predict the limits of perfect scaling (or the maximum useful batch size) before running extensive experiments. We could then prioritize workloads to run on specialized hardware or decide whether gaining access to specialized hardware would be useful for a given workload of interest. On the one hand, our results are bad news for practitioners because they show that accurate scaling predictions must depend on a combination of non-obvious properties of the model, optimizer, and data set. On the other hand, we have a lot of control over the choice of model and optimizer and there is some indication that they might be responsible for the largest portion of the variation between workloads. Our results comparing SGD and SGD with momentum (or Nesterov momentum) show that, at least for the problems we tried, momentum can extend perfect scaling to much larger batch sizes, offering clear guidance for practitioners. Other optimizers, such as KFAC \citep{martens2015optimizing,grosse2016kronecker,ba2016distributed}, or optimization techniques designed specifically for massively data parallel systems \citep[e.g.][]{li2014efficient}, might allow perfect scaling to extend much further. Intuitively, it seems plausible that optimizers that estimate local curvature information might be able to benefit more from large batches than optimizers that only use gradients.

Although the model seems to have a large effect on the maximum useful batch size and the limit of perfect scaling, our results do not give definitive answers on exactly how to design models that scale better for a given optimizer and data set. Even when we kept the model family fixed, we observed somewhat inconsistent results from changing the model width and depth. \citet{chen2018effect} suggested that wider models can exploit larger batch sizes than narrower models, but their theoretical arguments only apply to linear networks and fully connected networks with a single hidden layer. In contrast, we found that \textit{narrower} variants of the Transformer model scaled better to larger batch sizes, although it is unclear if the same notion of ``width'' transfers between different types of neural networks.

Unlike the model and optimizer, we generally have much less control over the data set. Unfortunately, properties of the data set also affect how well training scales in practice. Our results are equivocal on whether the number of training examples has any effect, but changing the data set entirely can certainly change the scaling behavior with respect to batch size. 

Finally, our results at least partially reconcile conflicting stances in the literature on whether increasing the batch size degrades model quality. Our experiments show that:
\begin{enumerate}
    \item Any study that only tunes the learning rate for one batch size and then uses a heuristic to choose the learning rate for other batch sizes \citep{goyal2017accurate,keskar2016large,hoffer2017train,lin2018don,devarakonda2017adabatch,golmant2018computational} gives a systematic advantage to the batch size used in tuning (as well as nearby batch sizes). Our results did not show a simple relationship between the optimal learning rate and batch size that scales indefinitely (see Figures~\ref{fig:effective-lr} and~\ref{fig:raw-lr}), so the use of simple heuristics for batch sizes sufficiently far from the base batch size could very well explain the degraded solutions and divergent training reported in prior work. Similarly, the optimal values of other metaparameters, such as the momentum and learning rate decay schedule, should not be assumed to remain constant or scale in a simple way as the batch size increases.

    \item Assuming an epoch budget when comparing solution quality between batch sizes \citep{masters2018revisiting,goyal2017accurate,lin2018don,devarakonda2017adabatch}, in effect, limits an investigation to the perfect scaling region of the steps to result vs batch size curve (see Figure~\ref{fig:stt-problems}). This budget favors smaller batch sizes because they will perform more optimizer steps for the same number of training examples (see Section~\ref{sec:solution-quality}). Certainly, there are situations where an epoch budget is appropriate, but there may exist budgets just outside the perfect scaling region that can achieve the same quality solution, and those budgets may still represent a significant reduction in the number of training steps required.
    Moreover, even for a fixed model and data set, simply changing the optimizer can significantly extend the perfect scaling regime to larger batch sizes. For example, \citet{masters2018revisiting} found that test performance of ResNet-8 (without batch normalization) on CIFAR-10 with a fixed epoch budget degraded after batch size 16, but considered only plain mini-batch SGD. Our experiments confirmed that perfect scaling ends at batch size 16 with plain mini-batch SGD, but using Nesterov momentum extends the perfect scaling regime to batch size 256 (see Figure~\ref{fig:stt-problems-resnet-cifar}).
    
    \item Assuming a step budget when comparing solution quality between batch sizes \citep{hoffer2017train} might favor larger batch sizes because they will see more training examples for the same number of gradient updates (see Section~\ref{sec:solution-quality}). A step budget is likely sufficient for a larger batch size to reach \emph{at least} the same performance as a smaller batch size: we never saw the number of steps to reach a goal validation error increase when the batch size was increased (see Figure~\ref{fig:stt-problems}).
    
    \item Increasing the batch size reduces noise in the gradient estimates (see Equation~\ref{eq:batch-stoch-grad}). However, the noise in updates due to small batches might, in some cases, provide a helpful regularization effect \citep{GoodfellowEtAlBook2016,smith2018bayesian}. Thankfully, other regularization techniques, such as label smoothing, can replace this effect (see Section~\ref{sec:label-smoothing}). Others have also used regularization techniques, such as data augmentation \citep{keskar2016large} and $L_2$~regularization \citep{smith2018bayesian}, to eliminate the ``generalization gap'' between two batch sizes.
    
    \item Finally, although we do not believe there is an inherent degradation in solution quality associated with increasing the batch size, depending on the compute budget, it may become increasingly difficult to find good values for the metaparameters with larger batch sizes. Specifically, increasing the batch size may shrink the region in metaparameter space corresponding to rapid training in terms of epochs (see Figure~\ref{fig:hparams-sensitivity-transformer-epoch-budget}), as previously reported by \citet{breuel2015effects}.
    On the other hand, increasing the batch size may increase the region in metaparameter space corresponding to rapid training in terms of steps (see Figure~\ref{fig:hparams-sensitivity-transformer-step-budget}).
\end{enumerate}

\subsection{Limitations of our experimental protocol}

When interpreting our results, one should keep in mind any limitations of our experimental protocol. We do not believe any of these limitations are debilitating, and we hope that describing these potential areas of concern will spur methodological innovation in future work.

Firstly, we were unable to avoid some amount of human judgment when tuning metaparameters. Although we did not tune metaparameters by hand, we specified the search spaces for automatic tuning by hand and they may not have been equally appropriate for all batch sizes, despite our best efforts.
We are most confident in our search spaces that tuned the fewest metaparameters (such as in our experiments that only tuned learning rate and momentum). We found it quite difficult to be confident that our tuning was sufficient when we searched over learning rate decay schedules; readers should be aware that the steps to result measurement is generally quite sensitive to the learning rate schedule. Thus, we may not have sampled enough trials at some batch sizes or, nearly equivalently, our search spaces may have been too wide at some batch sizes. Even though we verified that the best trial was not on the boundary of the search space, this by no means guarantees that we found the globally optimal metaparameters.

Smaller batch sizes typically had more opportunities to measure validation error and, when validation error was noisy, got more chances to sample a lucky validation error. Batch sizes (usually larger ones) that did not reach the goal validation error using the first search space used revised search spaces that gave them an extra bite of the apple, so to speak.

Finally, our analysis does not consider how robustly we can reach a goal error rate. For instance, we did not distinguish between batch sizes where all 100 trials achieved the goal validation error and batch sizes where only one of the 100 trials achieved the goal.
The maximum or minimum value over a set of trials is not usually a very robust statistic, but something like the 50\nth\ percentile trial mostly reveals information about the search space. We tried to strike a balance between studying realistic workloads and being able to repeat our experiments so many times that these uncertainty questions became trivial. Ultimately, we opted to study realistic workloads and simply report results for the optimal trials.

\section{Conclusions and Future Work}

Increasing the batch size is a simple way to produce valuable speedups across a range of workloads, but, for all workloads we tried, the benefits diminished well within the limits of current hardware. Unfortunately, blindly increasing the batch size to the hardware limit will not produce a large speedup for all workloads. However, our results suggest that some optimization algorithms may be able to consistently extend perfect scaling across many models and data sets. Future work should perform our same measurements with other optimizers, beyond the closely-related ones we tried, to see if any existing optimizer extends perfect scaling across many problems.
Alternatively, if we only need speedups for specific, high-value problems, we could also consider designing models that extend perfect scaling to much larger batch sizes.
However, unlike the optimizer, practitioners are likely to tailor their model architectures to the specific problems at hand. Therefore, instead of searching for model architectures that happen to scale extremely well, future work should try to uncover general principles for designing models that can scale perfectly to larger batch sizes. Even if such principles remain elusive, we would still benefit from methods to prospectively predict the scaling behavior of a given workload without requiring careful metaparameter tuning at several different batch sizes. 
Finally, the deep learning community can always benefit from methodical experiments designed to test hypotheses, characterize phenomena, and reduce confusion, to balance more exploratory work designed to generate new ideas for algorithms and models.

\subsection*{Acknowledgements}

We thank Tomer Koren for helpful discussions. We also thank Justin Gilmer and Simon Kornblith for helpful suggestions and comments on the manuscript. Finally, we thank Matt J.~Johnson for lending us some computing resources.

\appendix

\section{Data Set Details}\label{appendix:datasets}

This section contains details of the data sets summarized in Table~\ref{table:datasets}.

\subsection{Data Set Descriptions and Pre-Processing}

\textbf{MNIST} \citep{lecun1998mnist} is a classic handwritten digit image classification data set with 10 mutually exclusive classes. We split the original training set into 55,000 training images and 5,000 validation images, and used the official test set of 10,000 images. We did not use data augmentation.

\textbf{Fashion MNIST} \citep{xiao2017fashion} is another reasonably simple image classification data set with 10 mutually exclusive classes. It was designed as a drop-in replacement for MNIST. We split the original training set into 55,000 training images and 5,000 validation images, and used the official test set of 10,000 images. We did not use data augmentation.

\textbf{CIFAR-10} \citep{krizhevsky2009learning} is an image classification data set of $32 \times 32$ color images with 10 mutually exclusive classes. We split the original training set into 45,000 training images and 5,000 validation images. We used the official test set of 10,000 images. We pre-processed each image by subtracting the average value across all pixels and channels and dividing by the standard deviation.\footnote{We used the TensorFlow op 
\texttt{tf.image.per\_image\_standardization}.} We did not use data augmentation.

\textbf{ImageNet} \citep{russakovsky2015imagenet} is an image classification data set with 1,000 mutually exclusive classes. We split the official training set into 1,281,167 training images and 50,045 test images, and used the official validation set of 50,000 images. We pre-processed the images and performed data augmentation in a similar way to \citet{simonyan2014very}. Specifically, at training time, we sampled a random integer $S \in [256, 512]$, performed an aspect-preserving resize so that the smallest side had length $S$, and took a random crop of size $(224, 224)$. We randomly reflected the images horizonally, but unlike \citet{simonyan2014very}, we did not distort the colors. At evaluation time, we performed an aspect-preserving resize so that the smallest side had length 256, and took a central crop of size $(224, 224)$. In both training and evaluation, we then subtracted the global mean RGB value from each pixel using the values computed by \citet{simonyan2014very}.\footnote{See \url{https://gist.github.com/ksimonyan/211839e770f7b538e2d8\#description} for the mean RGB values used.}

\textbf{Open Images v4} \citep{openimages} is a data set of 9 million images that are annotated with image-level labels and object bounding boxes.\footnote{Available at \url{https://storage.googleapis.com/openimages/web/index.html}.} The image labels were generated by a computer vision model and then verified as either \textit{positive} or \textit{negative} labels by human annotators. We only considered the 7,186 ``trainable'' classes with at least 100 human-annotated positives in the training set. We filtered the official subsets by keeping only images with at least one positive trainable label, which produced training, validation and test sets of size 4,526,492; 41,225; and 124,293 images, respectively. On average, each image in the training set has 2.9 human-annotated positive labels, while each image in the validation and test sets have 8.4 human-annotated positive labels. We only considered the human-annotated positives and assumed all other classes were negative. We pre-processed the images and performed data augmentation identically to ImageNet.

\textbf{LM1B} \citep{chelba2014one}
is a text data set of English news articles.\footnote{Available at \url{http://www.statmt.org/lm-benchmark/}.} We used the official training set and created validation and test sets using files
\texttt{news.en.heldout\-00000-of-00050} and \texttt{news.en.heldout-00001-of-00050},
respectively. These splits contain 30,301,028; 6,075; and 6,206 sentences, respectively. We used an invertable word tokenizer to split the text into sub-word tokens with a vocabulary of size 32,000.\footnote{The code for processing the raw data and generating the vocabulary is available at \url{https://github.com/tensorflow/tensor2tensor/blob/master/tensor2tensor/data_generators/lm1b.py}} On average, the training set contains around 20 tokens per sentence and the validation and test sets contain around 29 tokens per sentence. At training time, we clipped long sentences to the first 64 tokens, which affected only about 2\% of sentences. We did not clip long sentences at evaluation time. The maximum sentence across the validation and test sets has 476 tokens.

\textbf{Common Crawl} is a repository of web data containing over 3 billion web pages.\footnote{Available at \url{http://commoncrawl.org/2017/07/june-2017-crawl-archive-now-available/}.} We filtered and processed the data set identically to \citet{anil2018large}.\footnote{See \url{https://github.com/google-research/google-research/tree/master/codistillation} for document IDs.} The vocabulary contains 24,006 sub-word tokens. We randomly partitioned the sentences into a training set (99.98\%) and a holdout set (0.02\%). Our training set contains ${\sim}25.8$ billion sentences. We used the first 6,075 sentences of the holdout set as our validation set, which is the same number of sentences in our LM1B validation set. Some sentences are tens of thousands of tokens long. To maintain consistency with our LM1B processing, we clipped sentences to 64 tokens at training time and 476 at evaluation time. 

\subsection{Evaluation Metrics}\label{appendix:metrics}

We use \textbf{classification error} for MNIST, Fashion MNIST, CIFAR-10, and ImageNet. To compute this metric, we consider the model's classification for each image to be the class it assigns the highest probability. Then
\begin{equation*}
\text{classification error} = \frac{\text{\# incorrect classifications}}{\text{\# classifications}}.
\end{equation*}

We use class-agnostic \textbf{average precision} ($\mathit{AP}$) for Open Images. To compute this metric, we first rank each image-class pair by the predicted likelihood of the class being a true positive for that image. Then
\begin{equation}\label{eq:averageprecision}
    \mathit{AP} = \frac{1}{w} \sum_{k=1}^{nm} \text{Precision}(k) \cdot \text{Relevance}(k),
\end{equation}
where $\text{Precision}(k)$ is the precision when considering the top $k$ image-class pairs, $\text{Relevance}(k)$ is an indicator function equal to 1 if the $k$\nth\ image-class pair is a verified positive and 0 otherwise, $n$ is the number of images in the validation set, $m$ is the number of classes, and $w$ is the number of positive labels. Average precision was proposed for Open Images by \citet{veit2017learning}. Due to false negatives in the validation set, \citet{veit2017learning} only computed $\mathit{AP}$ over the the human-annotated classes in each image. However, on average, each image in the validation set only has 8.4 positive and 4 negative human-annotated classes, so each image is only evaluated over ${\sim}12$ classes out of 7,186. This yields misleadingly high values of $\mathit{AP}$. Instead, we compute $\mathit{AP}$ over all classes in each image, which may underestimate the true $\mathit{AP}$ due to false negatives in the validation set, but is more indicative of the true performance in our experience. We compute $\mathit{AP}$ using an efficient approximation of the area under the discrete precision-recall curve.\footnote{Equation~\ref{eq:averageprecision} can be interpreted as a right Riemann sum of the discrete precision-recall curve $\{ (r_i, p_i) | i = 1, ..., w \}$, where $r_i = i/w$ and $p_i$ is the maximum precision among all values of precision with recall $r_i$ (each value of recall may correspond to different values of precision at different classification thresholds). We use the TensorFlow op \texttt{tf.metrics.auc} with \texttt{curve="PR"}, \texttt{num\_thresholds=200}, and \texttt{summation\_method="careful\_interpolation"}.}

We use average per-token \textbf{cross entropy error} for LM1B and Common Crawl. For a single sentence $s=(w_1,...,w_m)$, let $p(w_j | w_1,...,w_{j-1})$ denote the model's predicted probability of the token $w_j$ given all prior tokens in the sentence. Thus, the predicted log-probability of $s$ is $\log p(s) = \sum_{j=1}^m \log p(w_j | w_1,...,w_{j-1})$. We compute the average per-token cross entropy error over a data set $\{ s_1, ..., s_n \}$ as
\begin{equation*}
    \text{cross entropy error} = \frac{\sum_{i=1}^n\log p(s_n)}{\sum_{i=1}^n \text{len}(s_n)},
\end{equation*}
where $\text{len}(s)$ denotes the number of tokens in $s$. This is the logarithm of the per-token perplexity.

\section{Model Details}\label{appendix:models}

In this section we give the architectural details of the models summarized in Table~\ref{table:models}. In addition to the descriptions below, each model has a task-specific output layer. Models trained on MNIST, Fashion MNIST, CIFAR-10, and ImageNet (classification with mutually exclusive labels) use a softmax output layer to model the probability distribution over classes. Models trained on Open Images (classification with multiple labels per image) use a sigmoid output layer to model the probability of each class. Models trained on LM1B and Common Crawl (language modeling) use a softmax output layer to model the probability of the next word in a sentence given all prior words in the sentence.

\textbf{Fully Connected} is a fully connected neural network with ReLU activation function. Hidden layers use dropout with probability 0.4 during training. We vary the number of layers and number of units per layer in different experiments to investigate the impact of model size. We use the notation FC-$N_1$-...-$N_k$ to denote a fully connected neural network with $k$ hidden layers and $N_i$ units in the $i$\nth\ layer.

\textbf{Simple CNN} consists of 2 convolutional layers with max-pooling followed by 1 fully connected hidden layer. The convolutional layers use $5 \times 5$ filters with stride length 1, ``same'' padding \citep{GoodfellowEtAlBook2016}, and ReLU activation function. Max pooling uses $2 \times 2$ windows with stride length 2. The fully connected layer uses dropout with probability 0.4 during training. We used three different model sizes: \textbf{base} has 32 and 64 filters in the convolutional layers and 1,024 units in the fully connected layer; \textbf{narrow} has 16 and 32 filters in the convolutional layers and 512 units in the fully connected layer; and \textbf{wide} has 64 and 128 filters in the convolutional layers and 2,048 units in the fully connected layer. We used the base model unless otherwise specified.

\textbf{ResNet-8} consists of 7 convolutional layers with residual connections followed by 1 fully connected hidden layer. We used the model described in section~4.2 of \citet{he2016deep} with $n=1$, but with the improved residual block described by \citet{he2016identity}. We removed batch normalization, which is consistent with \citet{masters2018revisiting}.

\textbf{ResNet-50} consists of 49 convolutional layers with residual connections followed by 1 fully connected hidden layer. We used the model described in section~4.1 of \citet{he2016deep}, but with the improved residual block described by \citep{he2016identity}. We replaced batch normalization \citep{ioffe2015batch} with ghost batch normalization to keep the training objective fixed between batch sizes and to avoid possible negative effects from computing batch normalization statistics over a large number of examples \citep{hoffer2017train}. We used a ghost batch size of 32 for all experiments. We also applied label smoothing \citep{szegedy2016rethinking} to regularize the model at training time, which was helpful for larger batch sizes. The label smoothing coefficient was a metaparameter that we tuned in our experiments.

\textbf{VGG-11} consists of 8 convolutional layers followed by 3 fully connected hidden layers. We used the model referred to as ``model A'' by \citet{simonyan2014very}.

\textbf{LSTM} is a one hidden-layer LSTM model \citep{hochreiter1997long}. It is a simpler variant of the LSTM-2048-512 model described by \citet{jozefowicz2016exploring}, with 1,024 embedding dimensions, 2,048 hidden units, and 512 projection dimensions. We did not use bias parameters in the output layer because we found this improved performance in our preliminary experiments.

\textbf{Transformer} is a self-attention model that was originally presented for machine translation \citep{vaswani2017attention}. We used it as an autoregressive language model by applying the decoder directly to the sequence of word embeddings for each sentence. We used four different sizes: the \textbf{base} model described by \citet{vaswani2017attention}; a \textbf{shallow} model that is identical to the base model except with only two hidden layers instead of six; a \textbf{narrow and shallow} model that is identical to the shallow model except with half as many hidden units and attention heads as well as half the filter size; and a \textbf{wide} model that is identical to the base model except with double the number of hidden units and attention heads as well as double the filter size. We used the base model unless otherwise specified.

\section{Learning Rate Schedules}\label{appendix:lr_decay}

We chose our learning rate schedule by experimenting with a variety of different schedules for ResNet-50 on ImageNet. For each schedule, we specified the following metaparameters:
\begin{itemize}
    \item $\eta_0$: initial learning rate
    \item $\alpha$: decay factor ($\alpha > 0$)
    \item $T$: number of training steps until the learning rate decays from $\eta_0$ to $\alpha \eta_0$
\end{itemize}

Each schedule corresponds to a decay function $d(t)$, such that the learning rate at training step $t$ is
\begin{equation*}
    \eta(t) = \begin{cases}
      d(t) \cdot \eta_0 & \text{if } t \le T, \\
      \alpha \eta_0 & \text{if } t > T.
    \end{cases}
\end{equation*}

We experimented with the following decay functions:
\begin{itemize}
    \item \textbf{Constant}: $d(t) = 1$
    \item \textbf{Linear}: $d(t) = 1 - (1- \alpha) \frac{t}{T}$
    \item \textbf{Cosine} \citep{loshchilov2016sgdr}: $d(t) =  \alpha + \frac{(1-\alpha)}{2} \left( 1 + \cos{\pi \frac{t}{T}} \right)$
    \item \textbf{Exponential Polynomial}: $d(t) =  \alpha + (1-\alpha) \left( 1 - \frac{t}{T} \right)^\lambda$, where $\lambda > 0$
    \item \textbf{Inverse Exponential Polynomial}: $d(t) = \frac{\alpha}{\alpha + (1-\alpha) \left( \frac{t}{T} \right)^\lambda}$, where $\lambda > 0$
    \item \textbf{Exponential}: $d(t) = \alpha^{t/T}$
\end{itemize}

We also tried piecewise linear learning rate schedules. These schedules are specified by a sequence of pairs $\{ (t_0, \eta_0), ..., (t_k, \eta_k) \}$, with $0 = t_0 < t_1 ... < t_k$, such that the learning rate at training step $t$ is 
\begin{equation*}
    \eta(t) = \begin{cases}
      \eta_i + \frac{\eta_{i+1} - \eta_i}{t_{i+1} - t_i} (t - t_i) & \text{if } t_i \le t < t_{i+1}, \\
      \eta_k & \text{if } t \ge t_k.
    \end{cases}
\end{equation*}
The schedules used by both \citet{he2016deep} (piecewise constant) and \citet{goyal2017accurate} (linear warm-up followed by piecewise constant) for ResNet-50 on ImageNet can both be expressed as piecewise linear.

We ran experiments with ResNet-50 on ImageNet, using Nesterov momentum with batch size 1,024 for 150,000 training steps, while tuning the momentum and all metaparameters governing the learning rate schedule. We used quasi-random metaparameter search as described in Section~\ref{sec:methods_and_experiments}. For piecewise linear schedules, we tried 1, 3, and 5 decay events. We found that it was possible to get good results with several of the schedules we tried, and it is likely that other schedules would also work well. Ultimately, we chose linear decay because it performed at least well as all other schedules we tried, while also being the simplest and requiring only two additional metaparameters.

\vspace*{\fill}
\section{Additional Plots}\label{appendix:additional-plots}

\begin{figure}[H]
    \centering
    \begin{subfigure}[b]{\threecolfigwidth}
        \includegraphics[width=\textwidth]{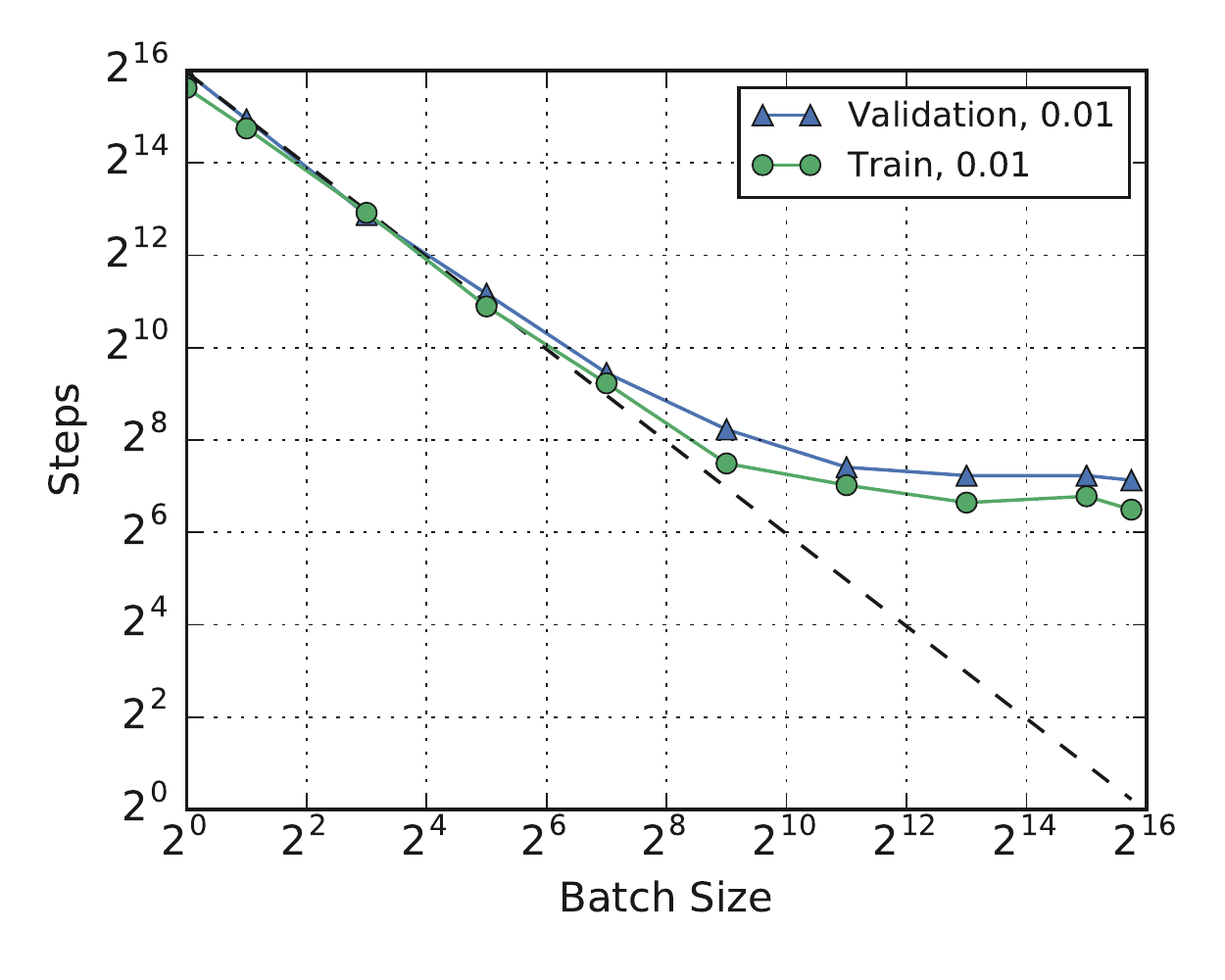}
        \vspace*{\capshift}
        \caption{Simple CNN on MNIST}
    \end{subfigure}
    \begin{subfigure}[b]{\threecolfigwidth}
        \includegraphics[width=\textwidth]{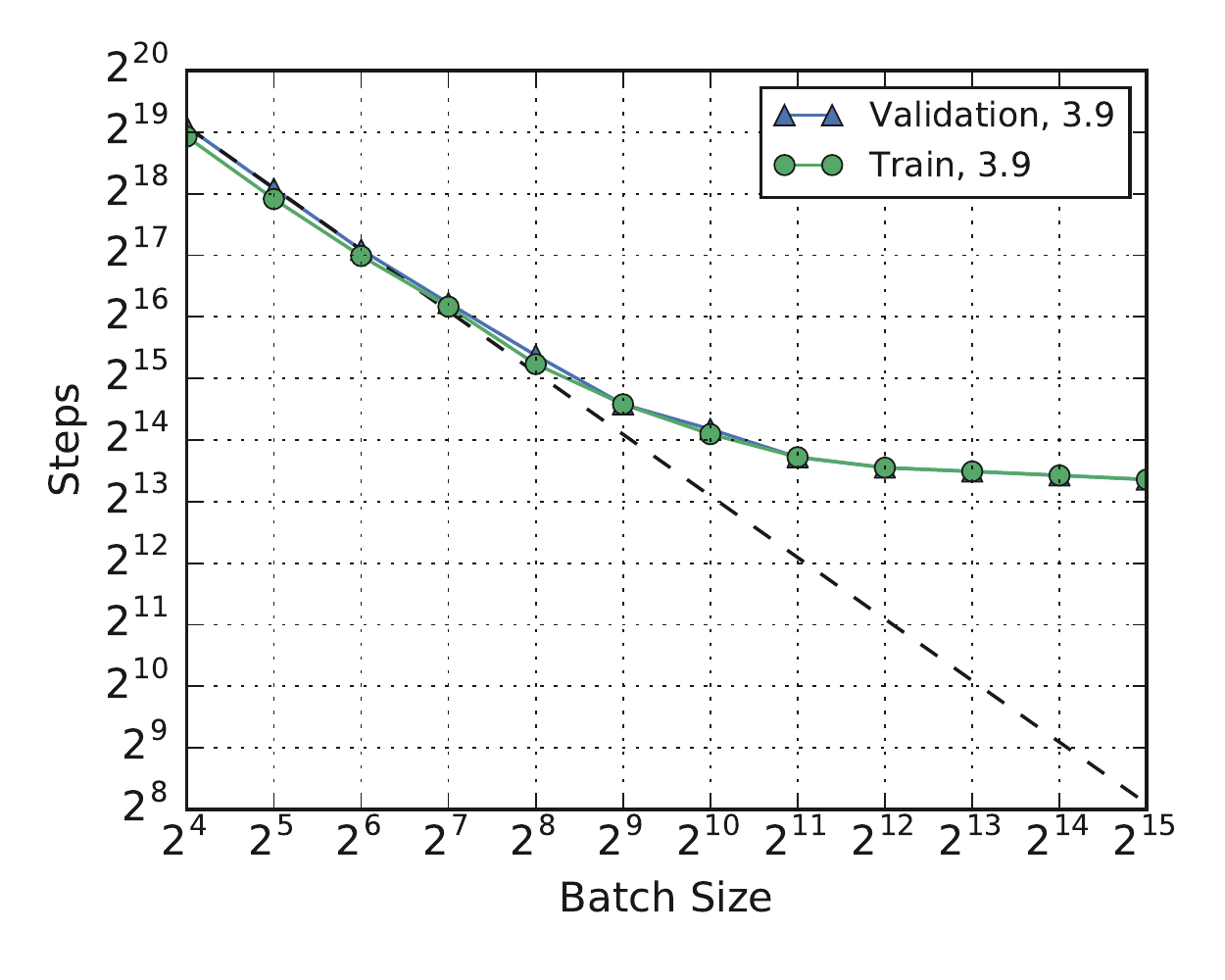}
        \vspace*{\capshift}
        \caption{Transformer on LM1B}
    \end{subfigure}
    \begin{subfigure}[b]{\threecolfigwidth}
        \includegraphics[width=\textwidth]{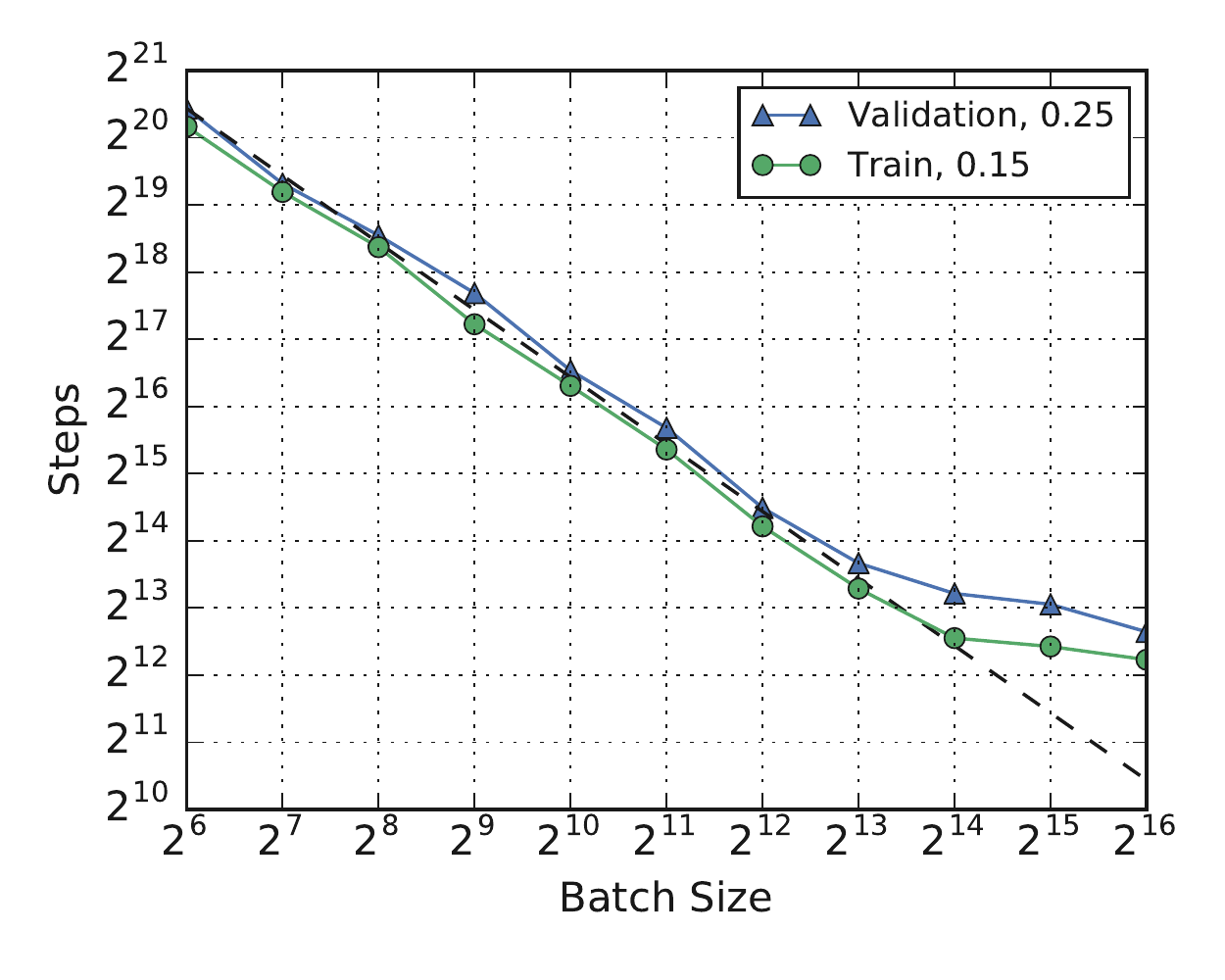}
        \vspace*{\capshift}
        \caption{ResNet-50 on ImageNet}
    \end{subfigure}
    \caption{{\bf Steps to result on the training set is almost the same as on the validation set.} The evaluation metrics are described in Appendix~\ref{appendix:metrics}. Error goals are specified in the plot legends.}
    \label{fig:stt-train-valid}
\end{figure}

\begin{figure}[H]
    \centering
    \includegraphics[width=0.9\textwidth]{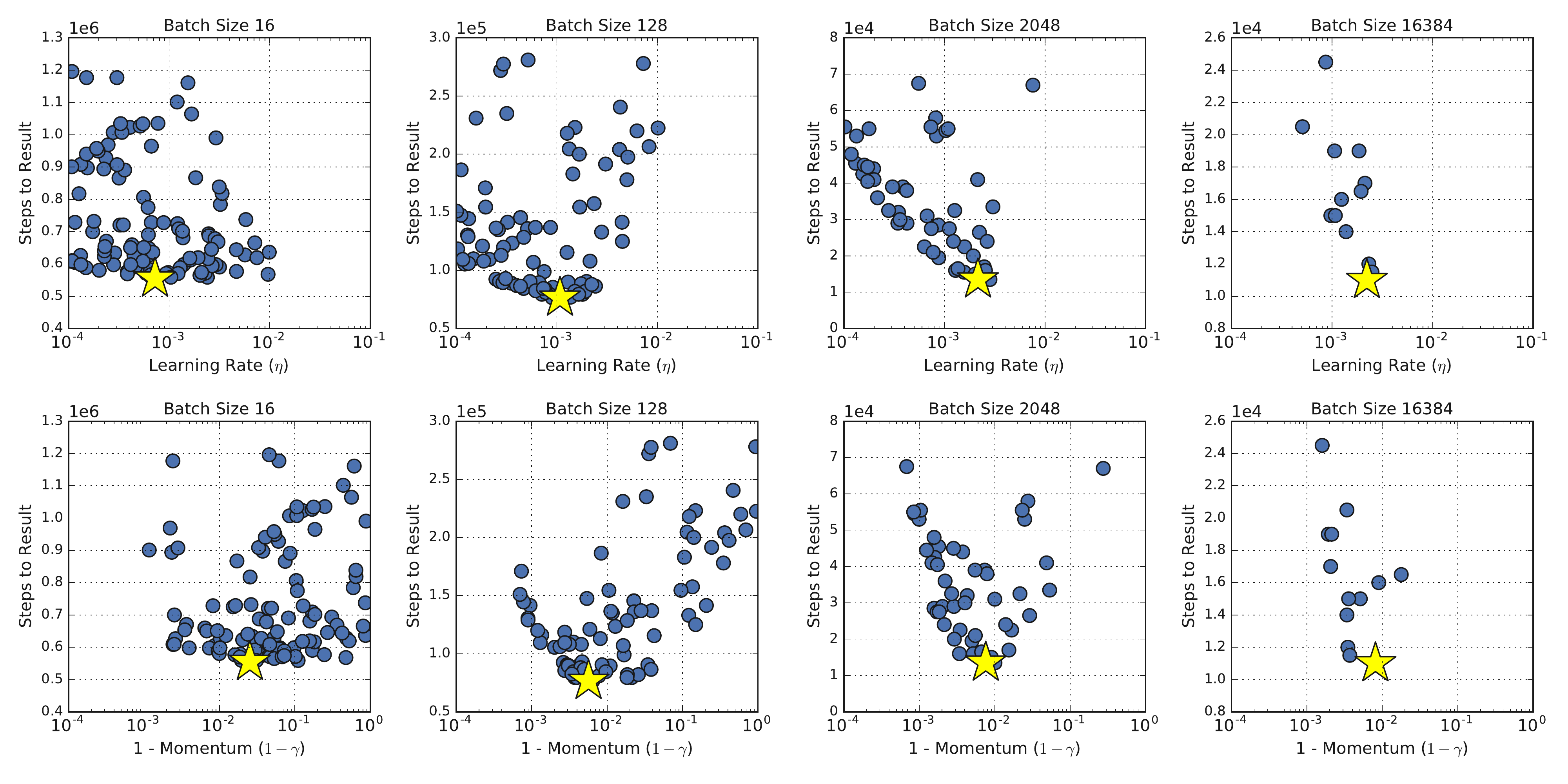}
    \caption{
    \textbf{Validating metaparameter search spaces for Transformer on LM1B.} Rows correspond to the metaparameters we tuned (learning rate $\eta$ and momentum $\gamma$) and columns correspond to different batch sizes. The $x$-axis is the search range that was sampled by the quasi-random search algorithm. Blue dots represent trials that reached the goal of 3.9 validation cross entropy error, and yellow stars correspond to trials that achieved the goal in the fewest steps. We deem these search spaces appropriate because the yellow stars are not on the boundaries.}
    \label{fig:hparams-validation-transformer}
\end{figure}

\begin{figure}[H]
    \centering
    \includegraphics[width=0.95\textwidth]{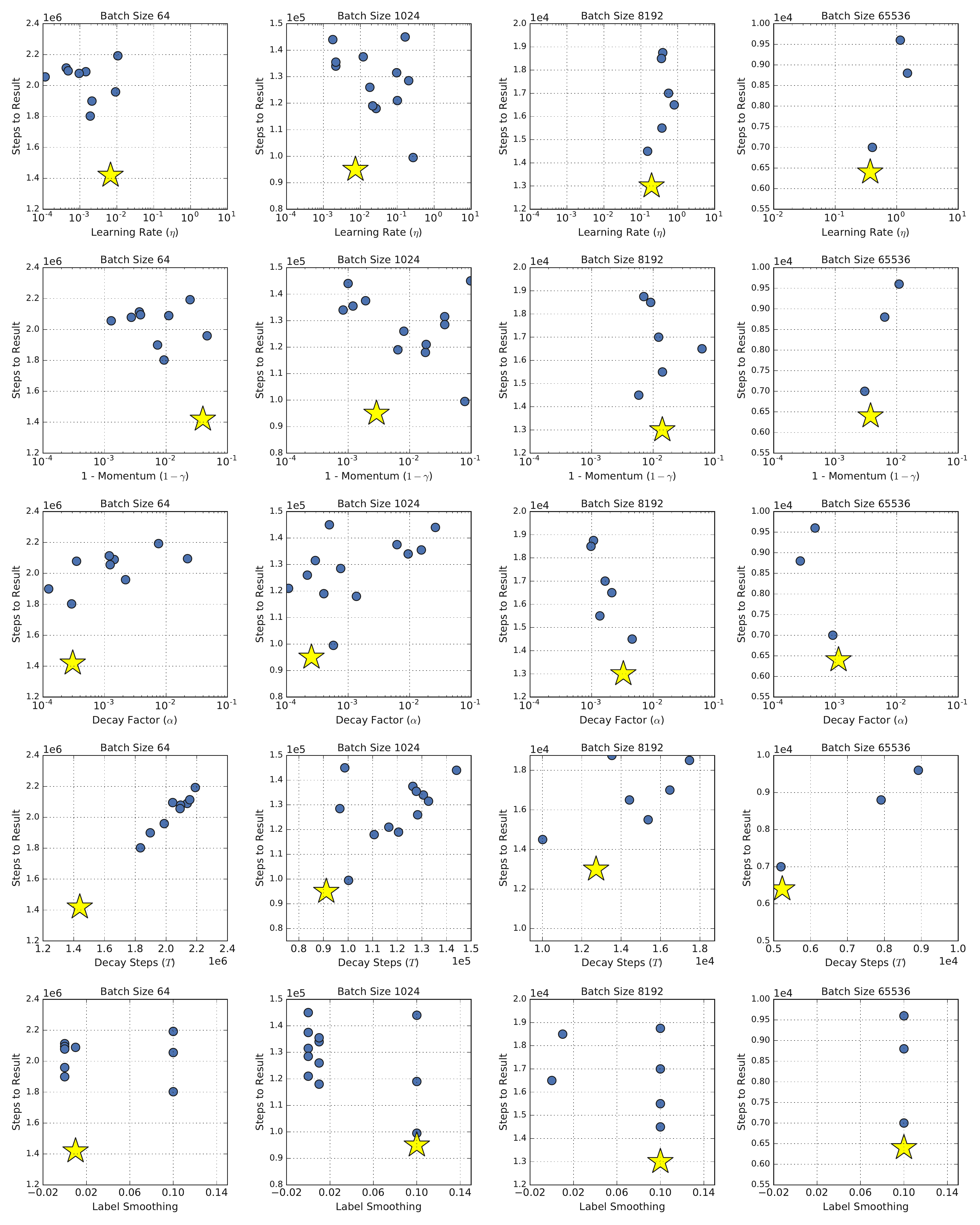}
    \caption{
    \textbf{Validating metaparameter search spaces for ResNet-50 on ImageNet.} Rows correspond to the metaparameters we tuned (initial learning rate $\eta_0$, momentum $\gamma$, learning rate decay parameters $\alpha$, $T$, and label smoothing parameter) and columns correspond to different batch sizes. For all parameters except the label smoothing parameter, the $x$-axis is the search range sampled by the quasi-random search algorithm. The label smoothing parameter was sampled uniformly in $\{0, 0.01, 0.1\}$ for $b\leq 2^{14}$ and $\{ 0, 0.1 \}$ for $b> 2^{14}$. Blue dots represent trials that reached the goal validation error rate of 0.25, and yellow stars correspond to trials that achieved the goal in the fewest steps. We deem these search spaces appropriate because the yellow stars are not on the boundaries.}
    \label{fig:hparams-validation-resnet}
\end{figure}

\newcommand{\multimodelfigwidth}{0.49\textwidth}
\begin{figure}[H]
    \centering
    \begin{subfigure}[b]{\multimodelfigwidth}
        \includegraphics[width=\textwidth]{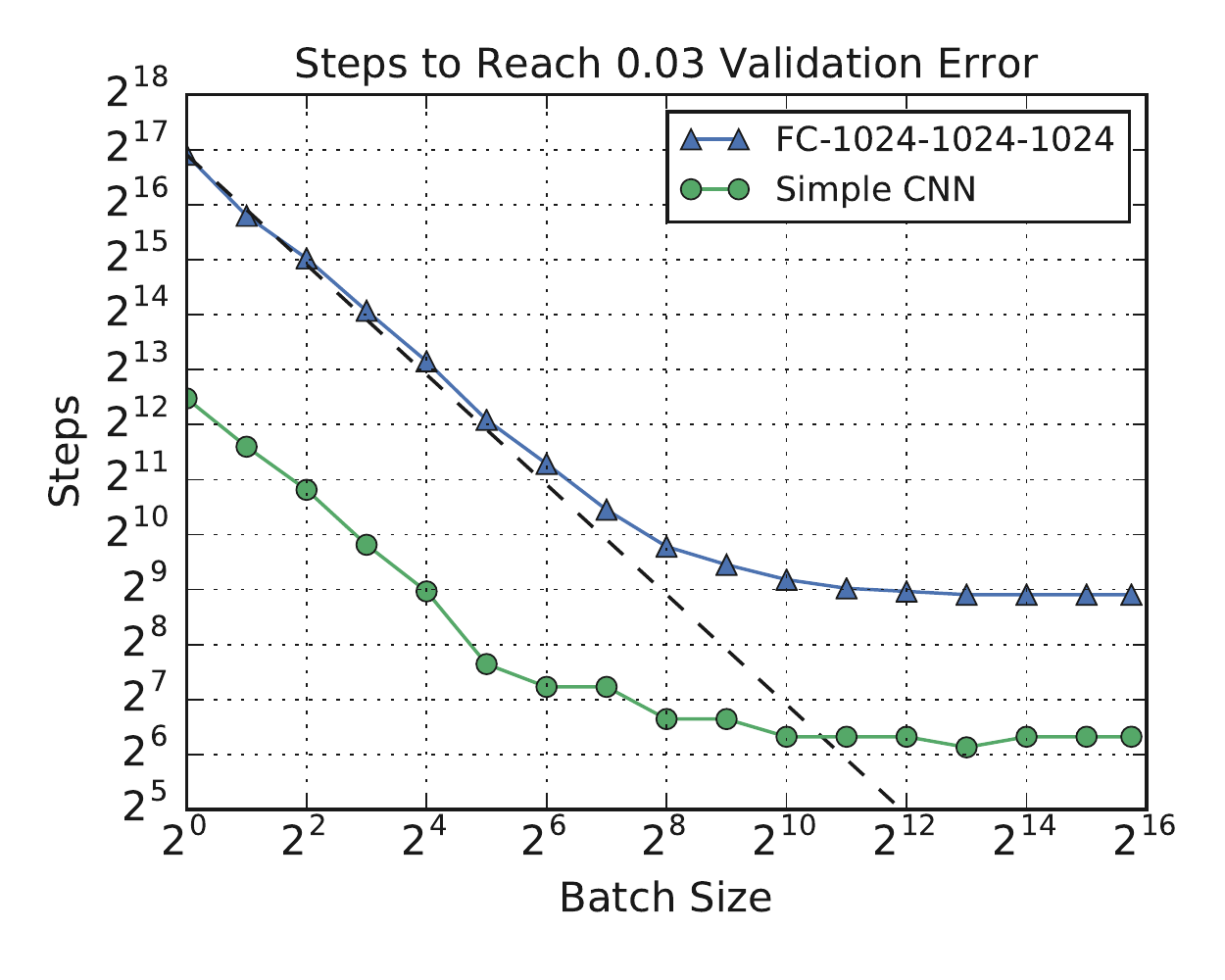}
        \vspace*{\capshift}
        \caption{Fully Connected vs Simple CNN on MNIST}
        \label{fig:stt-multiple-models-mnist-fc-cnn-raw}
    \end{subfigure}
    \begin{subfigure}[b]{\multimodelfigwidth}
        \includegraphics[width=\textwidth]{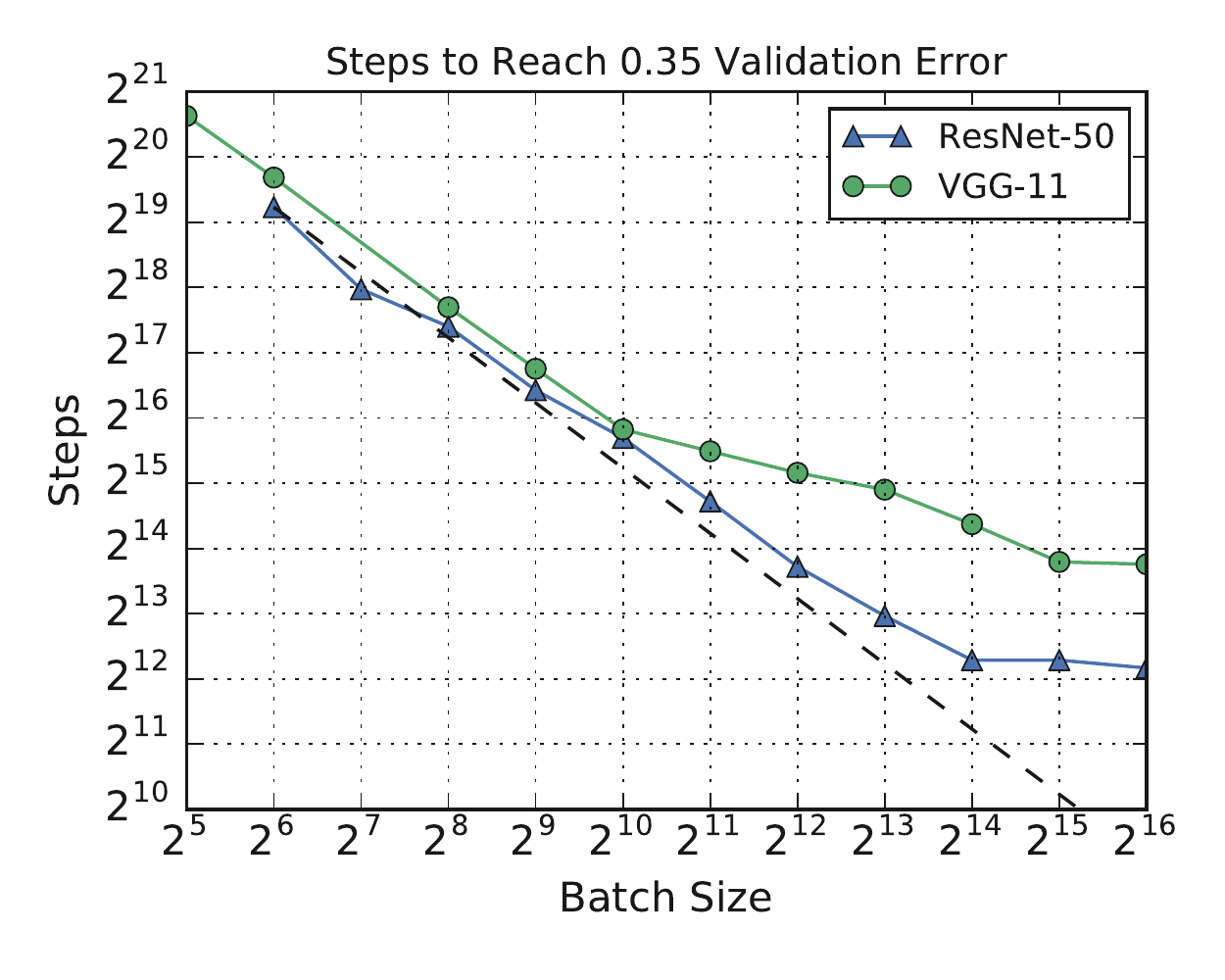}
        \vspace*{\capshift}
        \caption{ResNet-50 vs VGG-11 on ImageNet}
        \label{fig:stt-multiple-models-imagenet-raw}
    \end{subfigure}\\
    \vspace*{\lineshift}
    \begin{subfigure}[b]{\multimodelfigwidth}
        \includegraphics[width=\textwidth]{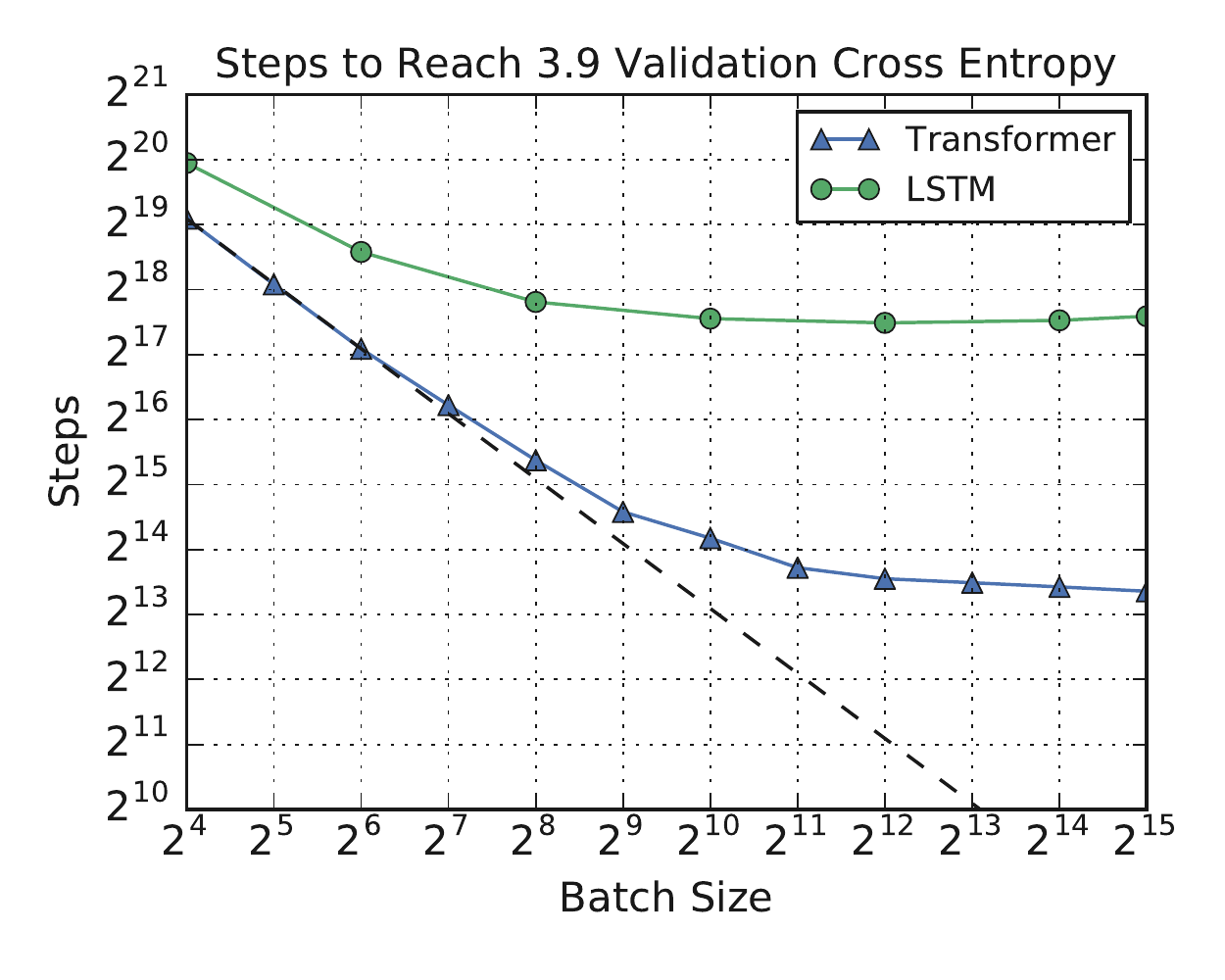}
        \vspace*{\capshift}
        \caption{Transformer vs LSTM on LM1B}
        \label{fig:stt-multiple-models-lm1b-raw}
    \end{subfigure}
    \begin{subfigure}[b]{\multimodelfigwidth}
        \includegraphics[width=\textwidth]{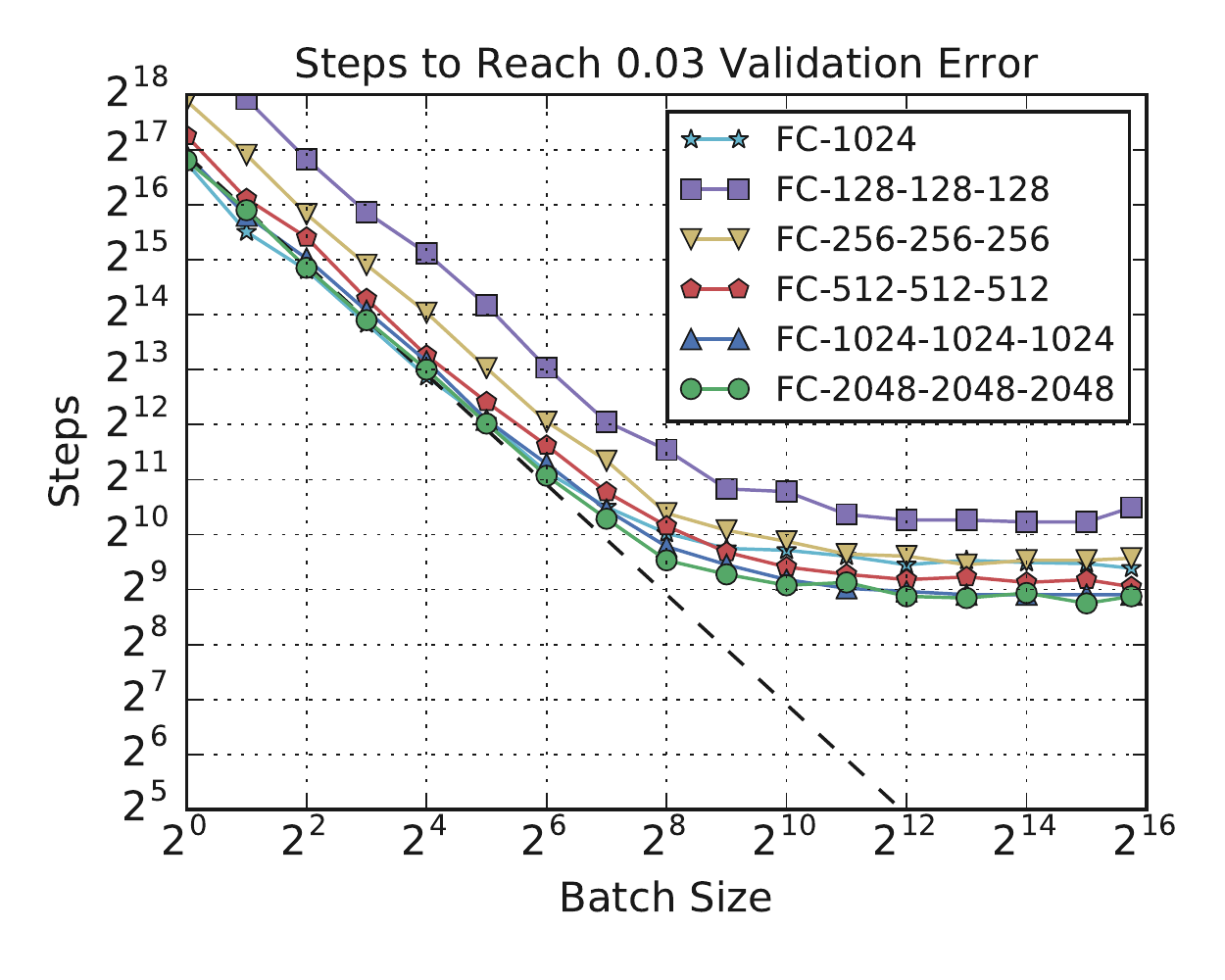}
        \vspace*{\capshift}
        \caption{Fully Connected sizes on MNIST}
        \label{fig:stt-multiple-models-fc-raw}
    \end{subfigure}\\
    \vspace*{\lineshift}
    \begin{subfigure}[b]{\multimodelfigwidth}
        \includegraphics[width=\textwidth]{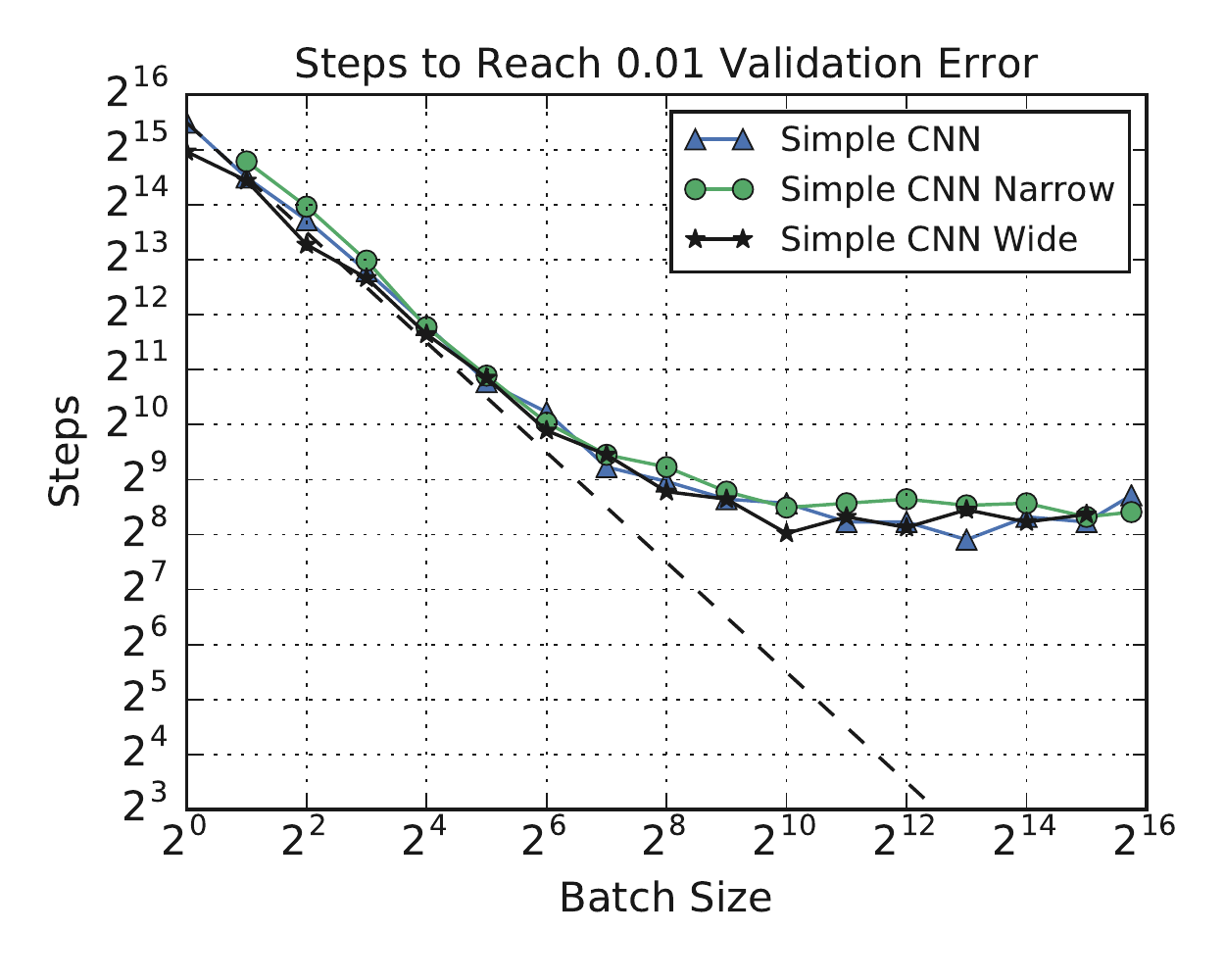}
        \vspace*{\capshift}
        \caption{Simple CNN sizes on MNIST}
        \label{fig:stt-multiple-models-cnn-raw}
    \end{subfigure}
    \begin{subfigure}[b]{\multimodelfigwidth}
        \includegraphics[width=\textwidth]{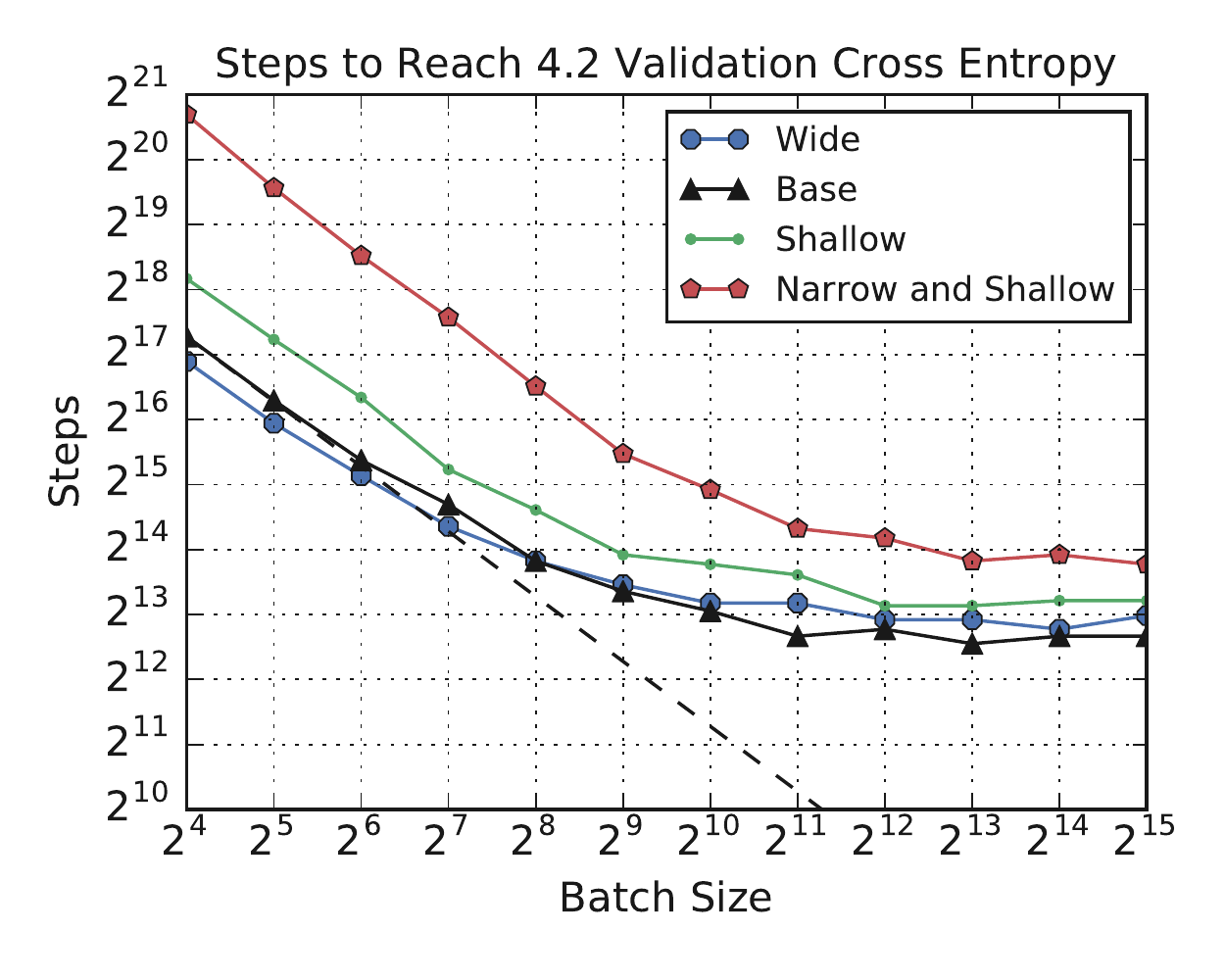}
        \vspace*{\capshift}
        \caption{Transformer sizes on LM1B}
        \label{fig:stt-multiple-model-sizes-lm1b-raw}
    \end{subfigure}
    \caption{\textbf{Figure~\ref{fig:stt-multiple-models} without the $y$-axis normalized.}}
    \label{fig:stt-multiple-models-raw}
\end{figure}

\begin{figure}
    \centering
    \begin{subfigure}[b]{\threecolfigwidth}
    \includegraphics[width=\textwidth]{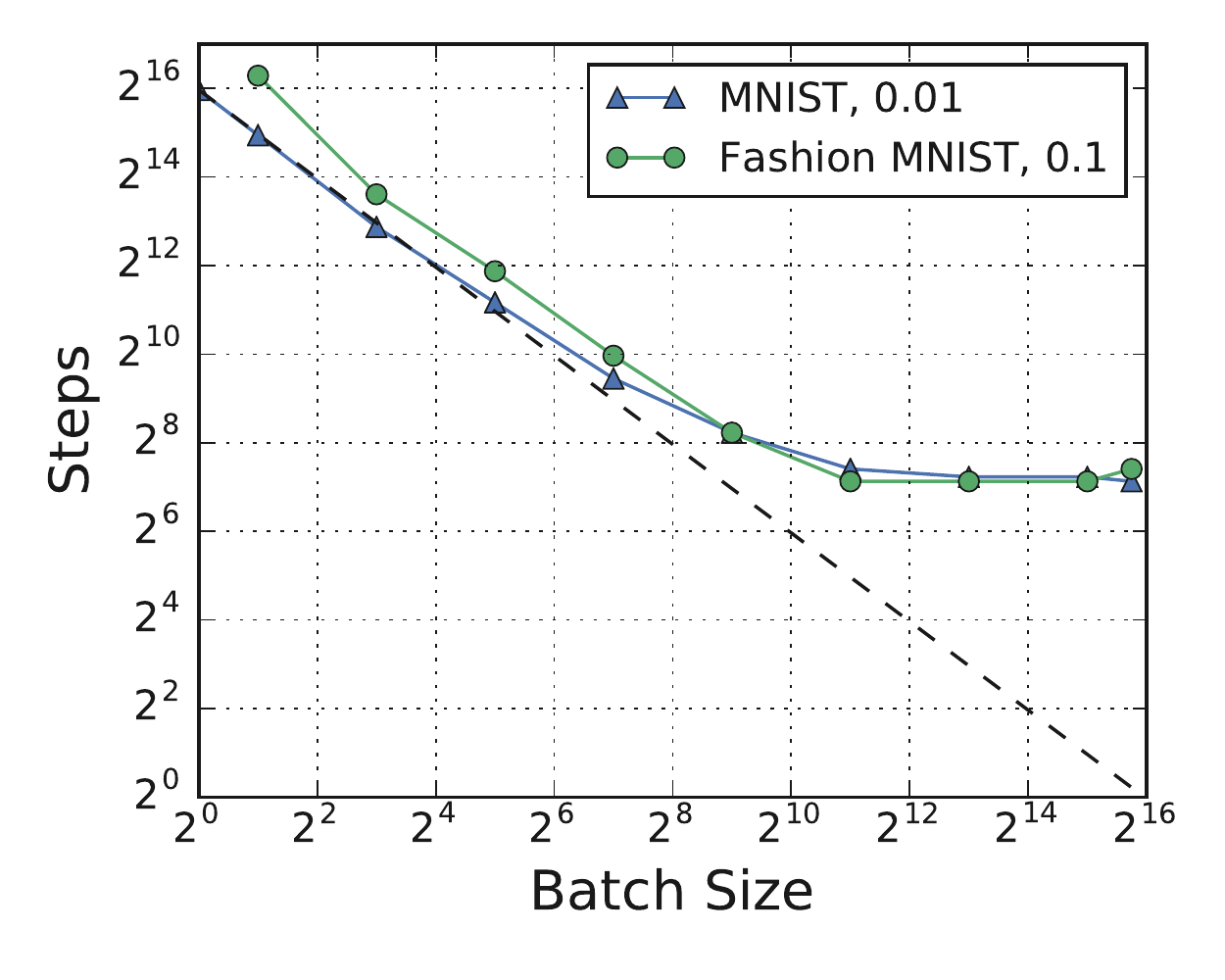}
    \vspace*{\capshift}
    \caption{Simple CNN on different data sets}
    \label{fig:mnist-and-fashion-raw}
    \end{subfigure}
    \begin{subfigure}[b]{\threecolfigwidth}
        \includegraphics[width=\textwidth]{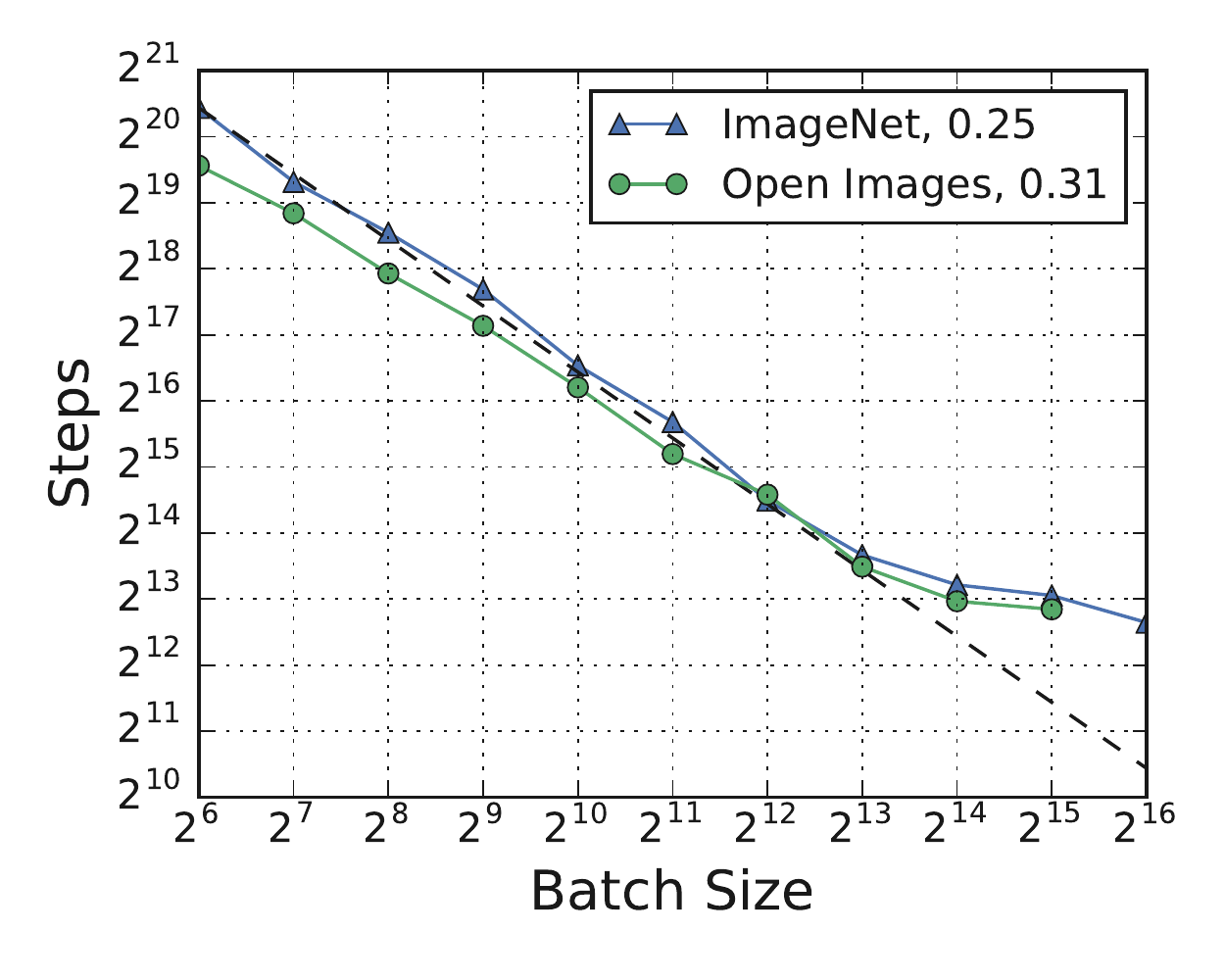}
        \vspace*{\capshift}
        \caption{ResNet-50 on different data sets}
        \label{fig:stt-resnet-imagenet-oi-raw}
    \end{subfigure}
    \begin{subfigure}[b]{\threecolfigwidth}
        \includegraphics[width=\textwidth]{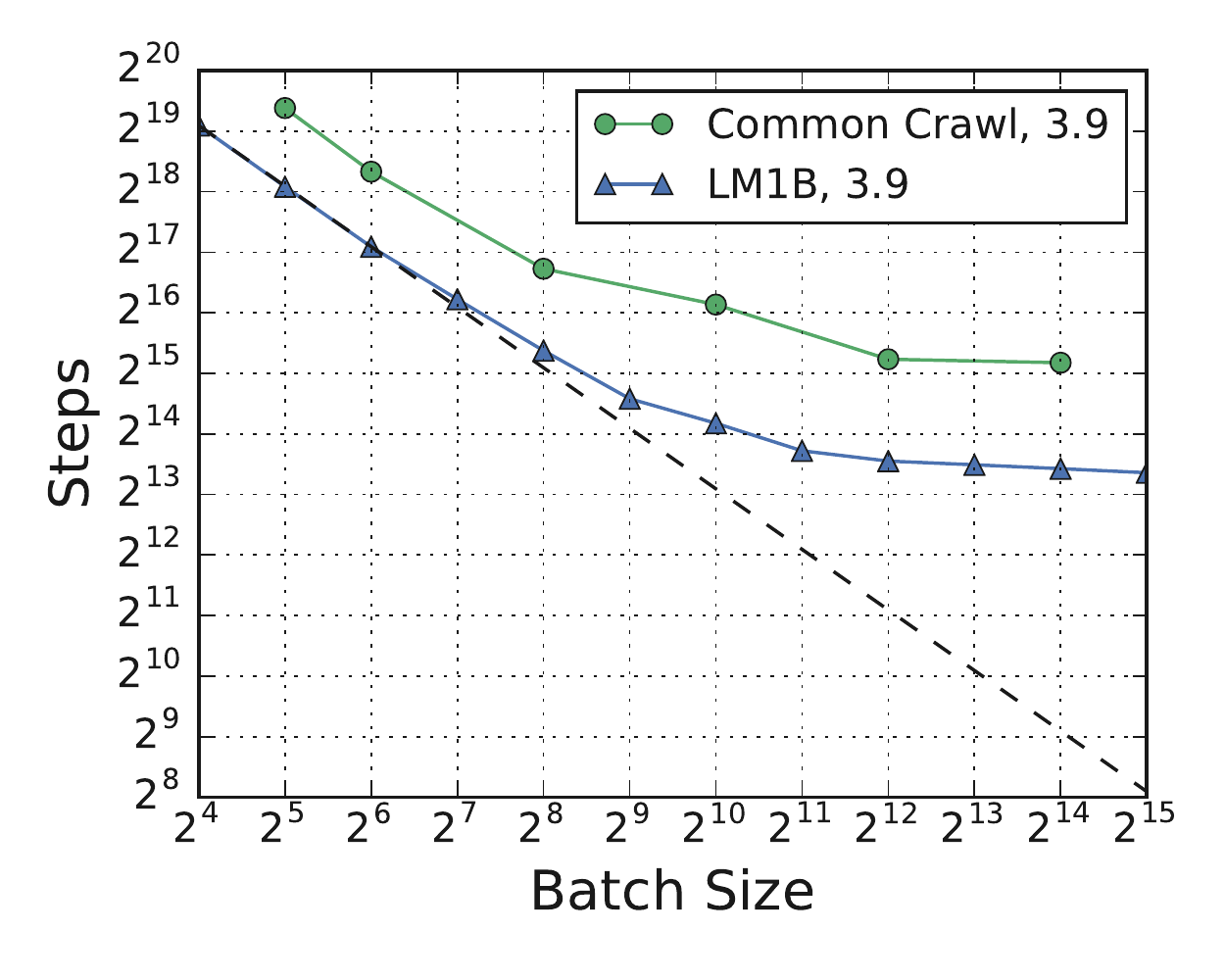}
        \vspace*{\capshift}
        \caption{Transformer on different data sets}
        \label{fig:stt-transformer-lm1b-cc-raw}
    \end{subfigure}
    \caption{\textbf{Figure~\ref{fig:stt-datasets} without the $y$-axis normalized.}}
    \label{fig:stt-datasets-raw}
\end{figure}

\begin{figure}
    \centering
    \begin{subfigure}[b]{0.48\textwidth}
        \includegraphics[width=\textwidth]{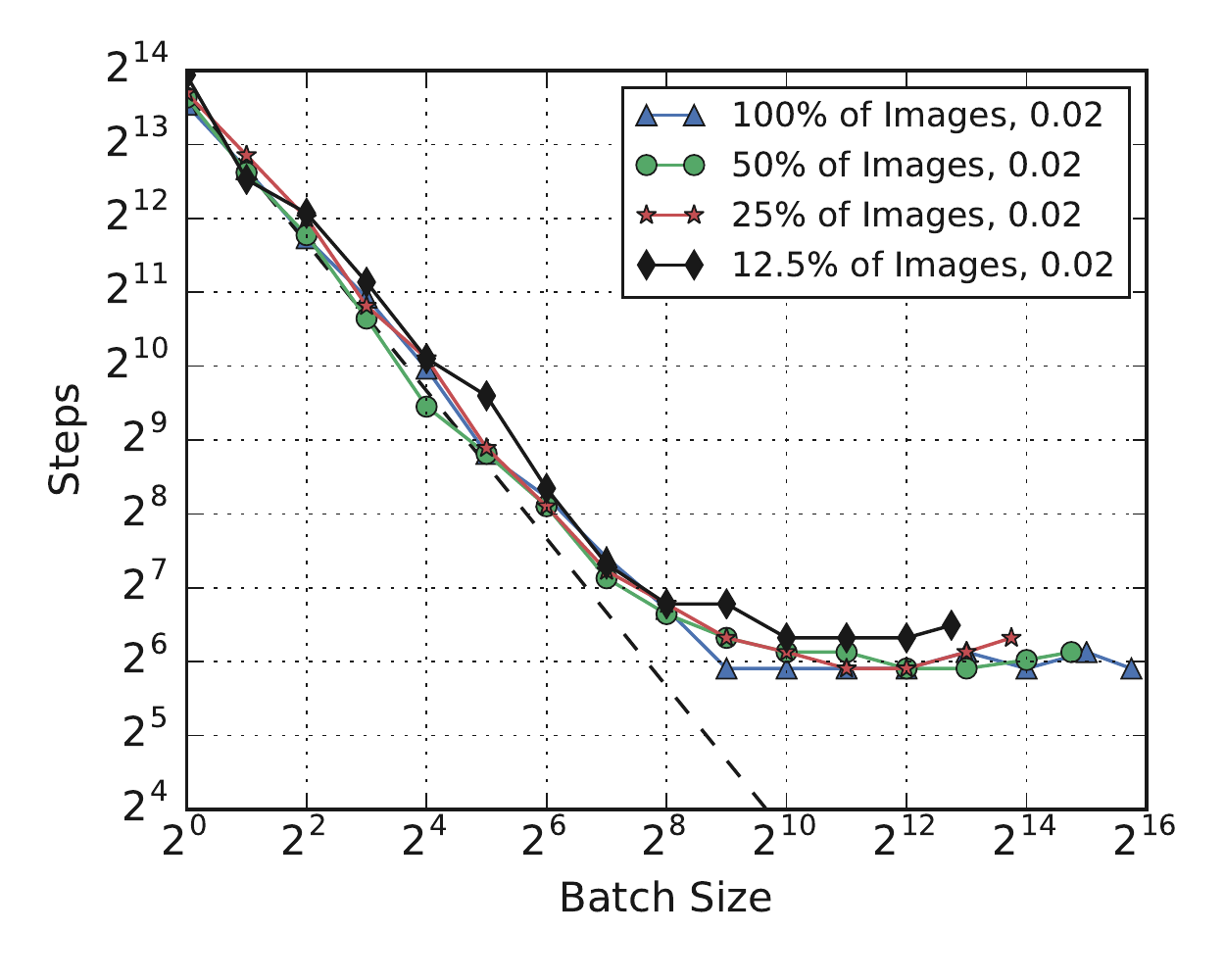}
        \vspace*{\capshift}
        \caption{Simple CNN on MNIST subsets}
        \label{fig:mnist-size-raw}
    \end{subfigure}
    \begin{subfigure}[b]{0.48\textwidth}
        \includegraphics[width=\textwidth]{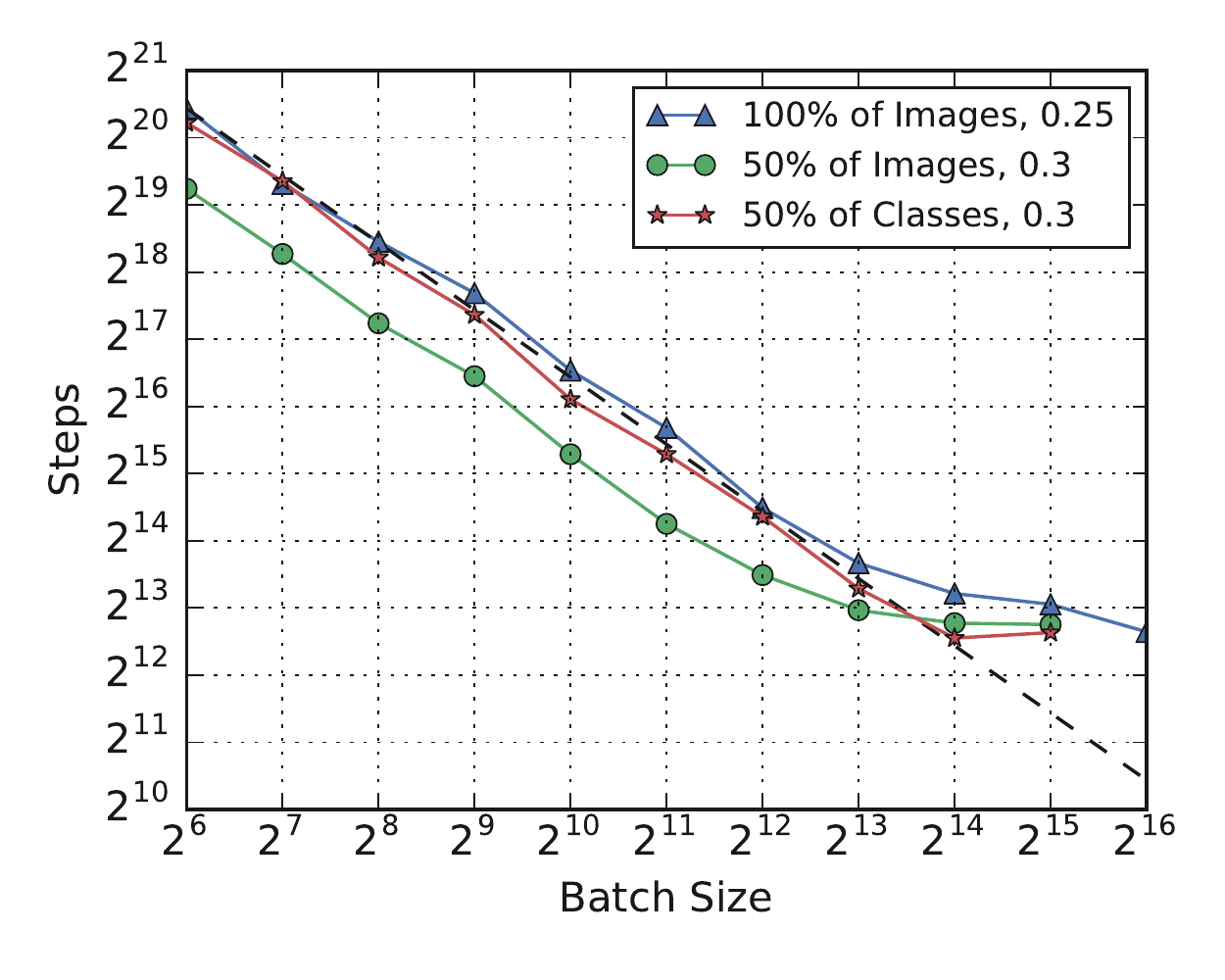}
        \vspace*{\capshift}
        \caption{ResNet-50 on ImageNet subsets}
        \label{fig:stt-resnet-half-raw}
    \end{subfigure}
    \caption{\textbf{Figure~\ref{fig:stt-dataset-sizes} without the $y$-axis normalized.}}
    \label{fig:stt-dataset-sizes-raw}
\end{figure}

\begin{figure}
    \centering
    \begin{subfigure}[b]{0.48\textwidth}
        \includegraphics[width=\textwidth]{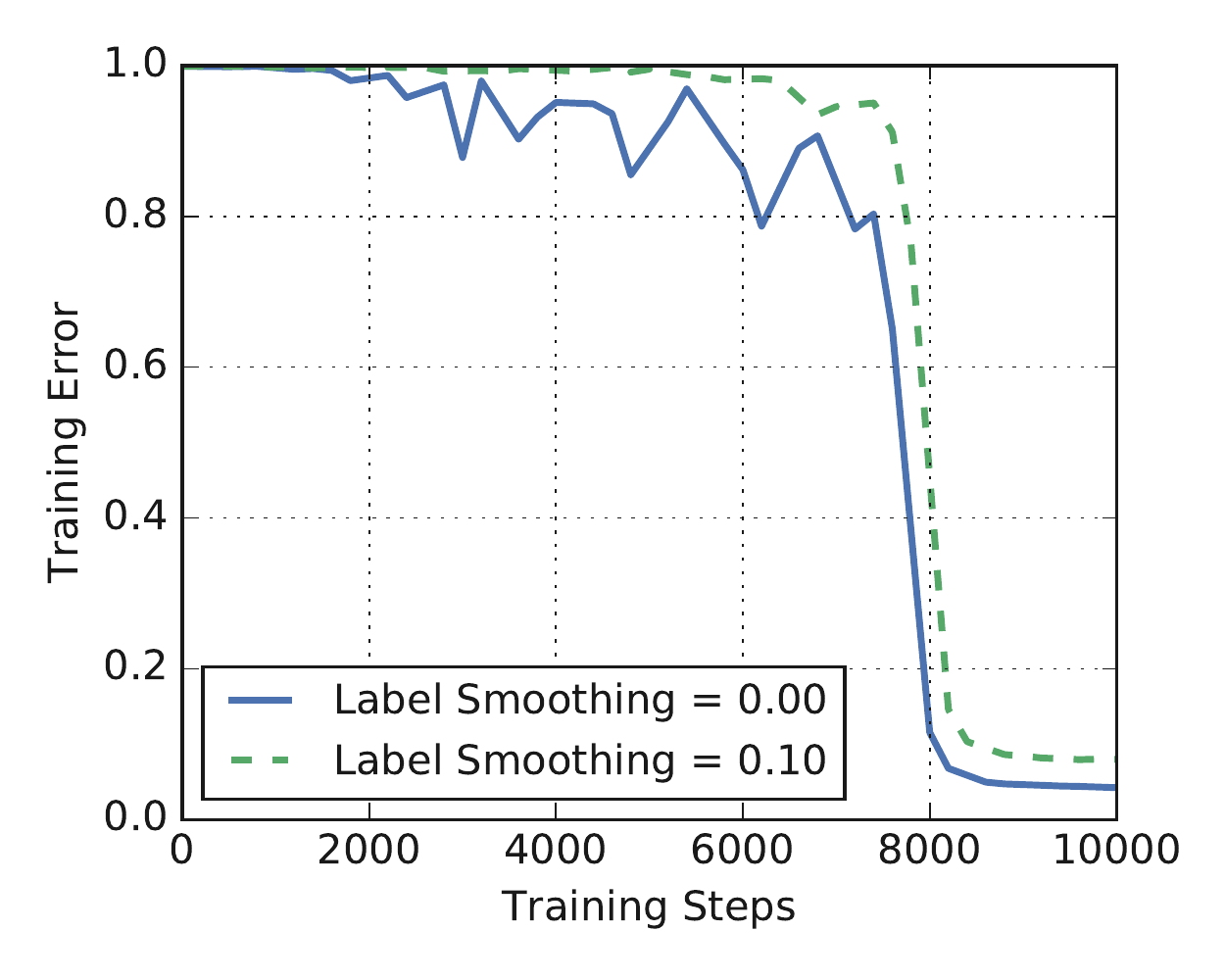}
    \end{subfigure}
    \begin{subfigure}[b]{0.48\textwidth}
        \includegraphics[width=\textwidth]{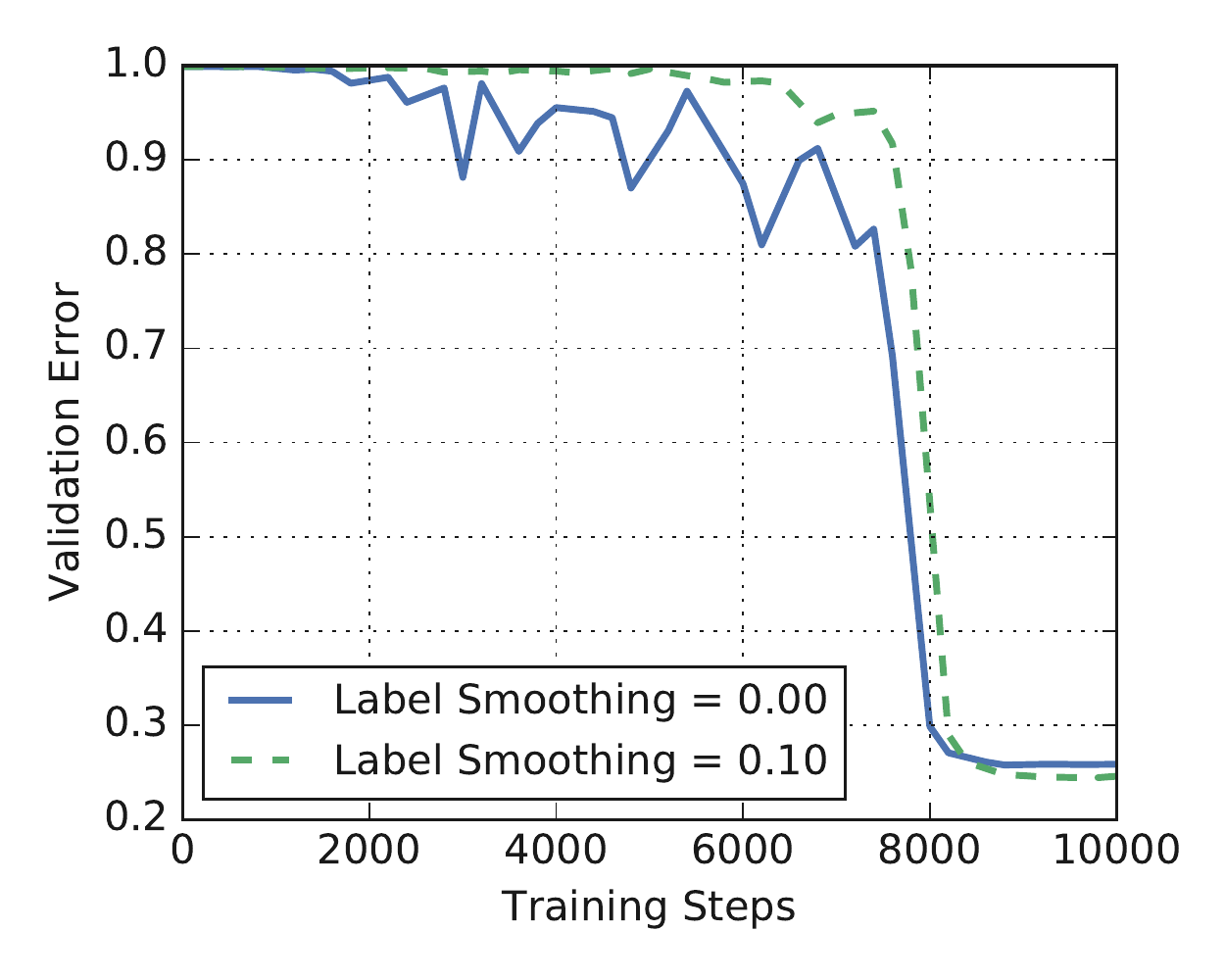}
    \end{subfigure}
    \caption{\textbf{Label smoothing reduces overfitting at large batch sizes.} Plots are training curves for the two best models with and without label smoothing for ResNet-50 on ImageNet with batch size $2^{16}$. The two models correspond to different metaparameter tuning trials, so the learning rate, Nesterov momentum, and learning rate schedule were independently chosen for each trial. The two trials shown are those that reached the highest validation error at any point during training, for label smoothing equal to 0 and $0.1$ respectively.}
    \label{fig:label-smoothing-training-curves}
\end{figure}

\begin{figure}
    \centering
    \begin{subfigure}[b]{0.49\textwidth}
        \includegraphics[width=\textwidth]{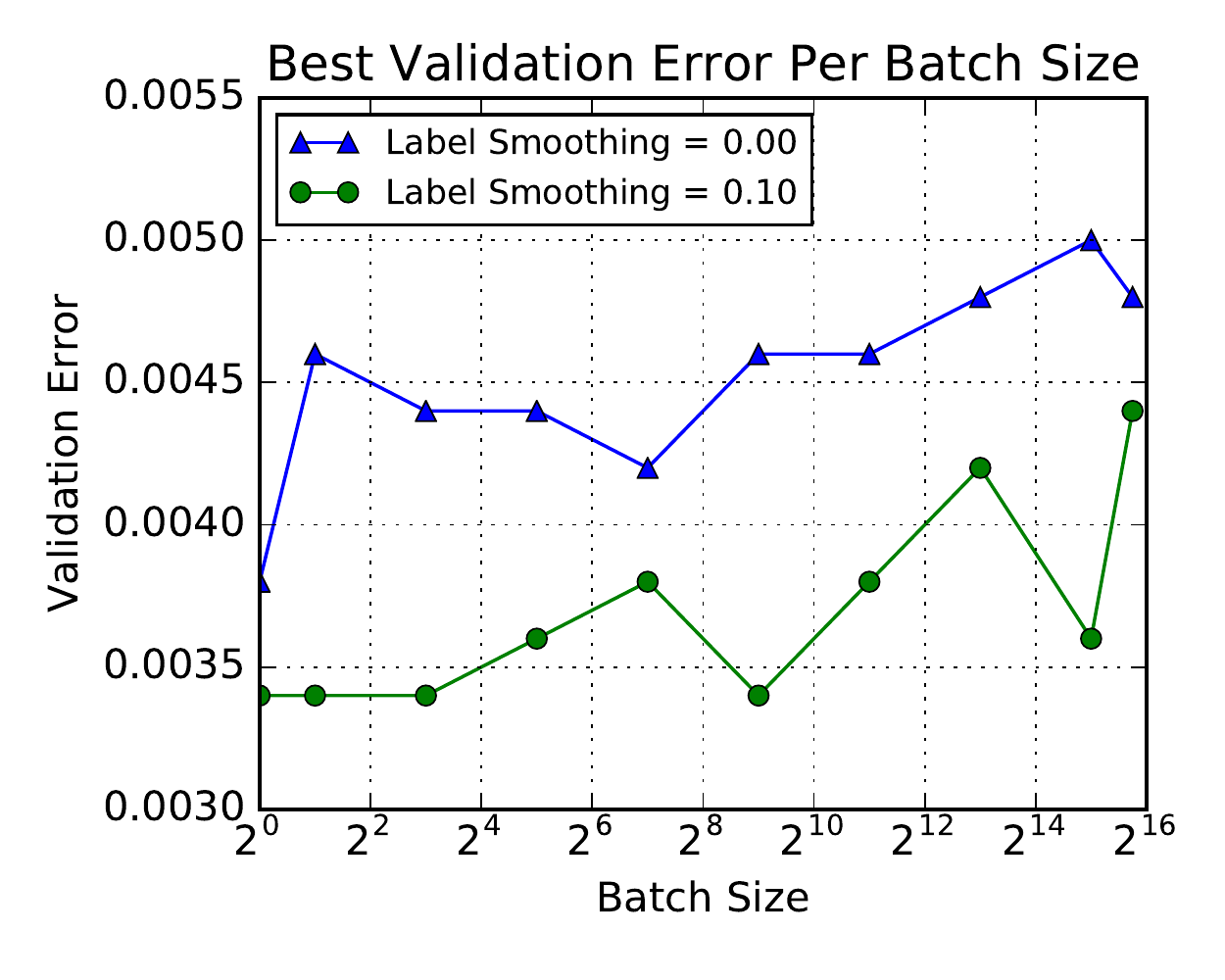}
        \vspace*{\capshift}
        \caption{Simple CNN on MNIST}
        \label{fig:label-smoothing-mnist}
    \end{subfigure}
    \begin{subfigure}[b]{0.49\textwidth}
        \includegraphics[width=\textwidth]{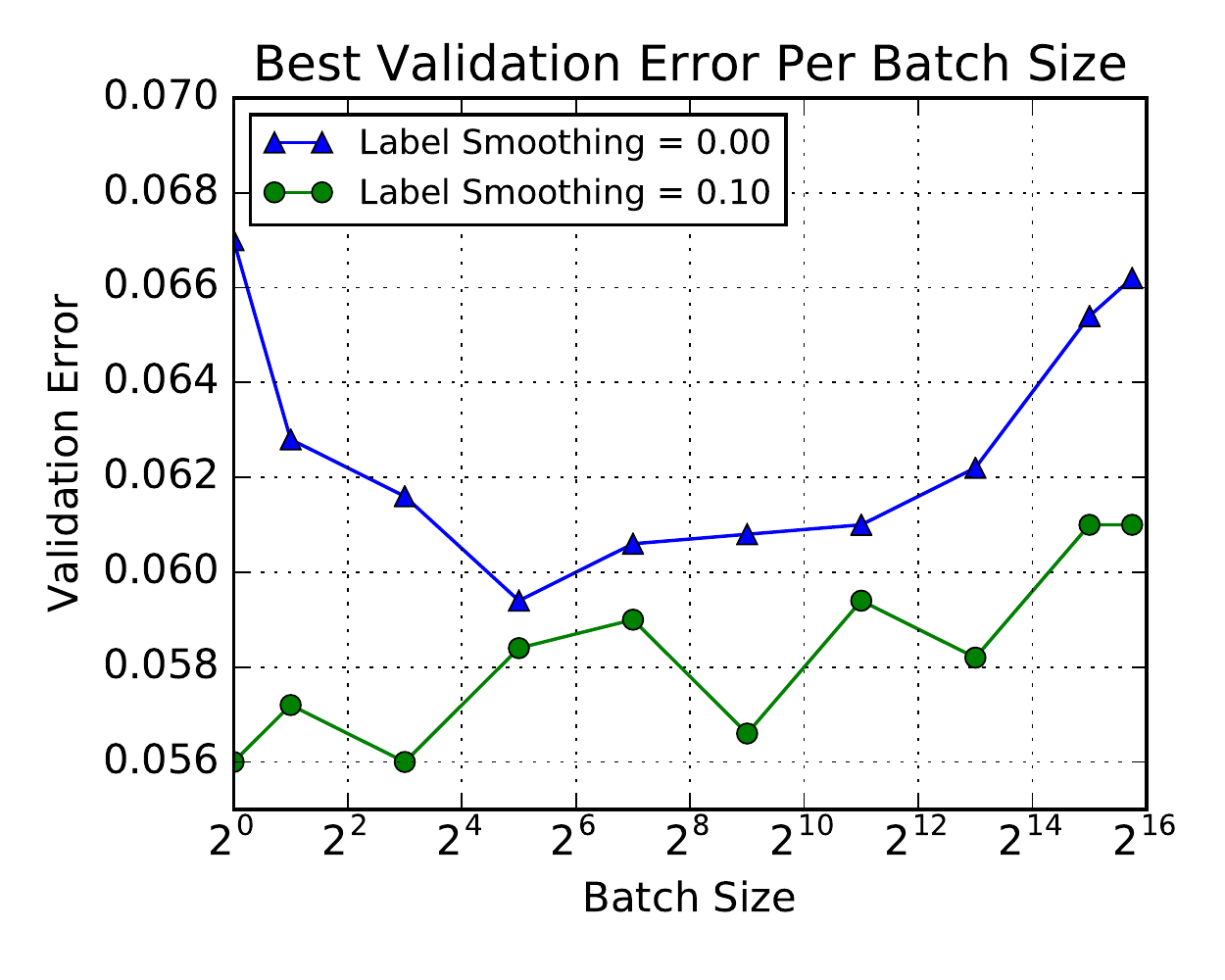}
        \vspace*{\capshift}
        \caption{Simple CNN on Fashion MNIST}
        \label{fig:label-smoothing-fashion-mnist}
    \end{subfigure}
    \caption{\textbf{Label smoothing helps all batch sizes for Simple CNN on MNIST and Fashion MNIST.} There is no consistent trend of label smoothing helping smaller or larger batch sizes more. Each point corresponds to a different metaparameter tuning trial, so the learning rate, Nesterov momentum, and learning rate schedule are independently chosen for each point. The training budget is fixed for each batch size, but varies between batch sizes.}
    \label{fig:label-smoothing-mnists}
\end{figure}

\begin{figure}
    \centering
    \begin{subfigure}[b]{\textwidth}
        \includegraphics[width=\textwidth]{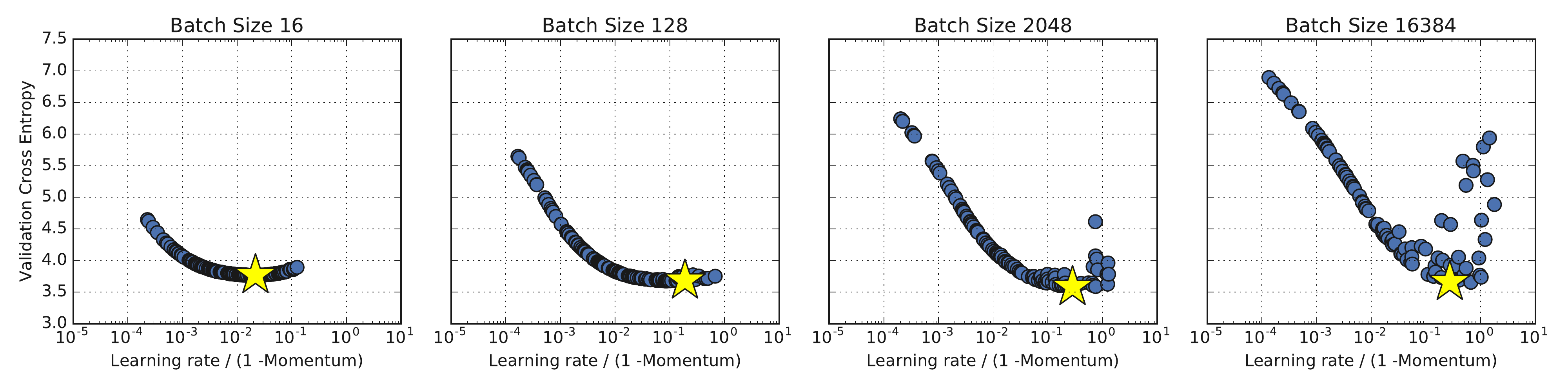}
        \vspace*{\capshift}
        \caption{\small Transformer on LM1B}
        \label{fig:hparams-error-vs-lr-transformer-lm1b}
    \end{subfigure}\\
    \vspace*{\lineshift}
    \begin{subfigure}[b]{\textwidth}
        \includegraphics[width=\textwidth]{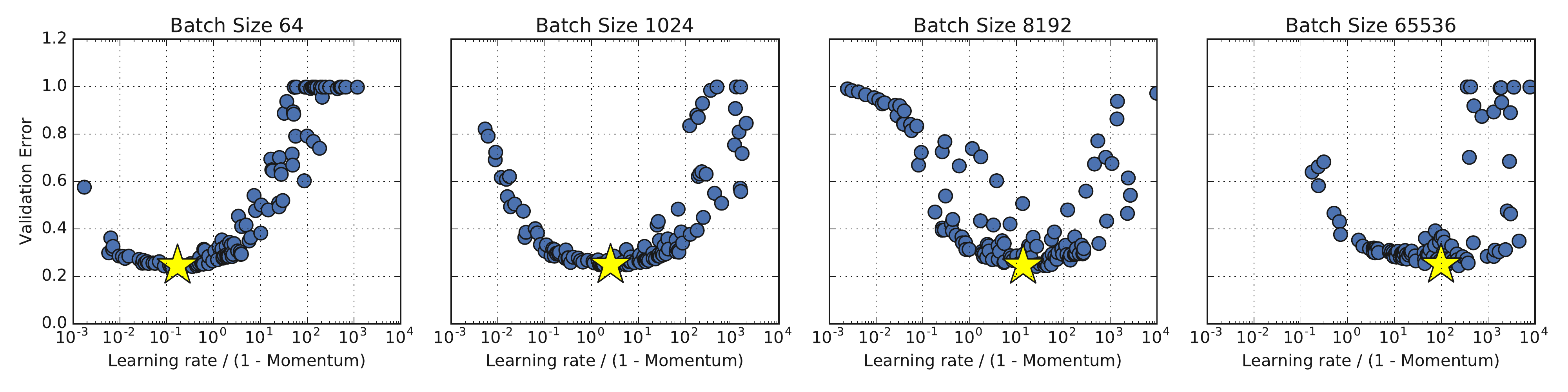}
        \vspace*{\capshift}
        \caption{\small ResNet-50 on ImageNet}
        \label{fig:hparams-error-vs-lr-resnet50-imagenet}
    \end{subfigure}
    \caption{\textbf{Validation error vs effective learning rate.} Training budgets are consistent for each batch size, but not between batch sizes. These plots are projections of the entire metaparameter search space, which is 2-dimensional for Transformer on LM1B (see Figure~\ref{fig:hparams-validation-transformer}) and 5-dimensional for ResNet-50 on ImageNet (see Figure~\ref{fig:hparams-validation-resnet}).}
    \label{fig:hparams-error-vs-lr}
\end{figure}

\begin{figure}
    \centering
    \begin{subfigure}[b]{\threecolfigwidth}
        \includegraphics[width=\textwidth]{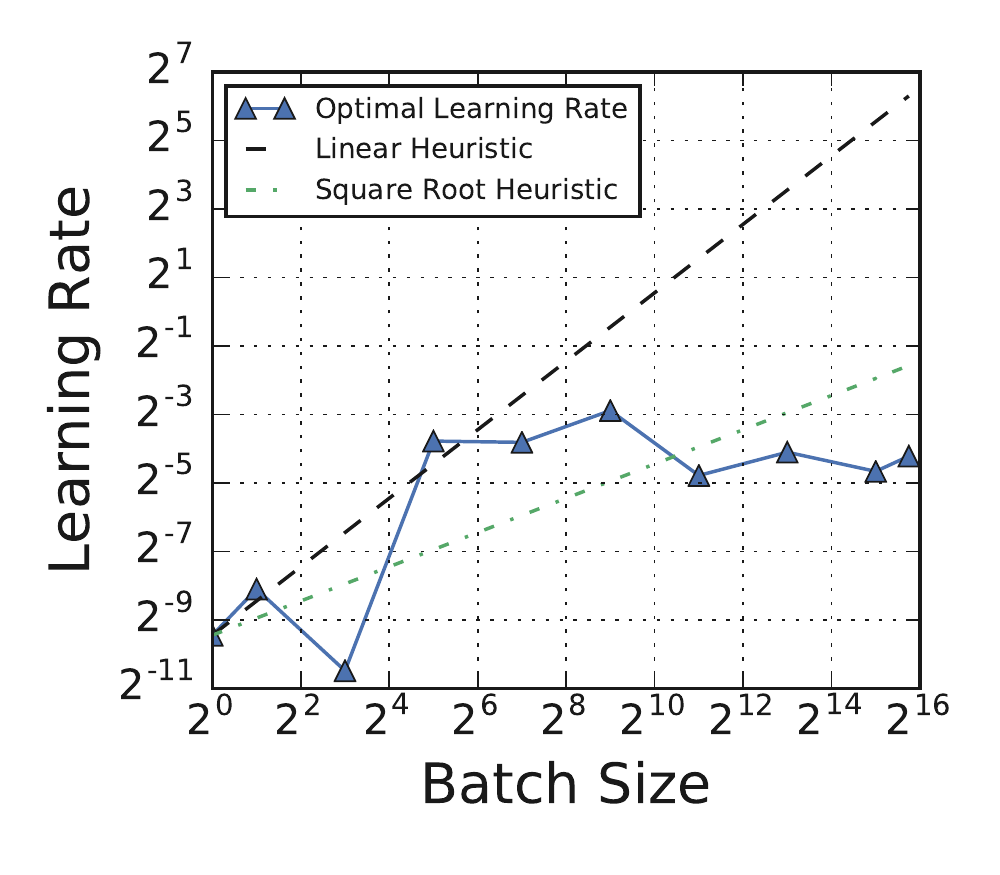}
        \vspace*{\capshift}
        \caption{Simple CNN on MNIST}
        \label{fig:hparams-raw-lr-cnn-mnist}
    \end{subfigure}
    \begin{subfigure}[b]{\threecolfigwidth}
        \includegraphics[width=\textwidth]{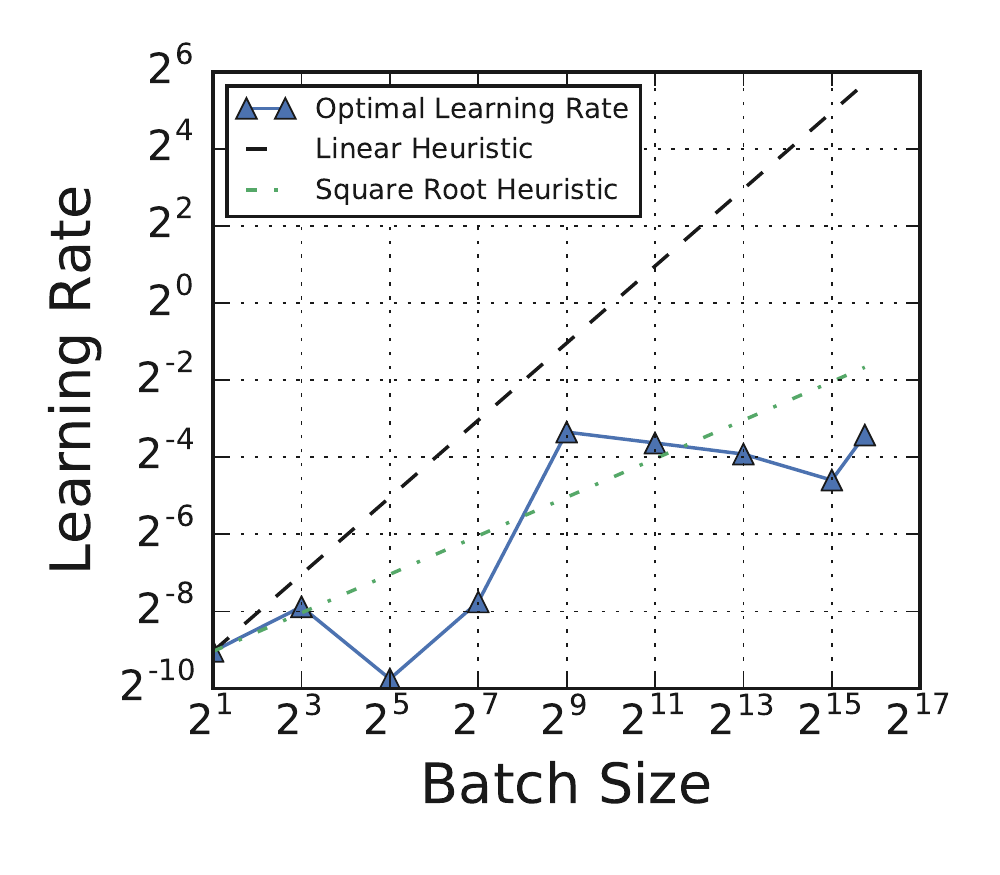}
        \vspace*{\capshift}
        \caption{Simple CNN on Fashion MNIST}
        \label{fig:hparams-raw-lr-cnn-fmnist}
    \end{subfigure}
    \begin{subfigure}[b]{\threecolfigwidth}
        \includegraphics[width=\textwidth]{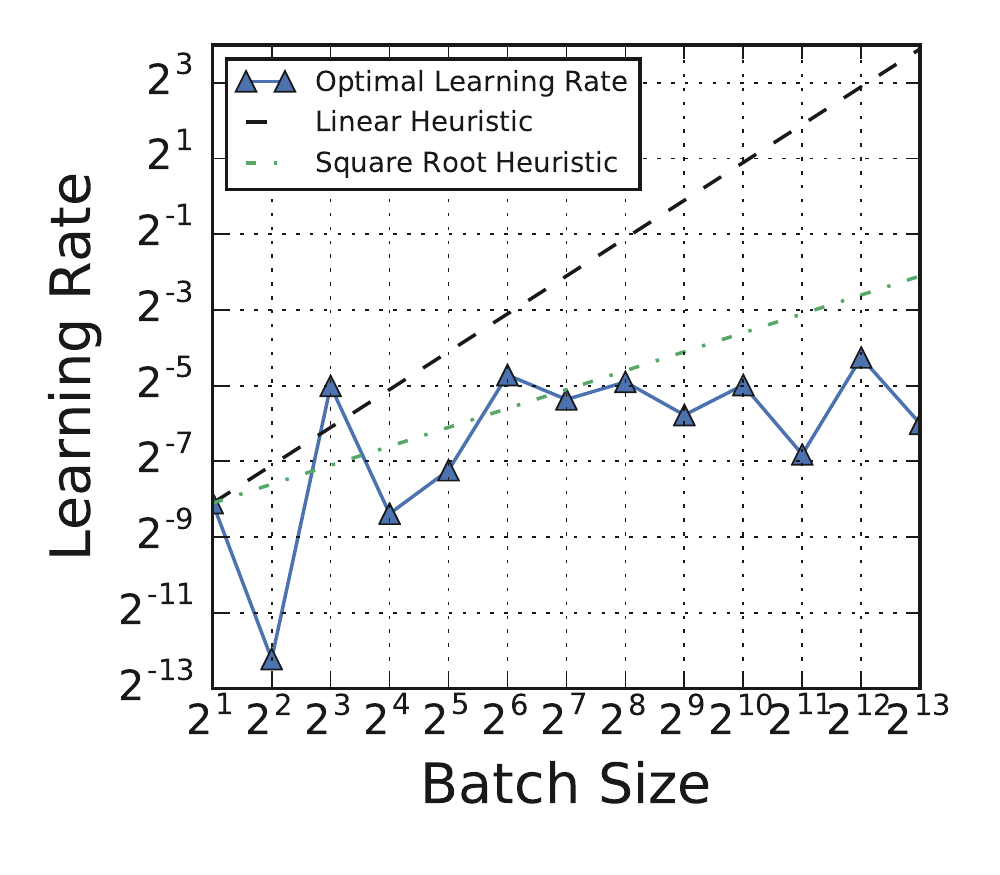}
        \vspace*{\capshift}
        \caption{ ResNet-8 on CIFAR-10}
        \label{fig:hparams-raw-lr-resnet-cifar}
    \end{subfigure} \\
    \vspace*{\lineshift}
    \begin{subfigure}[b]{\threecolfigwidth}
        \includegraphics[width=\textwidth]{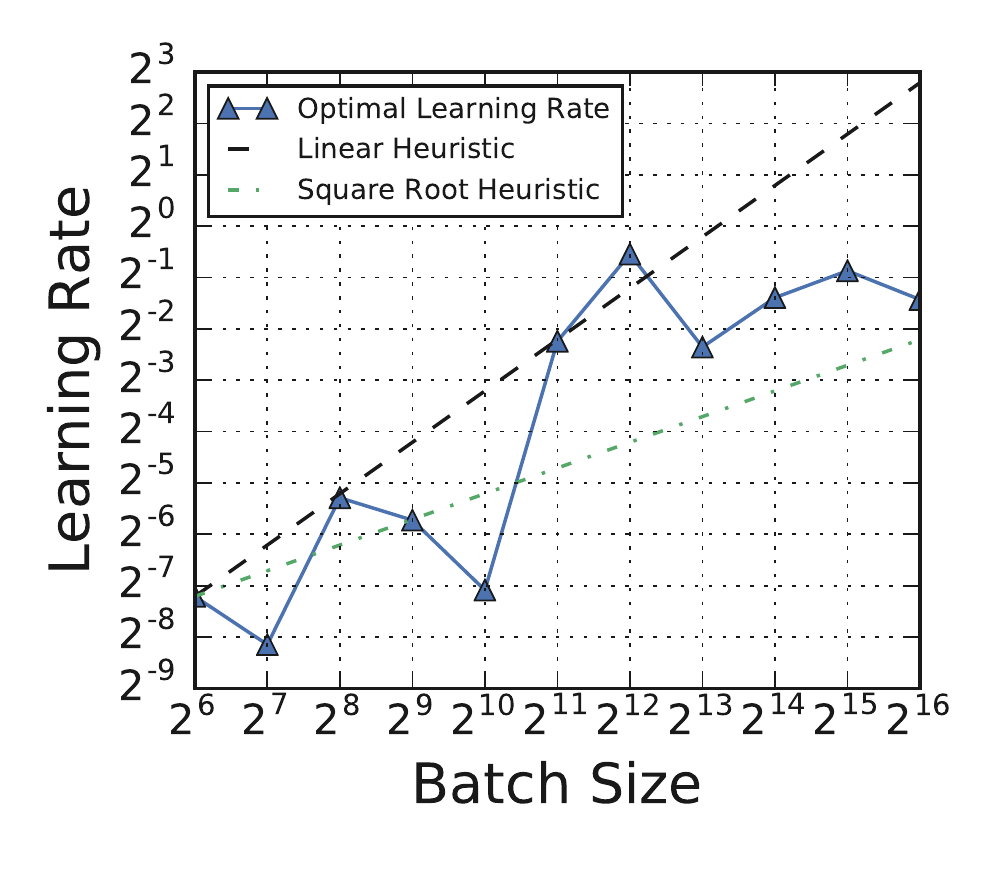}
        \vspace*{\capshift}
        \caption{ResNet-50 on ImageNet}
        \label{fig:hparams-raw-lr-resnet-imagenet}
    \end{subfigure}
    \begin{subfigure}[b]{\threecolfigwidth}
        \includegraphics[width=\textwidth]{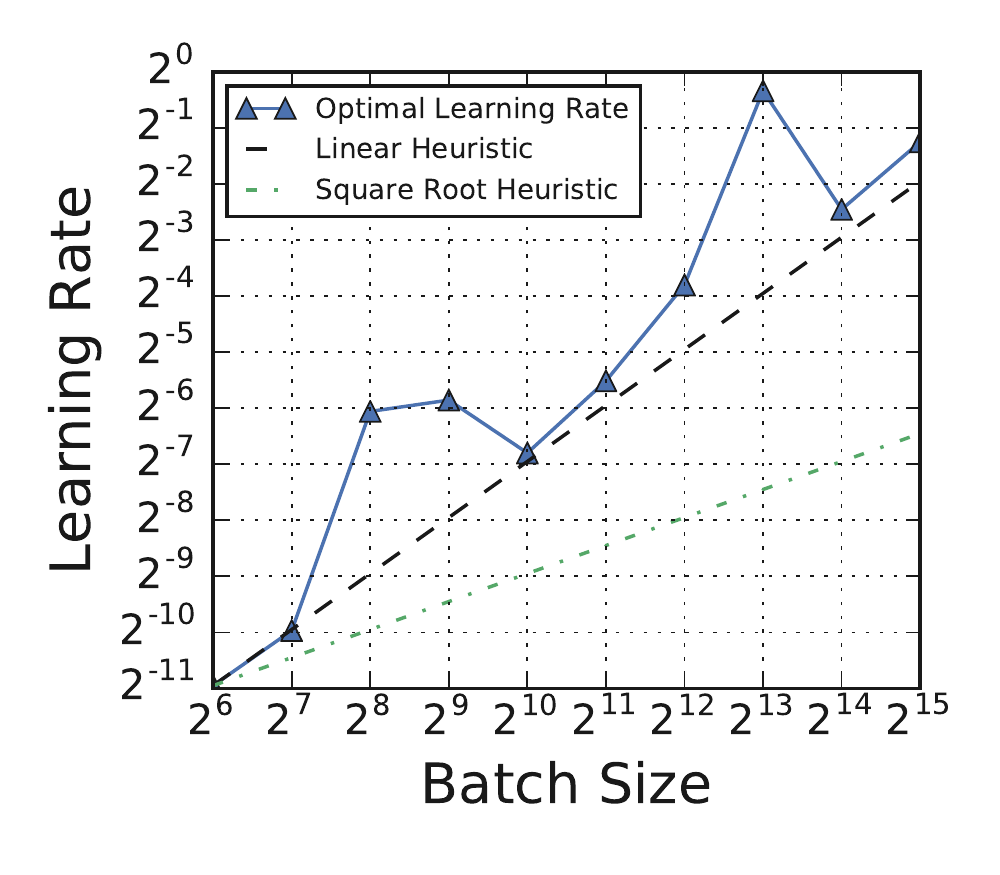}
        \vspace*{\capshift}
        \caption{ ResNet-50 on Open Images}
        \label{fig:hparams-raw-lr-resnet-oi}
    \end{subfigure}
    \begin{subfigure}[b]{\threecolfigwidth}
        \includegraphics[width=\textwidth]{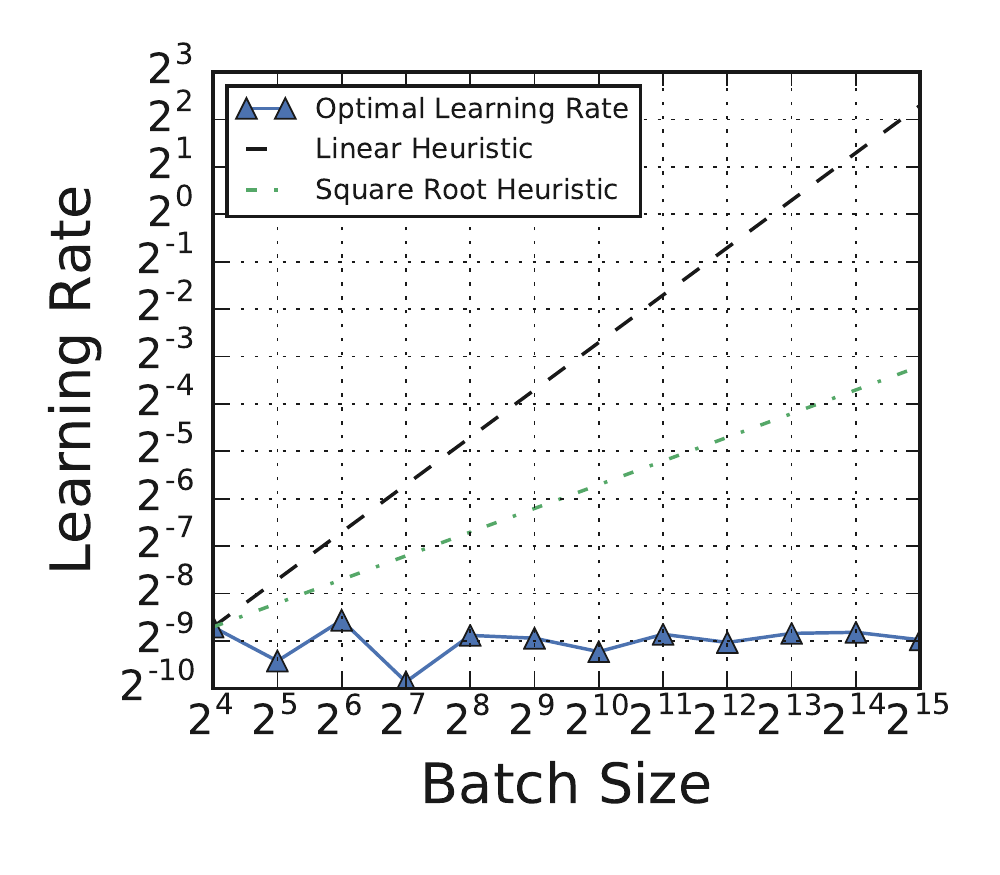}
        \vspace*{\capshift}
        \caption{ Transformer on LM1B}
        \label{fig:hparams-raw-lr-transformer-lm1b}
    \end{subfigure}\\
    \vspace*{\lineshift}
    \begin{subfigure}[b]{\threecolfigwidth}
        \includegraphics[width=\textwidth]{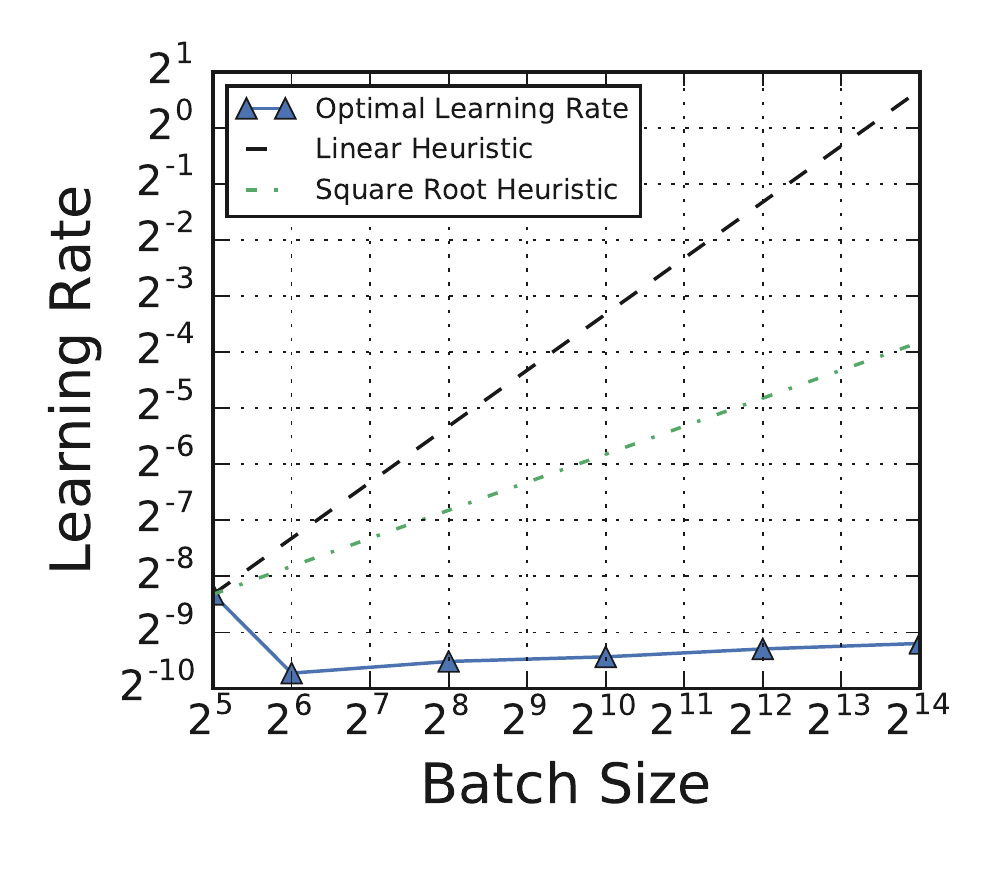}
        \vspace*{\capshift}
        \caption{Transformer on Common Crawl}
        \label{fig:hparams-raw-lr-transformer-cc}
    \end{subfigure}
    \begin{subfigure}[b]{\threecolfigwidth}
        \includegraphics[width=\textwidth]{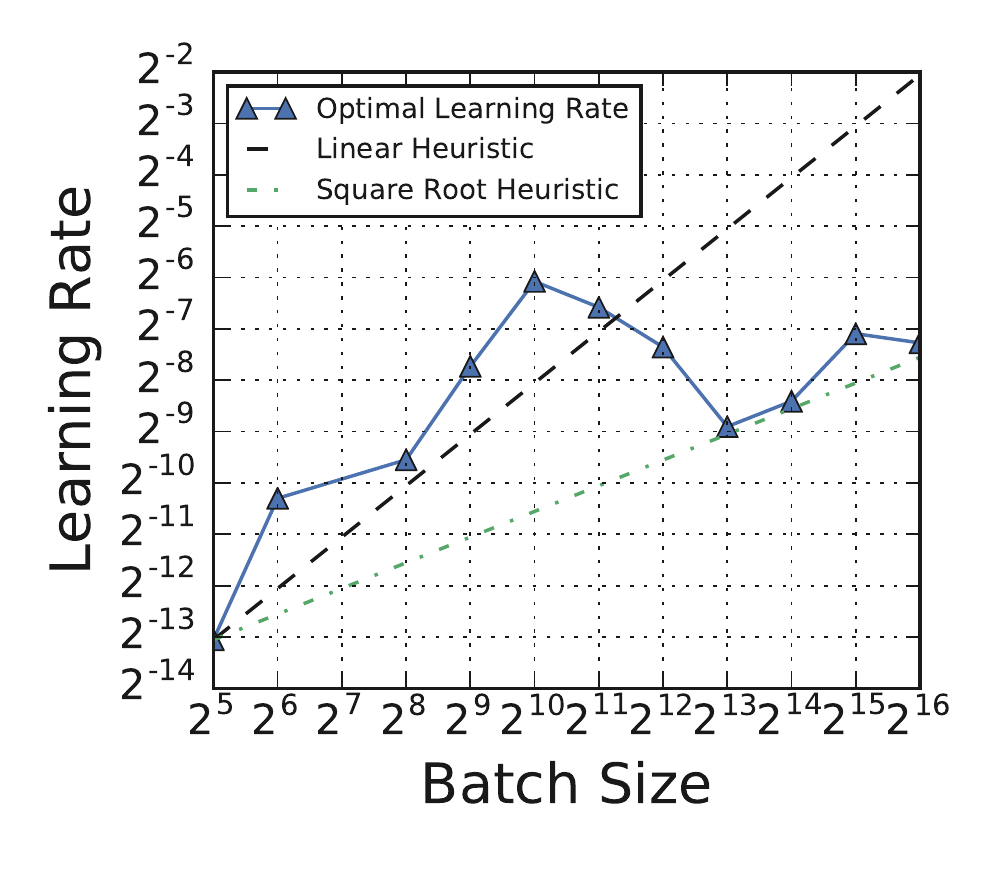}
        \vspace*{\capshift}
        \caption{ VGG-11 on ImageNet}
        \label{fig:hparams-raw-lr-vgg11-imagenet}
    \end{subfigure}
        \begin{subfigure}[b]{\threecolfigwidth}
        \includegraphics[width=\textwidth]{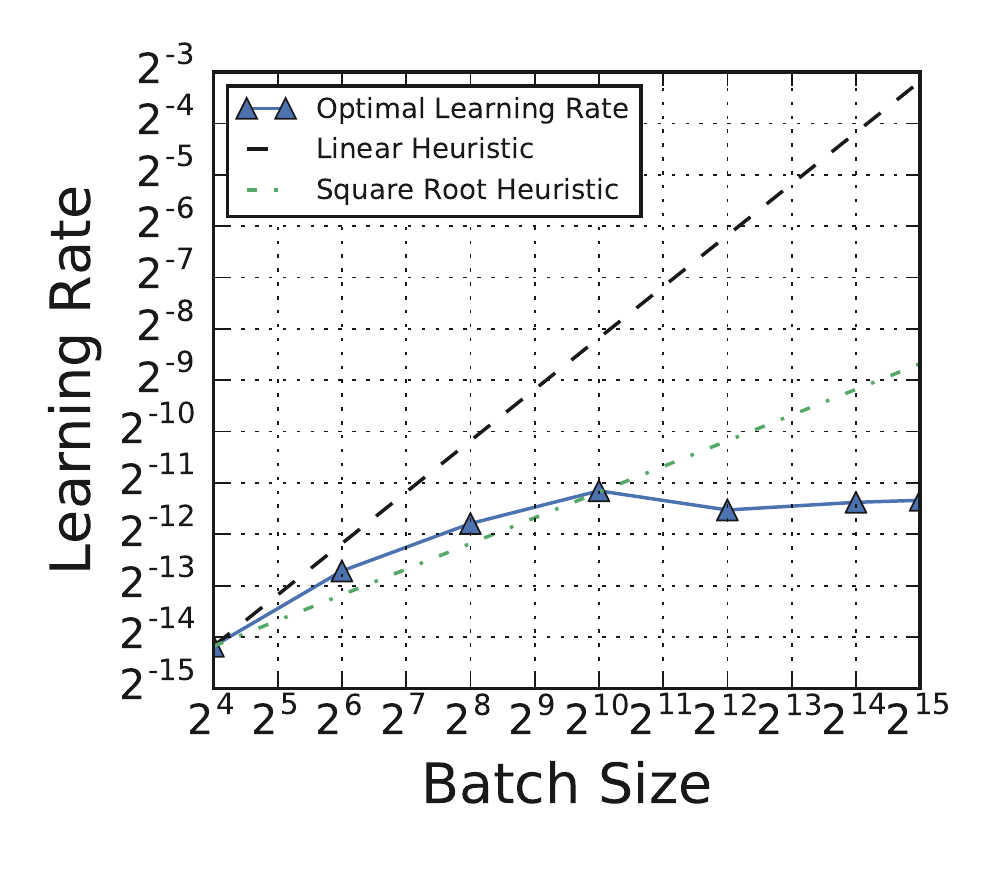}
        \vspace*{\capshift}
        \caption{LSTM on LM1B}
        \label{fig:hparams-raw-lr-lstm-lm1b}
    \end{subfigure}
    \caption{\textbf{Optimal learning rates do not always follow linear or square root scaling heuristics.} Learning rates correspond to the trial that reached the goal validation error in the fewest training steps (see Figure~\ref{fig:stt-problems}). For models using learning rate decay schedules (ResNet-8, ResNet-50, VGG-11), plots are based on the initial learning rate. See Figure~\ref{fig:momentum} for the corresponding plot of optimal momentum, and Figure~\ref{fig:effective-lr} for the corresponding plot of effective learning rate.}
    \label{fig:raw-lr}
\end{figure}

\begin{figure}
    \centering
    \begin{subfigure}[b]{\threecolfigwidth}
        \includegraphics[width=\textwidth]{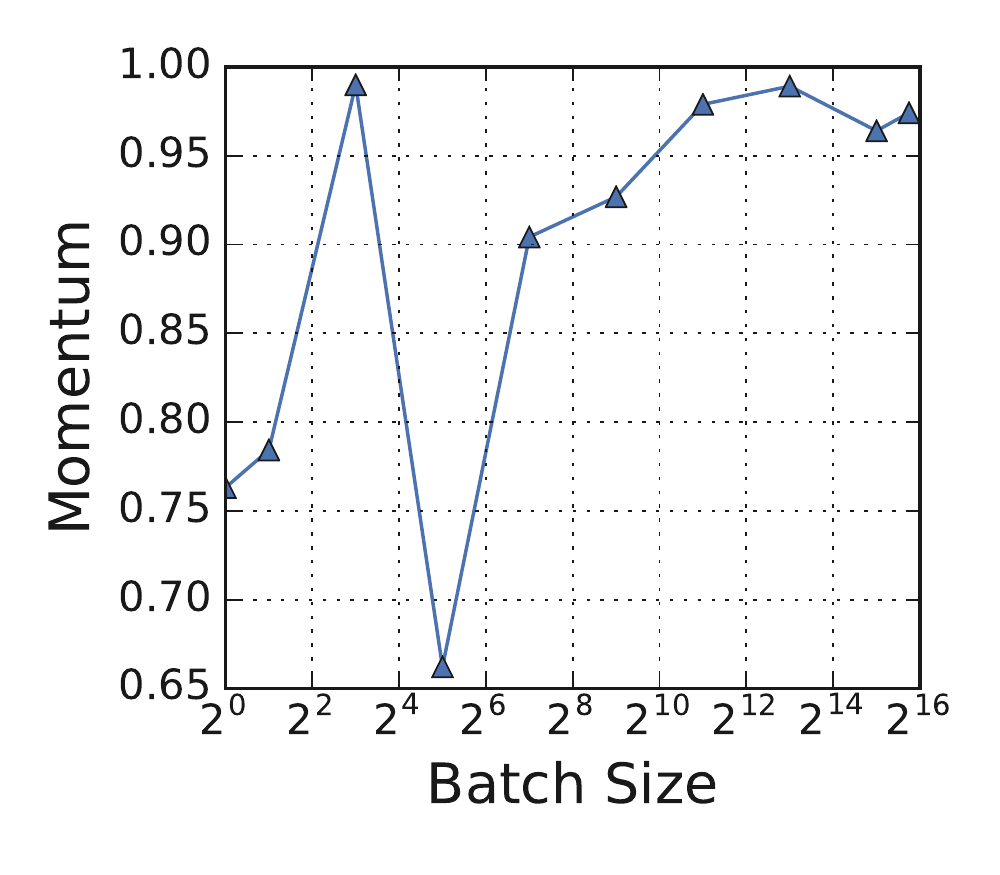}
        \vspace*{\capshift}
        \caption{Simple CNN on MNIST}
        \label{fig:hparams-momentum-cnn-mnist}
    \end{subfigure}
    \begin{subfigure}[b]{\threecolfigwidth}
        \includegraphics[width=\textwidth]{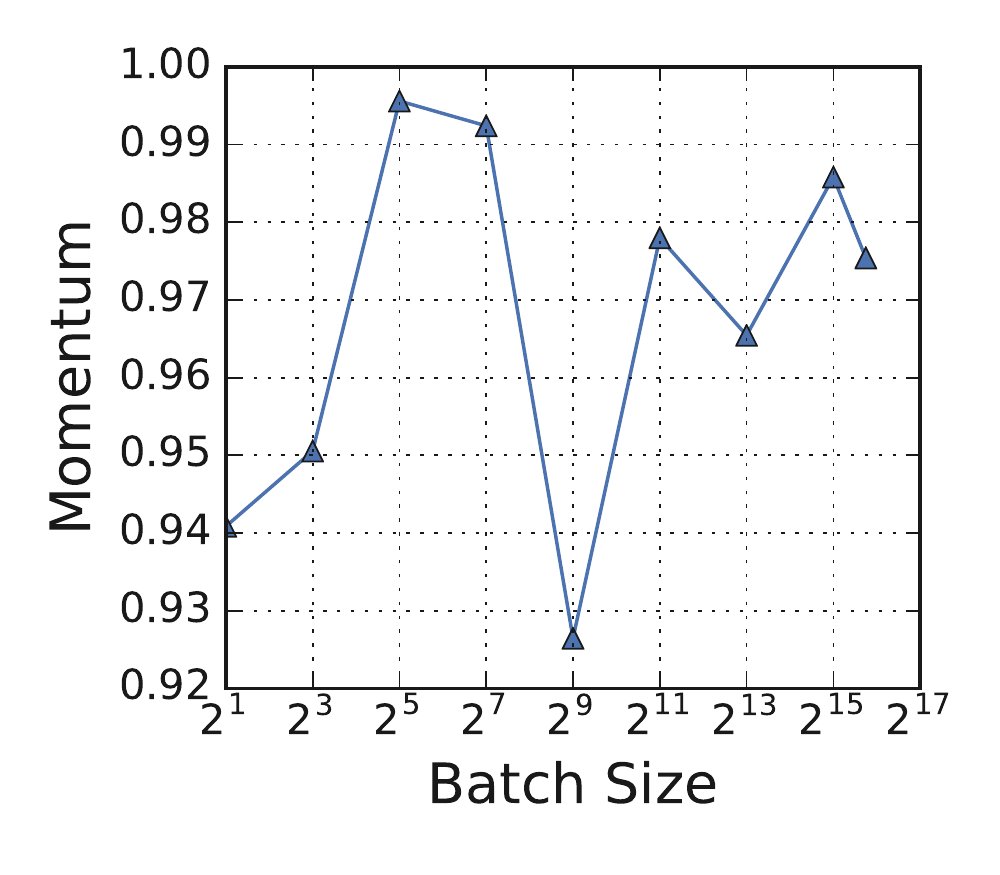}
        \vspace*{\capshift}
        \caption{Simple CNN on Fashion MNIST}
        \label{fig:hparams-momentum-cnn-fmnist}
    \end{subfigure}
    \begin{subfigure}[b]{\threecolfigwidth}
        \includegraphics[width=\textwidth]{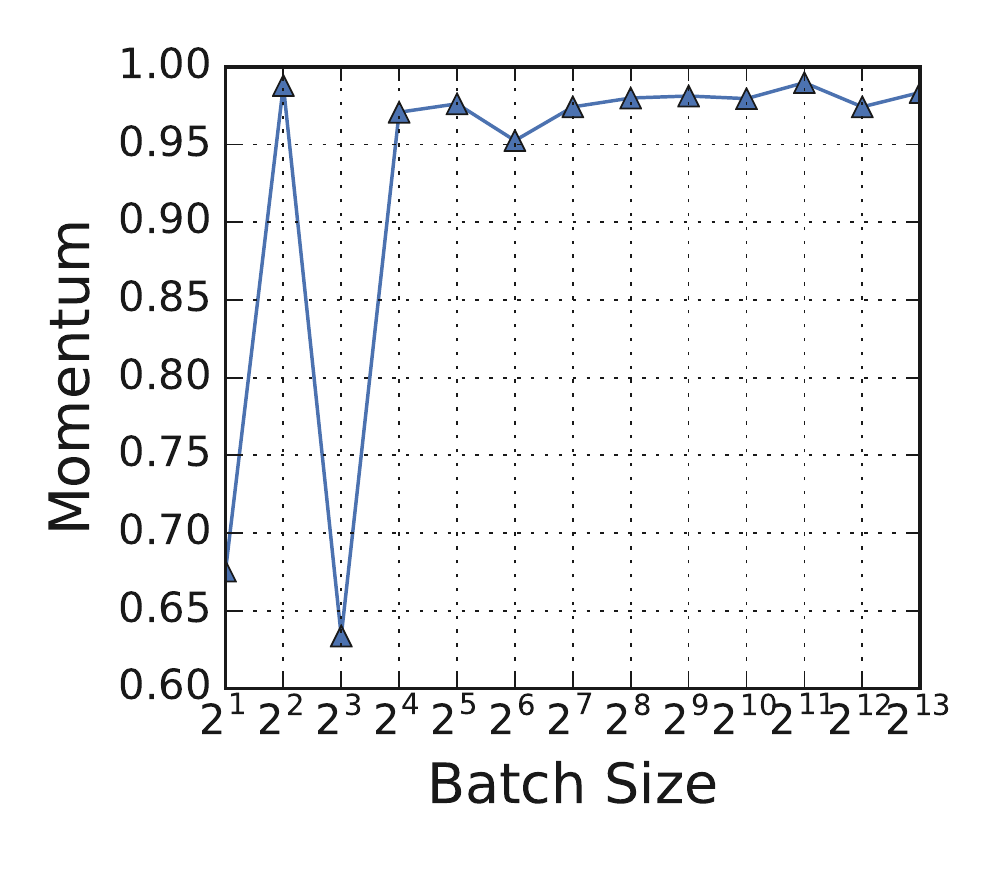}
        \vspace*{\capshift}
        \caption{ ResNet-8 on CIFAR-10}
        \label{fig:hparams-momentum-resnet-cifar}
    \end{subfigure} \\
    \vspace*{\lineshift}
    \begin{subfigure}[b]{\threecolfigwidth}
        \includegraphics[width=\textwidth]{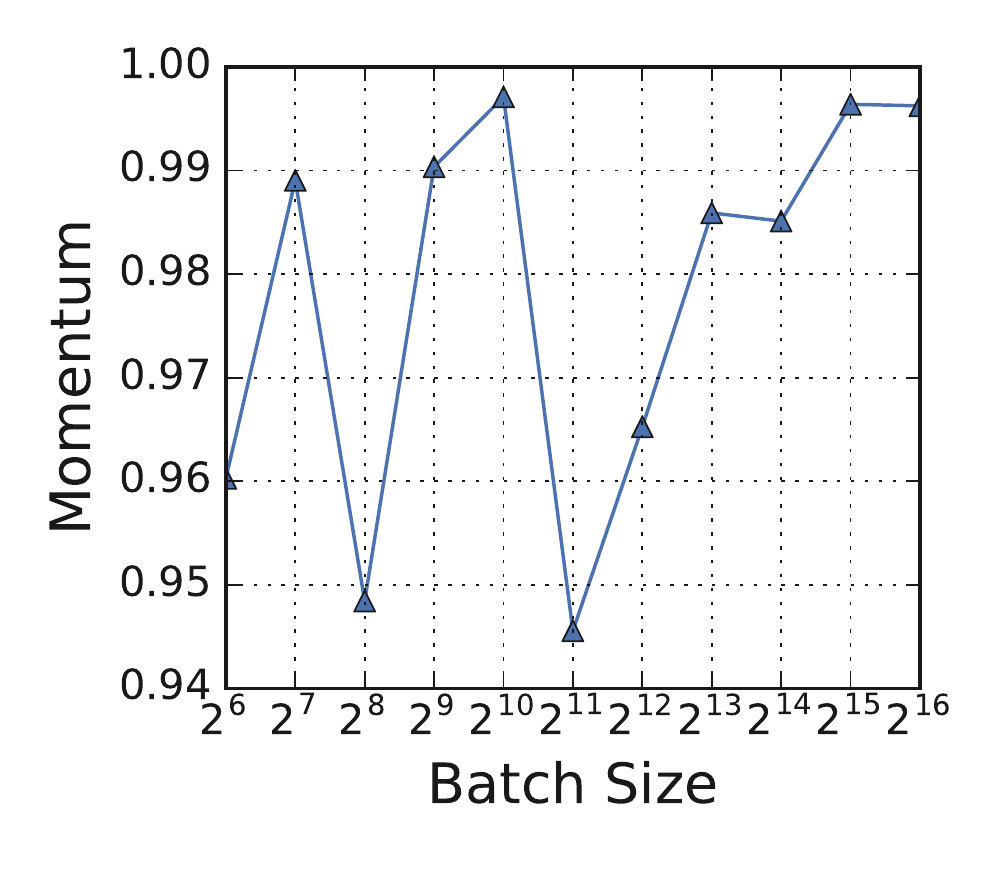}
        \vspace*{\capshift}
        \caption{ResNet-50 on ImageNet}
        \label{fig:hparams-momentum-resnet-imagenet}
    \end{subfigure}
    \begin{subfigure}[b]{\threecolfigwidth}
        \includegraphics[width=\textwidth]{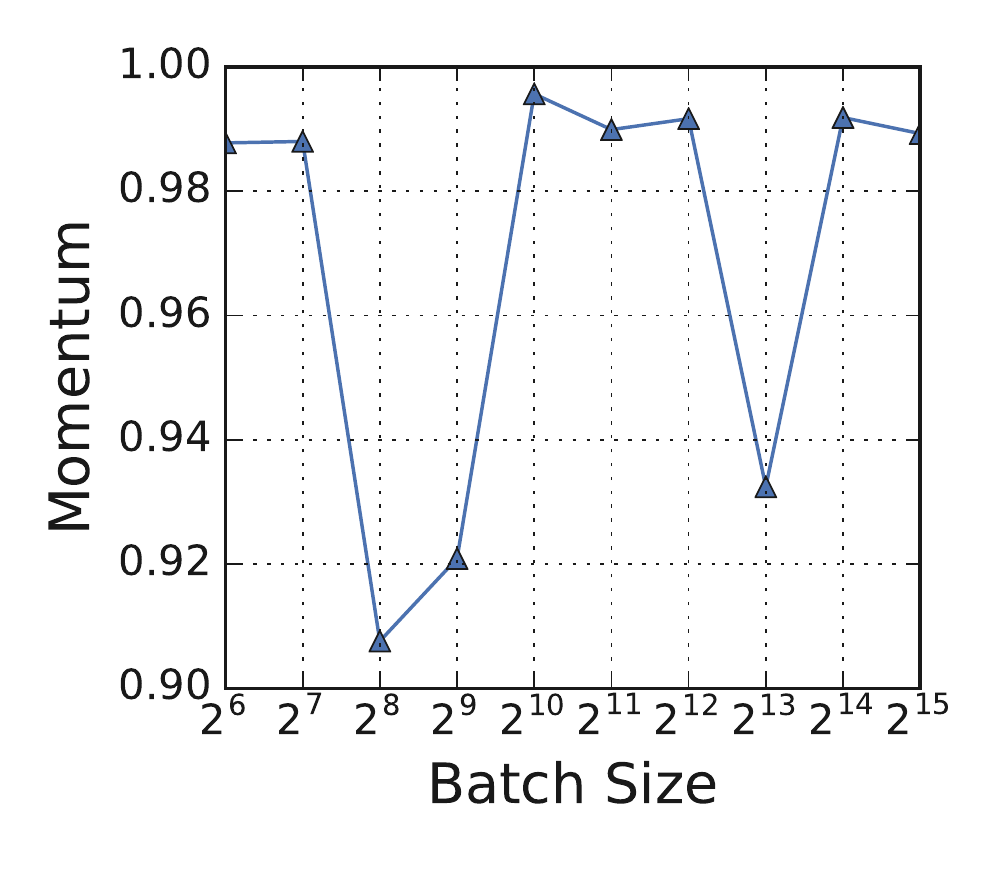}
        \vspace*{\capshift}
        \caption{ ResNet-50 on Open Images}
        \label{fig:hparams-momentum-resnet-oi}
    \end{subfigure}
    \begin{subfigure}[b]{\threecolfigwidth}
        \includegraphics[width=\textwidth]{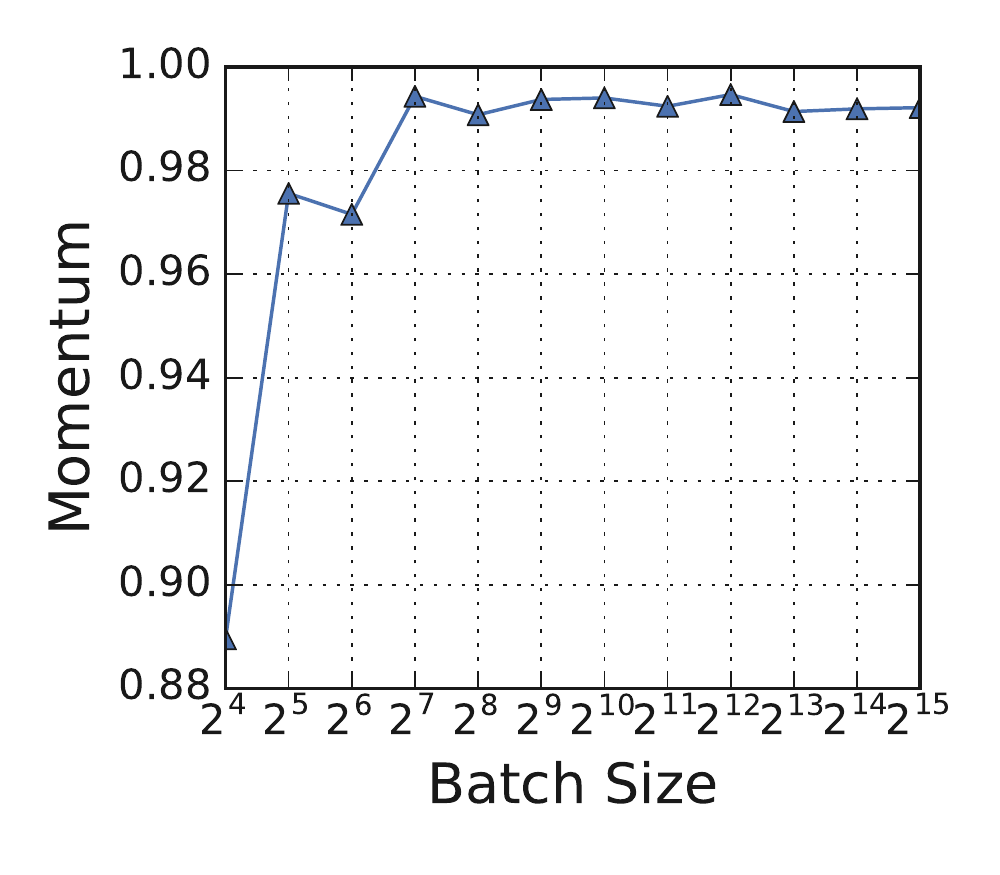}
        \vspace*{\capshift}
        \caption{ Transformer on LM1B}
        \label{fig:hparams-momentum-transformer-lm1b}
    \end{subfigure}\\
    \vspace*{\lineshift}
    \begin{subfigure}[b]{\threecolfigwidth}
        \includegraphics[width=\textwidth]{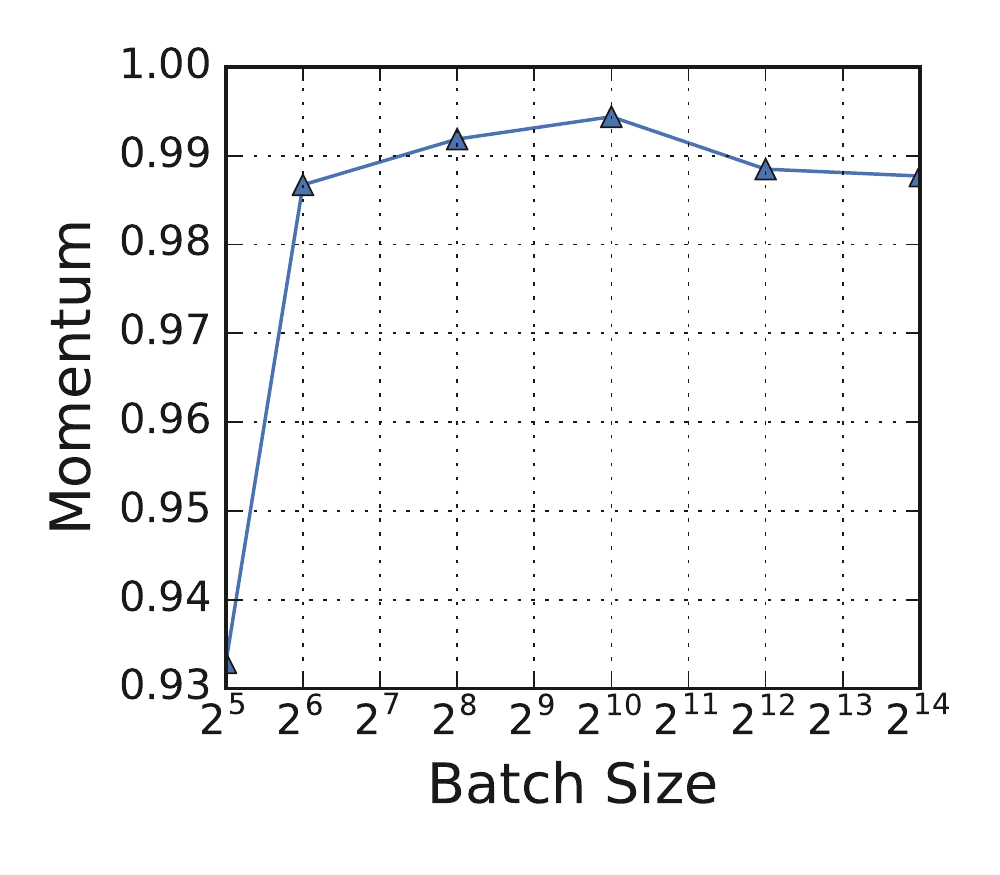}
        \vspace*{\capshift}
        \caption{Transformer on Common Crawl}
        \label{fig:hparams-momentum-transformer-cc}
    \end{subfigure}
    \begin{subfigure}[b]{\threecolfigwidth}
        \includegraphics[width=\textwidth]{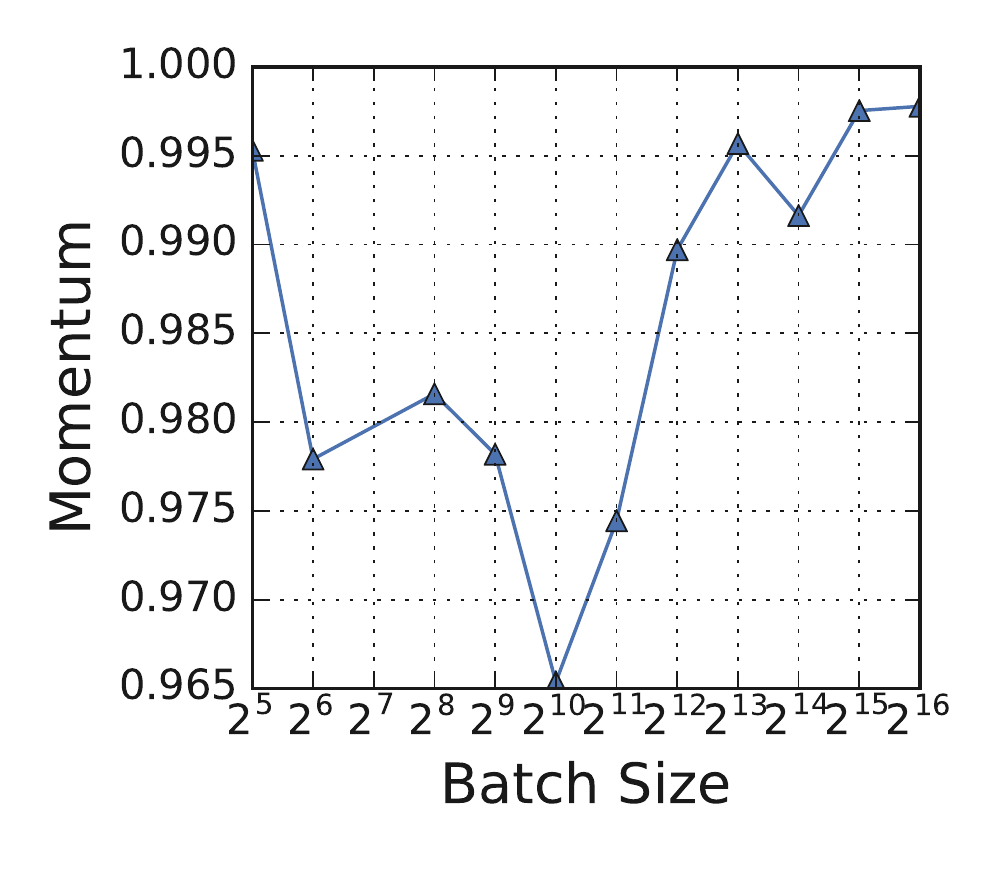}
        \vspace*{\capshift}
        \caption{ VGG-11 on ImageNet}
        \label{fig:hparams-momentum-vgg11-imagenet}
    \end{subfigure}
        \begin{subfigure}[b]{\threecolfigwidth}
        \includegraphics[width=\textwidth]{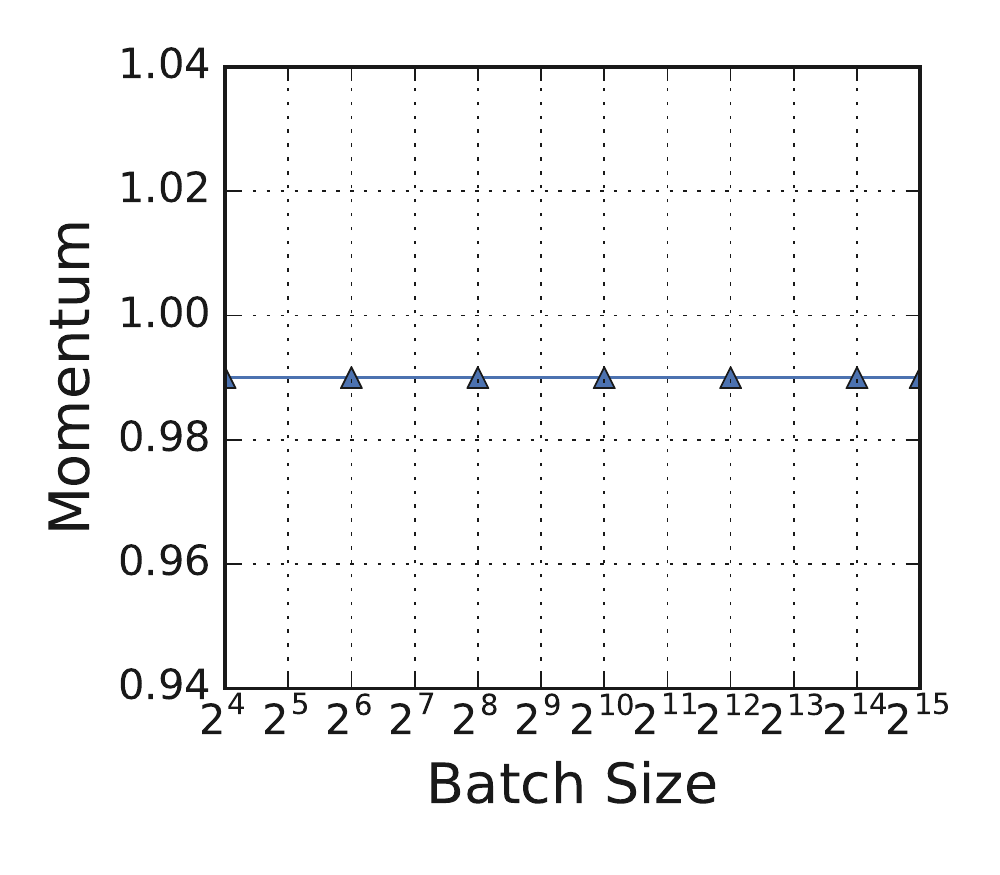}
        \vspace*{\capshift}
        \caption{LSTM on LM1B*}
        \label{fig:hparams-momentum-lstm-lm1b}
    \end{subfigure}
    \caption{\textbf{Optimal momentum has no consistent relationship with batch size.} Momentum corresponds to the trial that reached the goal validation error in the fewest training steps (see Figure~\ref{fig:stt-problems}). See Figure~\ref{fig:raw-lr} for the corresponding plot of optimal learning rate, and Figure~\ref{fig:effective-lr} for the corresponding plot of effective learning rate. *For LSTM on LM1B, we only tuned $\eta$ with fixed $\gamma=0.99$.}
    \label{fig:momentum}
\end{figure}

\newcommand{\solquazoomlhspace}{1mm}
\newcommand{\solqualzoomfigwidth}{0.48\textwidth}
\newcommand{\solqualzoomcapshift}{-1mm}
\newcommand{\solqualzoomlineshift}{3mm}

\begin{figure}
    \centering
    \begin{tabular}{
        >{\centering\arraybackslash}m{3in}
        >{\centering\arraybackslash}m{3in}
        }
    \toprule
    Step budget & Epoch budget \\
    \midrule
    \begin{subfigure}[b]{\textwidth}
        \centering
        \includegraphics[width=\solqualzoomfigwidth]{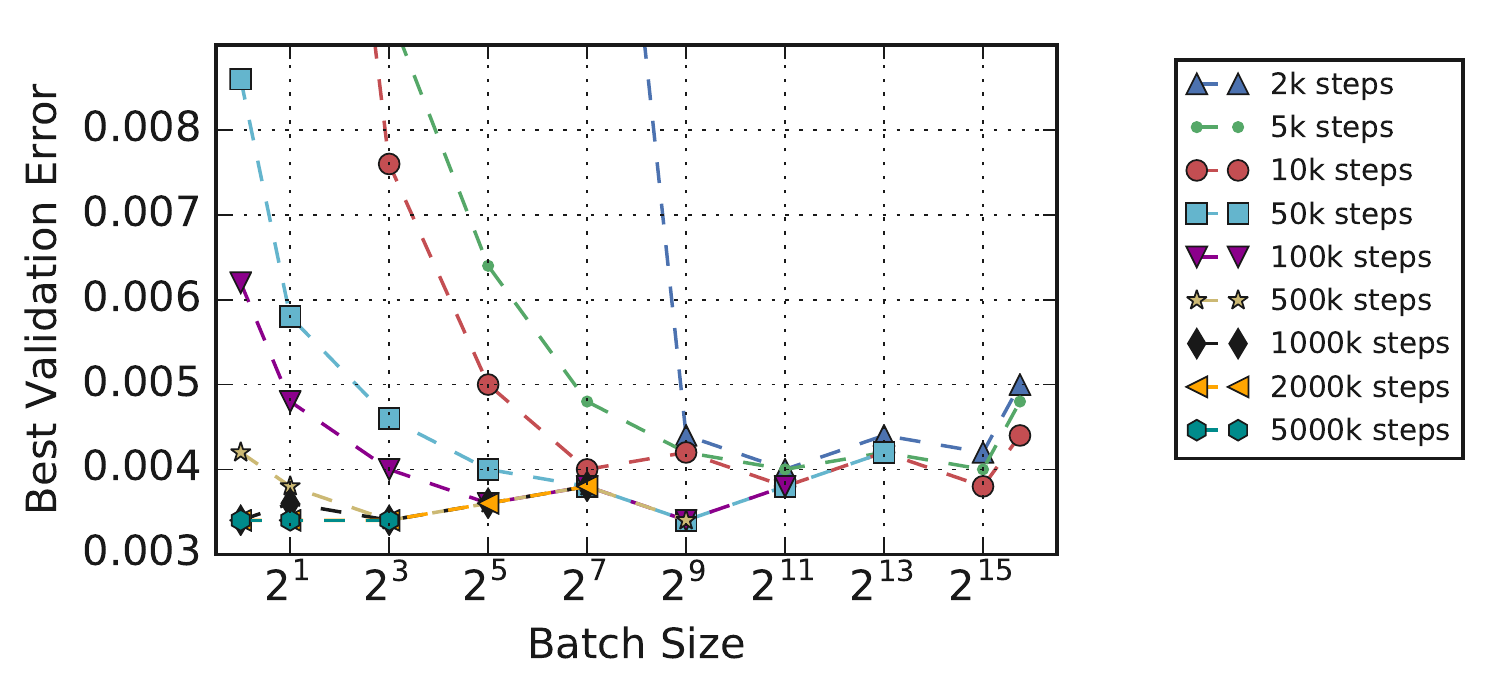}
        \hspace*{\solquazoomlhspace}
        \includegraphics[width=\solqualzoomfigwidth]{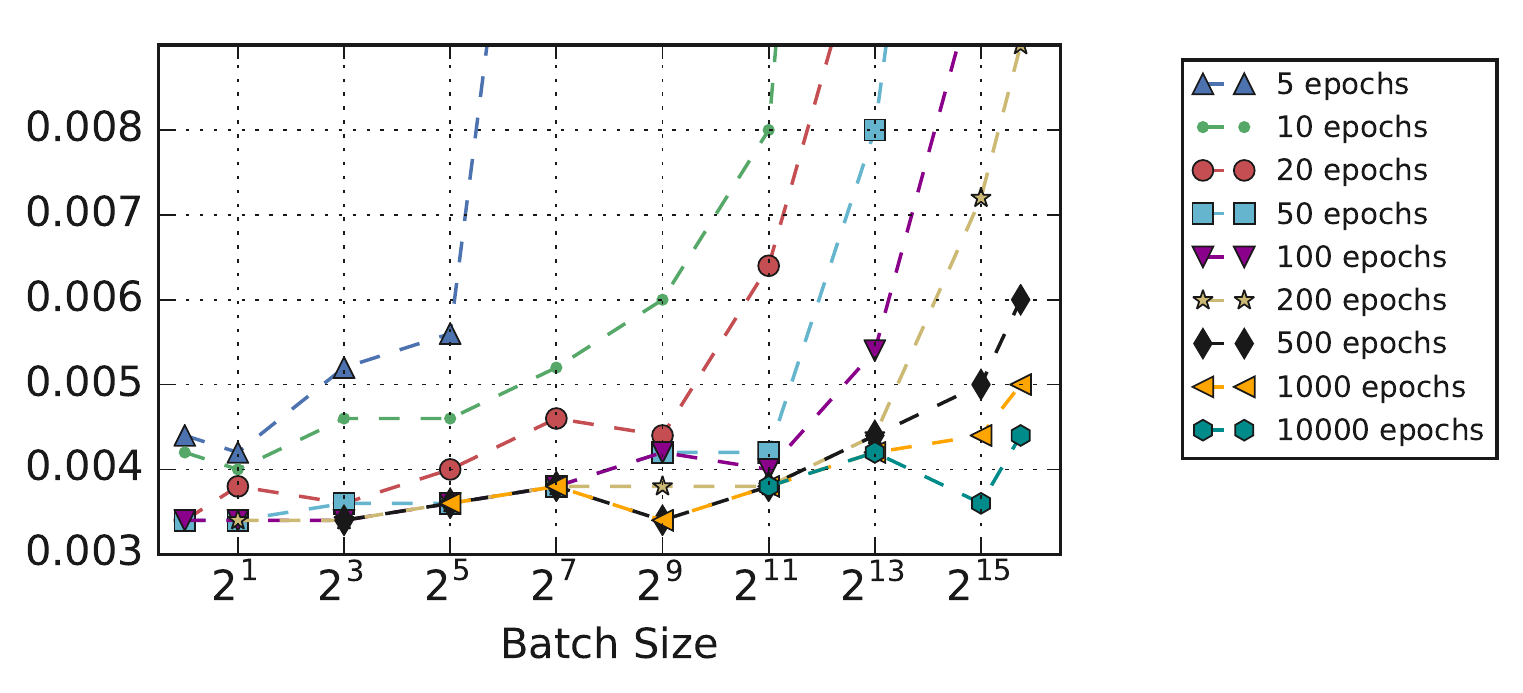}
        \vspace*{\solqualzoomcapshift}
        \caption{\small Simple CNN on MNIST: Validation Error}
    \end{subfigure}\\
    \vspace*{\solqualzoomlineshift}
    \begin{subfigure}[b]{\textwidth}
        \centering
        \includegraphics[width=\solqualzoomfigwidth]{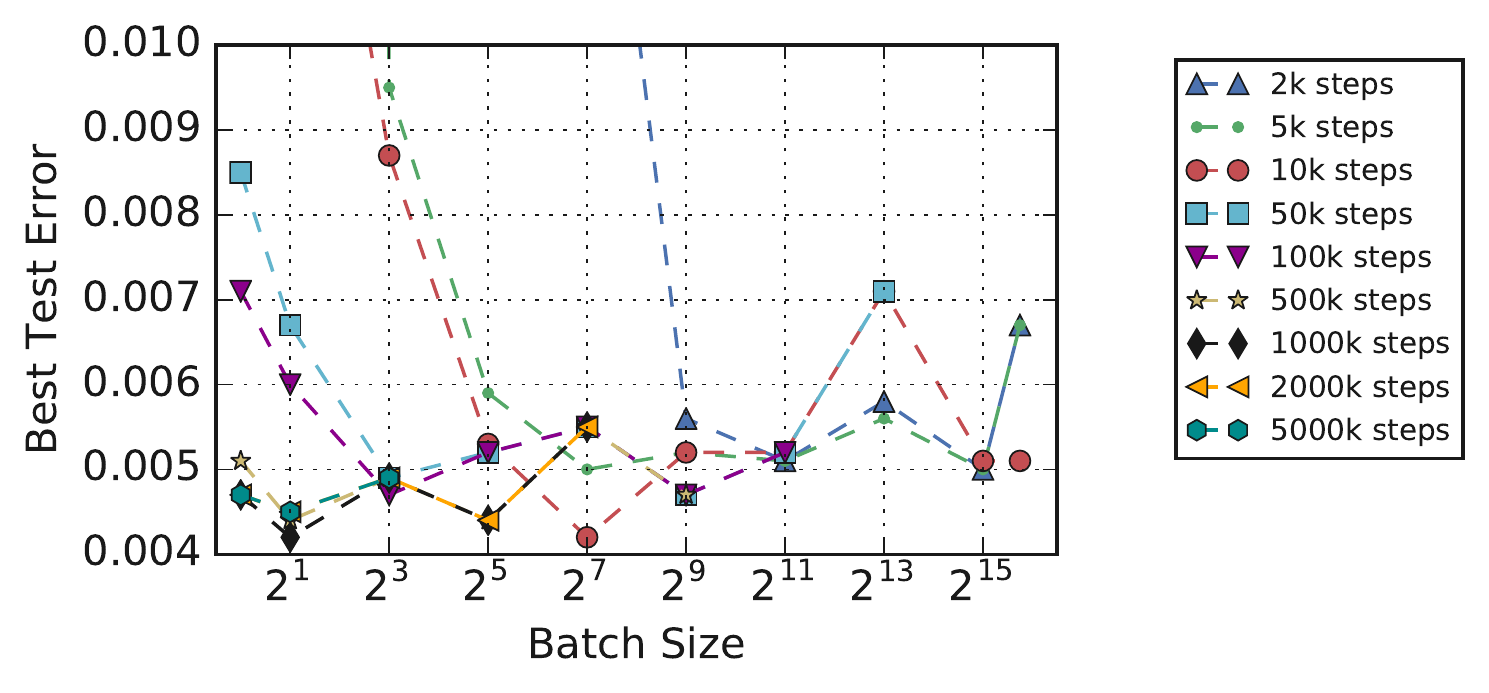}
        \hspace*{\solquazoomlhspace}
        \includegraphics[width=\solqualzoomfigwidth]{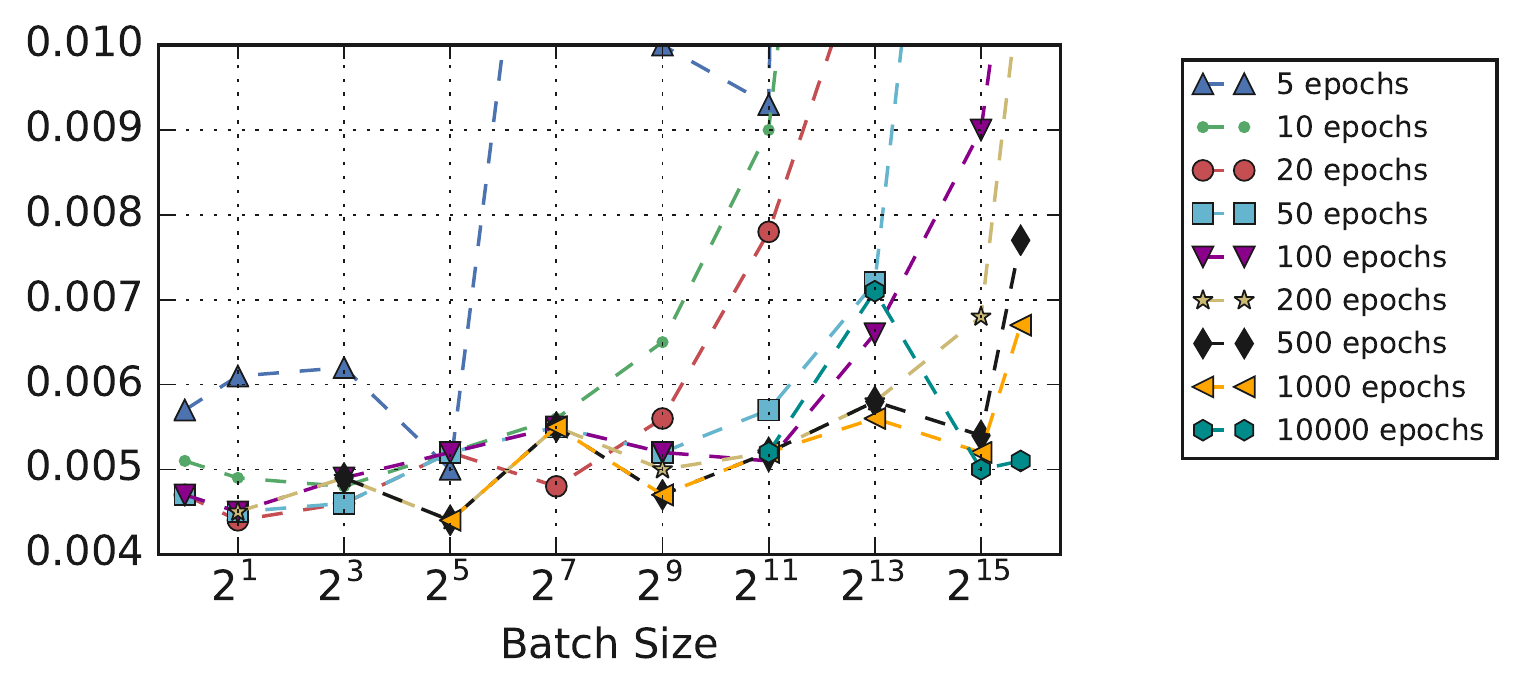}
        \vspace*{\solqualzoomcapshift}
        \caption{\small Simple CNN on MNIST: Test Error}
    \end{subfigure}    
    \end{tabular}
    \caption{\textbf{Zoomed version of Figure~\ref{fig:sol-qual-mnist}.}}
    \label{fig:sol-qual-zoom-mnist}
\end{figure}

\begin{figure}
    \centering
    \begin{tabular}{
        >{\centering\arraybackslash}m{3in}
        >{\centering\arraybackslash}m{3in}
        }
    \toprule
    Step budget & Epoch budget \\
    \midrule
    \begin{subfigure}[b]{\textwidth}
        \centering
        \includegraphics[width=\solqualzoomfigwidth]{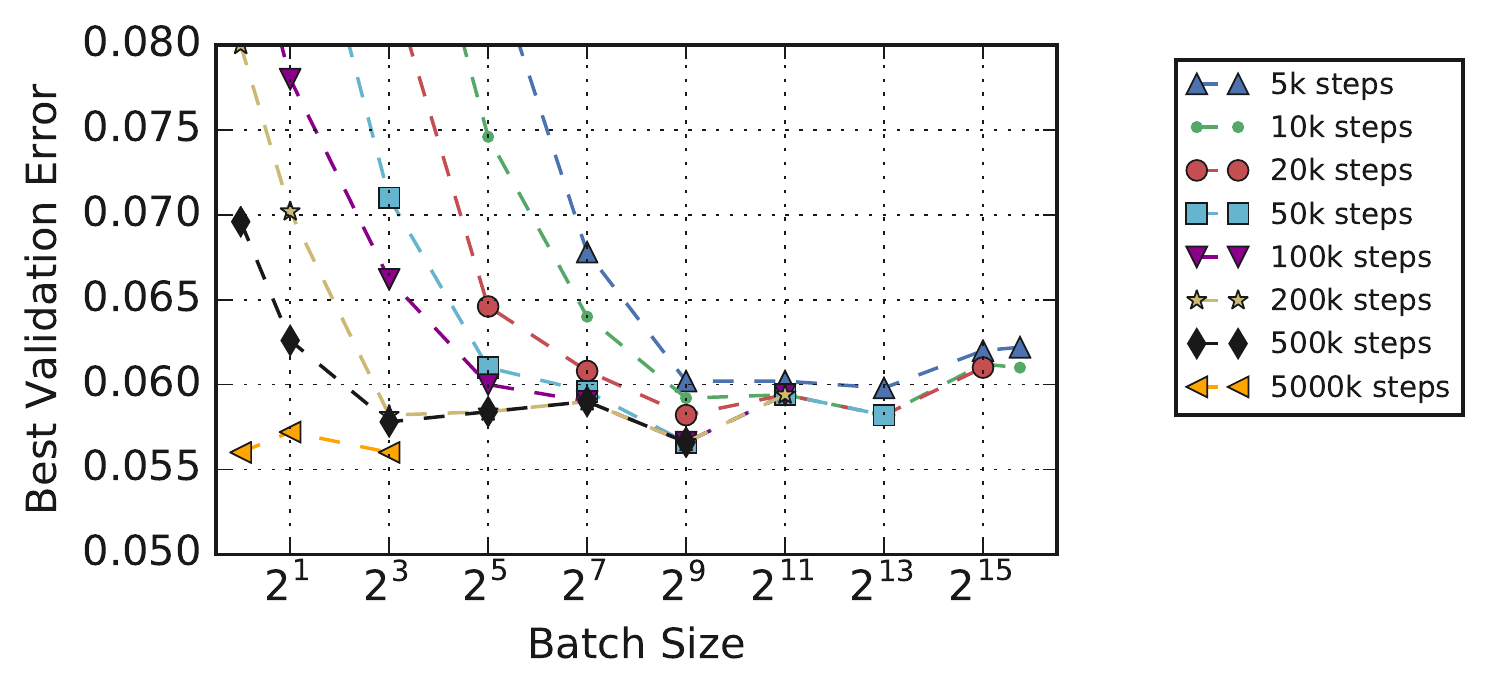}
        \hspace*{\solquazoomlhspace}
        \includegraphics[width=\solqualzoomfigwidth]{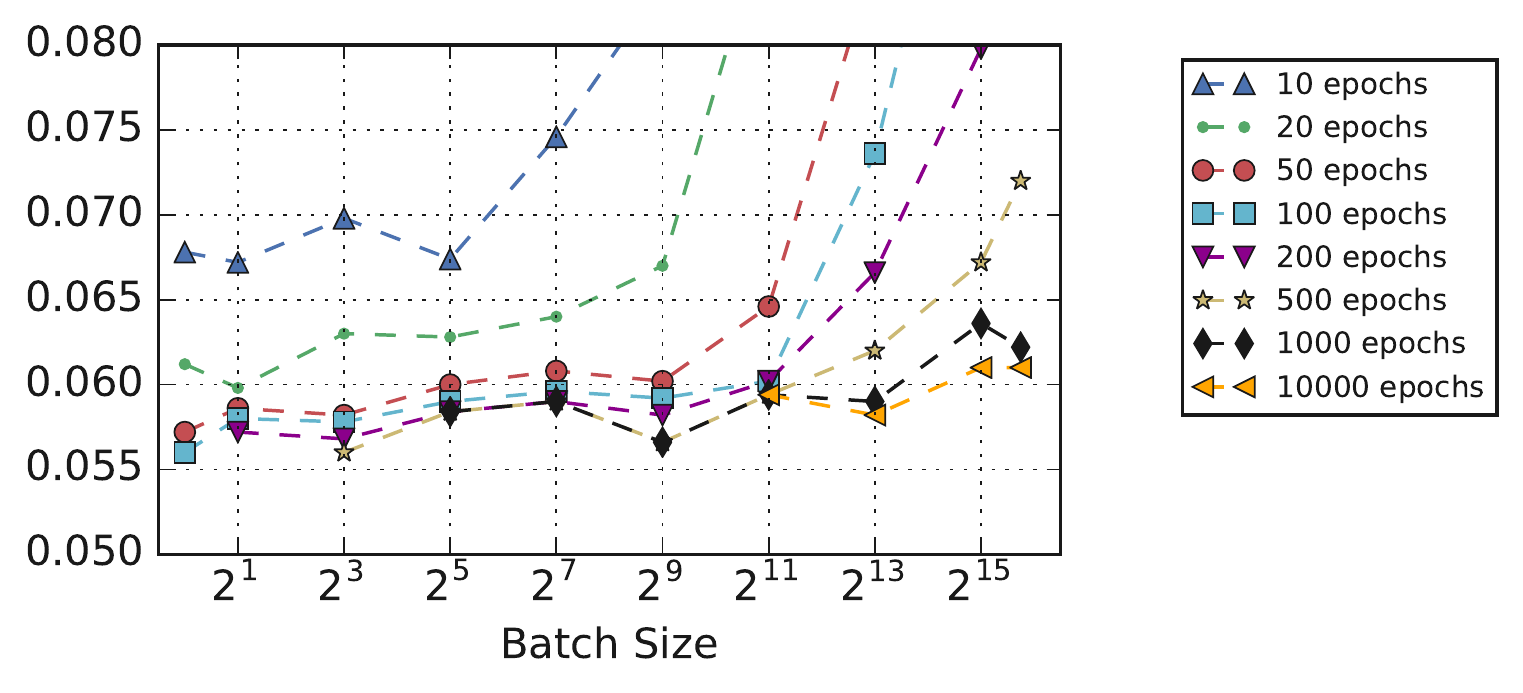}
        \vspace*{\solqualzoomcapshift}
        \caption{\small Simple CNN on Fashion MNIST: Validation Error}
    \end{subfigure}\\
    \vspace*{\solqualzoomlineshift}
    \begin{subfigure}[b]{\textwidth}
        \centering
        \includegraphics[width=\solqualzoomfigwidth]{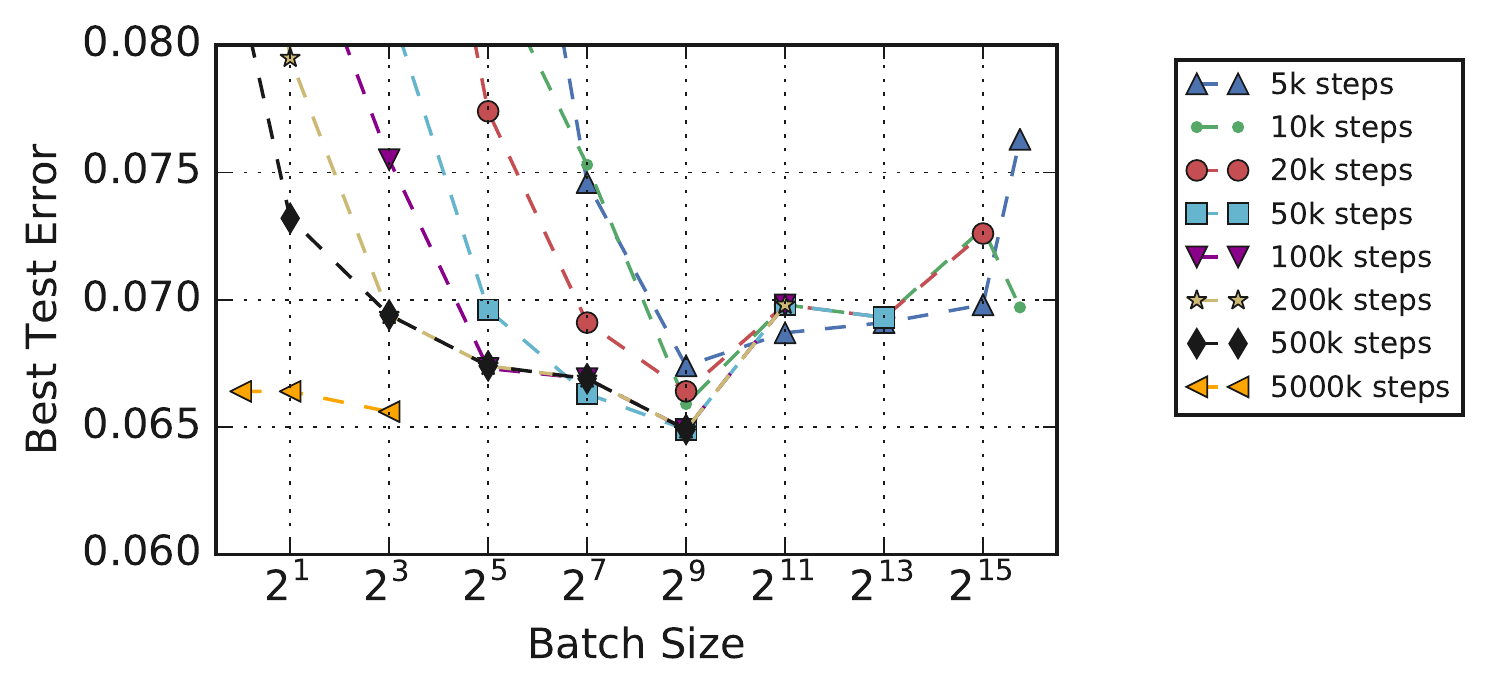}
        \hspace*{\solquazoomlhspace}
        \includegraphics[width=\solqualzoomfigwidth]{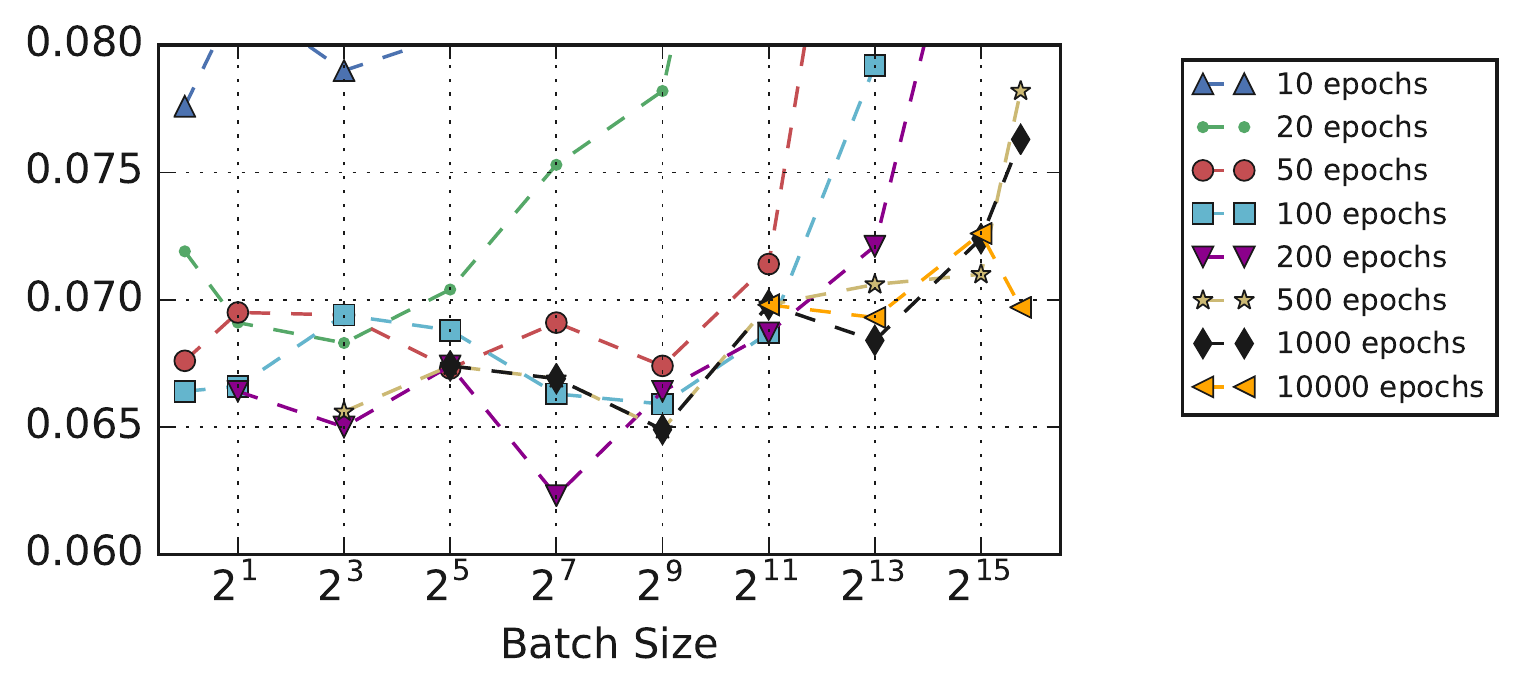}
        \vspace*{\solqualzoomcapshift}
        \caption{\small Simple CNN on Fashion MNIST: Test Error}
    \end{subfigure}    
    \end{tabular}
    \caption{\textbf{Zoomed version of Figure~\ref{fig:sol-qual-fashion-mnist}.}}
    \label{fig:sol-qual-zoom-fmnist}
\end{figure}

\pagebreak
\pagebreak
\bibliography{batchScience}

\end{document}